\documentclass[10pt]{article} 
\usepackage[accepted]{tmlr}

\usepackage{amsmath,amsfonts,bm}

\def\eqref#1{equation~\ref{#1}}

\def\1{\bm{1}}

\DeclareMathAlphabet{\mathsfit}{\encodingdefault}{\sfdefault}{m}{sl}
\SetMathAlphabet{\mathsfit}{bold}{\encodingdefault}{\sfdefault}{bx}{n}

\newcommand*{\MinNumber}{-1.0}%
\newcommand*{\MidNumber}{0.0} %
\newcommand*{\MaxNumber}{1.0}%

\newcommand{\ApplyGradient}[1]{%
        \ifdim #1 pt > \MidNumber pt
            \pgfmathsetmacro{\PercentColor}{max(min(100.0*(#1 - \MidNumber)/(\MaxNumber-\MidNumber),100.0),0.00)} %
            \hspace{-0.33em}\colorbox{green!\PercentColor!yellow}{#1}
        \else
            \pgfmathsetmacro{\PercentColor}{max(min(100.0*(\MidNumber - #1)/(\MidNumber-\MinNumber),100.0),0.00)} %
            \hspace{-0.33em}\colorbox{red!\PercentColor!yellow}{#1}
        \fi
}

\newcommand*{\MinNumberSub}{-1}%
\newcommand*{\MidNumberSub}{0.0} %
\newcommand*{\MaxNumberSub}{1}%
\newcommand{\ApplyGradientSub}[1]{%
        \ifdim #1 pt > \MidNumberSub pt
            \pgfmathsetmacro{\PercentColor}{max(min(100.0*(#1 - \MidNumberSub)/(\MaxNumberSub-\MidNumberSub),100.0),0.00)} %
            \hspace{-0.33em}\transparent{0.7}\colorbox{green!\PercentColor!yellow}{#1}
        \else
            \pgfmathsetmacro{\PercentColor}{max(min(100.0*(\MidNumberSub - #1)/(\MidNumberSub-\MinNumberSub),100.0),0.00)} %
            \hspace{-0.33em}\transparent{0.6}\colorbox{red!\PercentColor!yellow}{#1}
        \fi
}

\newcommand{\entry}[2]{{#1} \textsubscript{\ApplyGradientSub{#2}}}
\newcommand{\entrya}[2]{\entry{#1}{#2}}
\newcommand{\entrydualg}[2]{{\ApplyGradient{#1}}{\ApplyGradient{#2}}}
\newcommand{\entrydual}[2]{\entrydualg{#1}{#2}}
\newcommand{\entrydualvar}[2]{{#1}  {#2}}

\usepackage[utf8]{inputenc} 
\usepackage[T1]{fontenc}    
\usepackage{hyperref}       
\usepackage{url}            
\usepackage{booktabs}       
\usepackage{amsfonts}       
\usepackage{nicefrac}       
\usepackage{microtype}      
\usepackage{tcolorbox}         
\usepackage{amsmath}
\usepackage{cleveref}       
\usepackage{graphicx}
\usepackage{subcaption}
\usepackage{listings}
\usepackage{natbib}
\usepackage{multirow}
\usepackage{tabularx}
\usepackage{tikz}
\usepackage{pgfplots}
\usepackage{pgf}
\usepackage{collcell}
\usepackage{transparent}
\usepackage{wrapfig}
\usepackage{booktabs} 
\usepackage{float}
\pgfplotsset{compat=1.18}
\colorlet{light-gray}{gray!30}

\newcommand{\ourname}{\textbf{Syn}thetic Dataset \textbf{Qu}ality \textbf{E}stimation}
\newcommand{\ournameshort}{\textsc{SynQuE}}

\newcommand{\ourmethodname}{\textbf{L}LM-\textbf{E}valuated \textbf{N}ormalized \textbf{S}core}
\newcommand{\ourmethodnameshort}{\textsc{Lens}}

\newcommand{\dataset}{\mathcal{D}}
\newcommand{\samples}{\mathcal{U}}
\newcommand{\real}{\mathrm{r}}
\newcommand{\synth}{\mathrm{s}}
\newcommand{\performance}{M}
\newcommand{\proxy}{Q}
\newcommand{\characterization}{C}
\newcommand{\mmd}{\mathbf{MMD}^2}
\newcommand{\mdm}{\mathbf{MDM}}
\newcommand{\pad}{\mathbf{PAD}}
\newcommand{\mauve}{\textsc{Mauve}}
\newcommand{\perplexity}{\textsc{Perplexity}}
\newcommand{\expect}[1]{\mathbb{E}\left[{#1}\right]}

\title{\textsc{SynQuE}: Estimating Synthetic Dataset Quality Without Annotations}

\author{\name Arthur Chen \email haonan.chen@uwaterloo.ca \\
      \addr David R. Cheriton School of Computer Science \\
      University of Waterloo \\
      Waterloo, Ontario, Canada
      \AND
      \name Victor Zhong \email victor.zhong@uwaterloo.ca \\
      \addr David R. Cheriton School of Computer Science \\
      University of Waterloo \\
      Waterloo, Ontario, Canada
}

\def\month{04}  
\def\year{2026} 
\def\openreview{\url{https://openreview.net/forum?id=W4Pwb4SX3P}} 

\begin{document}

\maketitle

\begin{abstract}
We introduce and formalize the \textbf{\ourname} (\ournameshort) problem: ranking synthetic datasets by their expected real-world task performance using only limited unannotated real data.
This addresses a critical and open challenge where data is scarce due to collection costs or privacy constraints.
We establish the first comprehensive benchmarks for this problem by introducing and evaluating proxy metrics that choose synthetic data for training to maximize task performance on real data.
We introduce the first proxy metrics for \ournameshort\ by adapting distribution and diversity-based distance measures to our context via embedding models.
To address the shortcomings of these metrics on complex planning tasks, we propose \ourmethodnameshort, a novel proxy that leverages large language model reasoning.
Our results show that \ournameshort\ proxies correlate with real task performance across diverse tasks, including sentiment analysis, Text2SQL, web navigation, and image classification, with \ourmethodnameshort\ consistently outperforming others on complex tasks by capturing nuanced characteristics.
For instance, on text-to-SQL parsing, training on the top-3 synthetic datasets selected via \ournameshort\ proxies can raise accuracy from 30.4\% to 38.4 (+8.1)\% on average compared to selecting data indiscriminately.
This work establishes \ournameshort\ as a practical framework for synthetic data selection under real-data scarcity and motivates future research on foundation model-based data characterization and fine-grained data selection.
We release our code\footnote{\url{https://github.com/r2llab/SynQuE}}.
\end{abstract}

\section{Introduction}
\label{sec:introduction}

Data scarcity hinders effective machine learning, especially for tasks requiring specialized expertise like autonomous navigation or natural language interfaces, where data collection is costly and slow~\citep{xie_osworld_2024, yang_synthesizing_2024}.
In sensitive domains such as healthcare and finance~\citep{tan_large_2024, jordan_machine_2015}, privacy concerns further complicate data acquisition.
Large generative models have emerged as capable synthetic data generators, producing annotated data for tasks like policy learning~\citep{xu_agenttrek_2024}, Text2SQL~\citep{yang_synthesizing_2024}, sentiment analysis~\citep{ye_zerogen_2022, li_synthetic_2023}, and image classification~\citep{geng_unmet_2025}.
While synthetic data can improve real-world performance under scarcity, results vary widely depending on task and data quality~\citep{huang_survey_2025, geng_unmet_2025}.

Can we distinguish between high-quality synthetic data that improves real-world task performance and low-quality data that offers little benefit, \textit{without any annotated real data} and \textit{without costly model training}?
Crucially, increasing the size of synthetic datasets does not always lead to better downstream performance as it does with real data; in some cases, larger synthetic datasets can even degrade performance, exhibiting inverse scaling trends~\citep{geng_unmet_2025, li_synthetic_2023, setlur_rl_2024, gao_self-guided_2022, moller_parrot_2023}.
Therefore, selecting a synthetic dataset from a pool of datasets to train on to optimize downstream performance is important.

We introduce \ourname, or \ournameshort, the problem of ranking multiple synthetic datasets by quality using only limited unannotated samples of real data.
A synthetic dataset A is of higher quality than B if a model trained on A outperforms one trained on B on a real-world test set.
This ability is crucial when real data annotation is costly or infeasible.
For example, in text-to-SQL parsing, \ournameshort\ helps select the synthetic dataset that yields better generalization from a small set of unannotated real queries.
Similarly, for intelligent web agents, it identifies the synthetic interactions that produce agents performing best on real navigation tasks.

This work makes three main contributions.
1) We formalize the \ournameshort\ problem and establish the first comprehensive benchmark.
As part of this, we introduce and evaluate a suite of \textit{proxy metrics}: computable scores that estimate a synthetic dataset's quality using only a subset of unannotated real data.
We adapt proxy metrics using established distributional measures such as mean distance to medoids~\citep{cox_directed_2021}, diversity measures such as Proxy-A-Distance~\citep{ben-david_analysis_2006}, and divergence measures such as MAUVE~\citep{pillutla_mauve_2021} --- none have been systematically evaluated for the purpose of synthetic data selection.
2) We propose \ourmethodname\ (\ourmethodnameshort), a novel proxy measure that leverages large language model (LLM) reasoning to create \textit{dataset rubrics} that highlight difference between synthetic data and real data in language.
3) We conduct a comprehensive experiment across diverse domains in sentiment analysis, Text2SQL, web navigation, and image classification in order to evaluate how well these proxy metrics are able to select synthetic data to maximize performance on real test data.

Our empirical evaluation shows that \ournameshort\ proxies exhibit moderate to strong correlation with real task performance across diverse domains including sentiment analysis, Text-to-SQL, image classification, and web navigation.
While the best proxy varies by task, most can effectively predict downstream performance without \textit{any} labeled real data, enabling practical synthetic data selection that outperforms indiscriminate synthetic data selection.
We find that the reliability of proxies depends on task complexity, with higher variance on noisy data like synthetic images.
Among the proxies, \ourmethodnameshort, leveraging LLM reasoning and a principled debiasing strategy, consistently achieves superior gains on complex tasks like web navigation by capturing nuanced task details beyond embedding-based metrics.
These results establish \ournameshort\ as a robust framework for selecting high-quality synthetic data and motivate future work on stronger foundation model methods to characterize data and perform fine-grained, example-level data selection.

\begin{figure}[t]
    \centering
    \vspace{-2em}
    \includegraphics[width=1\linewidth]{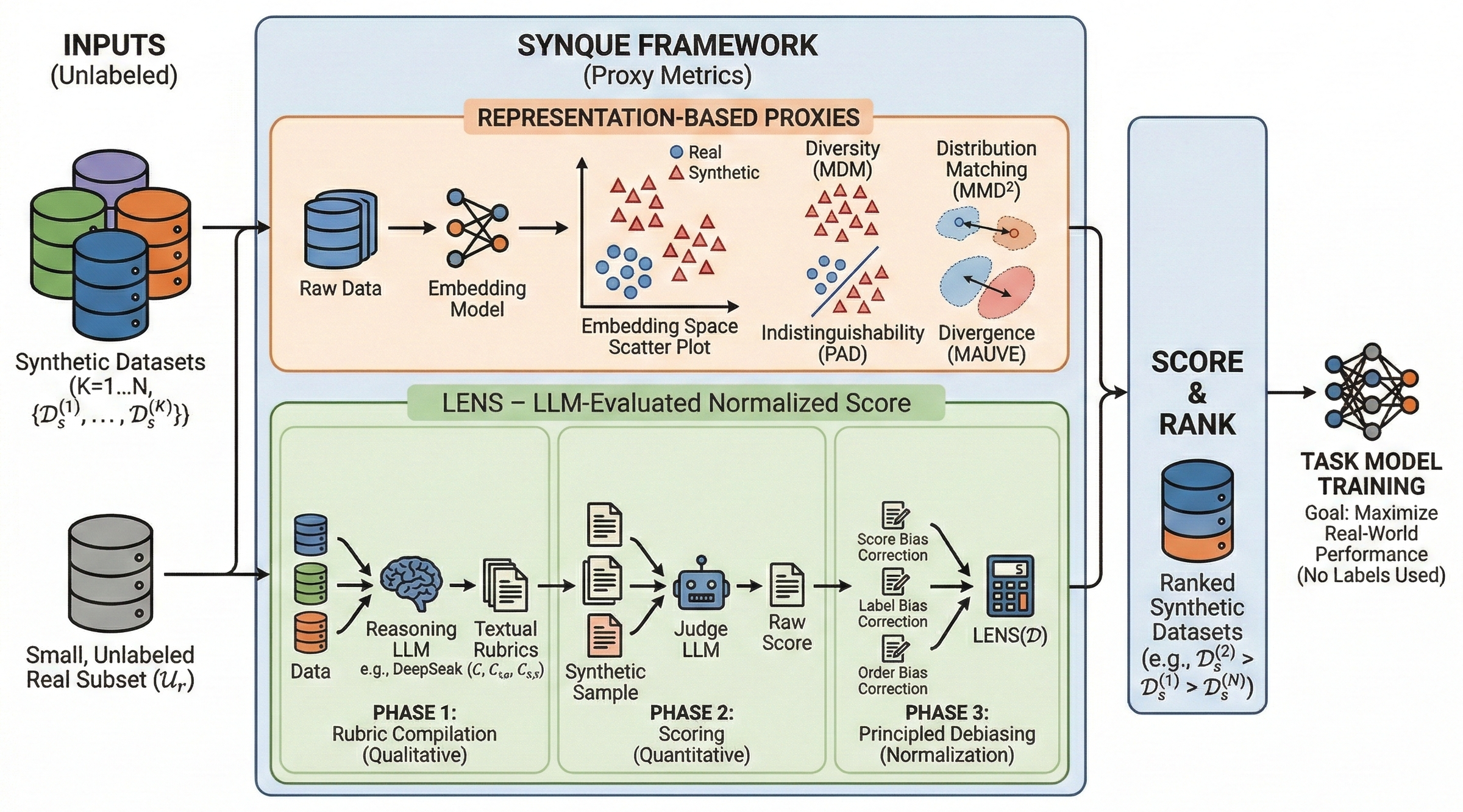}
    \caption{\ournameshort\ uses synthetic data and unlabeled samples of real data to estimate synthetic data quality. Proxy scores are used to rank and select the datasets that lead to the best task performance.}
    \label{fig:synque}
    \vspace{-1.5em}
\end{figure}

\section{Related Work}
\paragraph{Data synthesis}
Synthesizing training data with generative models is a promising way to address data scarcity by leveraging instruction-following abilities~\citep{touvron_llama_2023,ouyang_training_2022} and vast pre-trained knowledge~\citep{long_llms-driven_2024,honovich_unnatural_2022,mishra_cross-task_2022}.
This has been applied across domains such as text classification ~\citep{ye_zerogen_2022}, Text-to-SQL ~\citep{yang_synthesizing_2024, lei_spider_2024, li_codes_2024}, planning ~\citep{sun_os-genesis_2024, hu_agentgen_2024, xu_agenttrek_2024,murty_nnetnav_2025}, and computer vision ~\citep{geng_unmet_2025, li_bigdatasetgan_2022}.
While synthetic data often improves downstream models~\citep{liu_apigen_2024, ye_zerogen_2022}, challenges remain in ensuring quality due to issues such as hallucination~\citep{huang_survey_2025}, mode collapse~\citep{goodfellow_generative_2014, durall_combating_2020, shumailov_ai_2024}, and counterfactual artifacts~\citep{li_synthetic_2023, yu_large_2023}.
Our work introduces the \ournameshort\ framework for selecting synthetic data to maximize task performance without access to large amounts of real data and without training task models.

\paragraph{Distributional and diversity metrics}
Metrics like Proxy-$\mathcal{A}$-Distance ($\pad$) estimate domain divergence by training classifiers to distinguish source and target data~\citep{ben-david_theory_2010, ben-david_analysis_2006, quinonero-candela_dataset_2022}.
While effective in some NLP tasks~\citep{elsahar_annotate_2019}, $\pad$ struggles in noisy, complex settings such as agent planning~\citep{he_webvoyager_2024}.
Diversity metrics like mean-distance-to-medoids ($\mdm$)\citep{cox_directed_2021} and DCScore\citep{zhu_measuring_2025} approximate the coverage of synthetic data.
However, these measures do not capture common synthetic data factual inconsistencies from generative model hallucinations~\citep{huang_survey_2025, li_evaluating_2023,gunjal_detecting_2023}, which can harm downstream performance~\citep{geng_unmet_2025, yu_large_2023,casco-rodriguez_self-consuming_2023,li_synthetic_2023,hataya_will_2023,briesch_large_2023}.
\paragraph{Data selection and weighting}
A growing body of work focuses on selecting high-value training samples.
Some approaches utilize scaling laws by training regression models on short runs to predict the best data mixture~\citep{liu_regmix_2025, magnusson_datadecide_2025}.
Others employ influence functions or gradient-based heuristics to select influential subsets~\citep{zhang_harnessing_2024, hu_most_2025}, or utilize small real datasets to learn instance-level weights for synthetic data~\citep{kuo_not_2025}.
Crucially, these methods focus on \textit{instance-level} selection or weighting, often requiring access to labeled validation sets to compute gradients or train selection models.
In contrast, \ournameshort\ addresses \textit{dataset-level} ranking in a zero-shot, gradient-free manner, making it suitable for ``cold-start'' settings where no labeled data exists.
Finally, another group of works try to distinguish between human and machine text using divergence measures~\citep{pillutla_mauve_2021} or LLM-as-a-judge~\citep{gu_survey_2025, krumdick_no_2025, zheng_judging_2023}.
Neither of these two techniques have been studied for synthetic data selection.
Furthermore, the latter also suffers from limitations of generative models previously mentioned such as hallucination.
Our work establishes the first systematic benchmark and results for synthetic data selection in a practical, fully unsupervised regime.
\paragraph{Dataset Ranking and Selection}
Most prior work has focused on data selection at the sample-level, aiming to curate optimal subsets of individual data points to reduce training time or labeling costs. For example, \citet{kaushal_learning_2018} leverage submodular functions to maximize diversity and coverage in data subsets, while active learning frameworks like BatchBALD \citep{kirsch_batchbald_2019} acquire batches of informative points based on Bayesian mutual information. However, ranking or selecting at the dataset-level is equally important, as it accounts for dependencies and structure among samples within a dataset. 

Recent methodologies have expanded selection to this macro scale. For instance, \citet{nadas_synthetic_2025} introduce a hierarchical framework designed to retrieve entire real-world datasets from massive, decentralized sources under strict computational constraints. Similarly, previous work on dataset retrieval, such as \citep{chapman_dataset_2020}, studies data set ranking using keyword-based queries and defines data set quality primarily in terms of search relevance. 

In contrast to both instance-level subset curation and the macro-level routing or retrieval of real data, our setting addresses the emerging challenge of selecting \textit{synthetic} datasets under \textit{label scarcity}. We are concerned with ranking candidate model-generated datasets according to their potential downstream utility when annotated validation data is unavailable. Rather than optimizing search relevance or structural efficiency across data silos, our notion of quality relies on unannotated target data to estimate real-world task performance -- a fundamentally different objective.

\section{The \ourname\ Problem}
\label{sec:problem_setting}
We define the \ournameshort\ problem and establish notation. Let $\dataset_\real = \{x_\real^{(i)}, y_\real^{(i)}\}_{i=1}^{n_\real}$ be a real dataset, which we assume is scarce and for which labels are unavailable for validation.
Instead, we only have access to a small, unannotated collection of real-world inputs $\samples_\real = \{x_i\}_{i=1}^{m_\real}\in \dataset_\real$.
Our goal is to use $\samples_\real$ to estimate the quality of $K$ synthetic datasets $\{\dataset_\synth^{(1)}, \ldots, \dataset_\synth^{(K)}\}$.
These datasets might be generated by different methods and thus vary in quality.
We aim to select the synthetic dataset $\dataset_\synth^*$ that yields the best model performance on $\dataset_\real$, \textit{without using labeled real data or training models on any synthetic dataset}.

Consider a model $f(\cdot; \theta_k)$ trained on the synthetic dataset $\dataset_\synth^{(k)}$ with parameters $\theta_k$. Let $\performance(f, \dataset_e)$ denote model $f$'s performance on the evaluation dataset $\dataset_e$, for instance task completion rate or accuracy. The ideal synthetic dataset $\dataset_\synth^*$ satisfies:
\begin{equation}
\label{eq:ideal}
\dataset_\synth^* = \mathrm{argmax}_{\dataset_\synth^{(k)}} \performance \left( f \left( \cdot ; \theta_k \right), \dataset_e \right)
\end{equation}
In practice, Eq~\ref{eq:ideal} is infeasible because it requires labeled real data for evaluation and extensive model training across $K$ synthetic datasets.
Instead, \ournameshort\ seeks \textit{proxy metrics} $\proxy(\dataset_\synth^{(k)}, \samples_\real)$---computable using only the synthetic dataset $\dataset_\synth^{(k)}$ and unannotated real samples $\samples_\real$---that correlate strongly with true downstream performance.
The \ournameshort\ problem thus reduces to designing or learning a proxy function
$\proxy: (\dataset_\synth^{(k)}, \samples_\real) \rightarrow \mathbb{R}$
such that a higher score indicates the synthetic dataset $\dataset_\synth^{(k)}$ is more likely to produce better real-world task performance.
This formulation enables efficient and effective use of synthetic data in low-resource or privacy-sensitive scenarios. For ease of exposition, we refer to the score produced by the proxy function $\proxy$ as the \textbf{\ournameshort\ score}.

\section{\ournameshort\ Proxy Metrics}
\label{sec:proxy_metrics}
In this section, we introduce the proxy metrics designed for \ournameshort. To avoid circular reasoning, we explicitly ground these proxies in machine learning theory, linking them directly to our downstream objective in~\cref{eq:ideal} through two primary indicators: \textbf{distribution matching} and \textbf{diversity}.

We ground our similarity metrics ($\pad$, $\mmd$) in domain adaptation theory \citep{ben-david_analysis_2006}. Theoretical bounds show that downstream target risk is upper-bounded by source training risk plus domain divergence. Therefore, ranking synthetic datasets by their divergence from real data mathematically targets the upper bound of the downstream error defined in~\cref{eq:ideal}.

Similarly, we ground our diversity proxy ($\mdm$) in embedding space coverage. If a synthetic dataset lacks diversity (e.g., due to mode collapse), its empirical support is artificially restricted. Models trained on this limited support fail to generalize to the broader real-world target distribution, directly increasing target risk.

We adapt these traditional distance and diversity measures for \ournameshort\ by embedding raw text or image data into continuous representations. Finally, for complex, long-horizon tasks where representation-based metrics fail, we introduce \ourmethodnameshort. This LLM-based measure operates directly on raw data, providing deep contextual understanding without relying on a universal embedding model.

\subsection{Representation-Based Proxy Metrics}

\paragraph{Mean Distance to Medoid}
Our first proxy metric adapts Mean Distance to Medoid ($\mdm$), a measure of dataset diversity~\citep{cox_directed_2021, laliberte_distance-based_2010, lehman_abandoning_2011, risi_how_2009}.
The rationale is that high-diversity synthetic datasets may offer broader coverage in the embedding space and thus yield higher performance on real data.
For \ournameshort, $\mdm$ characterizes this diversity by measuring the sparsity of data points around medoids.
Given $\dataset_\synth$ of size $n_s$, we compute $K$ medoids using clustering algorithms such as kMedoids \footnote{\url{https://pypi.org/project/kmedoids/}}.
Let $\tilde{x}_k$ be the $k$-th medoid and $\mathcal{C}_k$ be the set of points assigned to this medoid. We aggregate the Euclidean distances from all points within their corresponding cluster:
$
\mdm = \frac{1}{n_s} \sum_{k=1}^K \sum_{x_i \in \mathcal{C}_k} d(x_i, \tilde{x}_k)
$.
Intuitively, if points within each cluster are centered around the medoid, $\mdm$ will be small, indicating the dataset is less diverse in the embedding space.
In contrast, high-diversity synthetic datasets to have broader coverage, and therefore higher performance on real data.
A higher $\mdm$ score suggests greater diversity, so we use it directly as the \ournameshort\ score.

\paragraph{Maximum Mean Discrepancy}
Next, we propose a proxy based on Maximum Mean Discrepancy ($\mmd$), a nonparametric test that assesses whether two samples originate from the same distribution~\citep{gretton_kernel_2012,borgwardt_integrating_2006, lu_improved_2022, li_generative_2015}.
As a \ournameshort\ proxy, we use $\mmd$ to measure the discrepancy between the synthetic dataset and the real data distribution.
Given $n_\synth$ synthetic input samples $x_i\in\dataset_\synth$ of size $n_\synth$ and $m_\real$ unannotated real input samples $y_i\in\samples_\real$, $\mmd$ quantifies the distance between these two empirical distributions in a reproducing kernel Hilbert space using a kernel function $k(\cdot)$.
\begin{equation}
\mmd=\frac{1}{{m_\real}^2}\sum_{i,j=1}^{m_\real} k(x_i,x_j)+\frac{1}{{n_\synth}^2}\sum_{i,j=1}^{n_\synth} k(y_i,y_j)-\frac{2}{m_\real n_\synth}\sum_{i,j=1}^{n_\synth, m_\real}k(x_i,y_j)
\end{equation}
A smaller $\mmd$ score indicates the synthetic data distribution is closer to the real one.
To maintain consistency with other proxies where higher is better, we use $-\mmd$ as the \ournameshort\ score.

\begin{table*}
  \small
  \noindent\begin{minipage}{.37\columnwidth}
    \begin{lstlisting}[label=lst:prompt_rubric,caption=Rubric prompt,frame=tlrb,backgroundcolor=\color{light-gray}]
You are shown samples from
datasets A and B. Give up to
10 points describing how 
dataset B is similar to dataset A.

Samples from dataset A:
${location A}

Samples from dataset B:
${location B}   
    \end{lstlisting}
  \end{minipage}\hfill
  \begin{minipage}{.6\columnwidth}
    \begin{lstlisting}[label=lst:prompt_scorer,caption=Scorer Prompt,frame=tlrb,backgroundcolor=\color{light-gray}]
Given similarities and differences between
datasets, how likely is the given sample from
dataset ${prediction}? Choose from very unlikely,
unlikely, unsure, likely, and very likely.

Similarities: ${similarities}

Differences: ${differences}

Sample: ${sample} 
    \end{lstlisting}
  \end{minipage}
\vspace{-0.1in}
\caption{Simplified \ourmethodnameshort\ prompt templates. Appendix~\ref{app:experiment} contains detailed prompts for each task.}
\label{tab:lens_prompts}
\vspace{-0.1in}
\end{table*}

\paragraph{Proxy-A-Distance}
Our third representation-based proxy adapts Proxy-A-Distance ($\pad$), a discriminative measure from domain adaptation that quantifies the divergence between two distributions~\citep{ben-david_analysis_2006, elsahar_annotate_2019}.
The $\pad$ proxy measures how well a classifier can discriminate between samples from the synthetic dataset $\dataset_\synth$ and real dataset $\dataset_\real$.
We compute $\pad=1-2\mathcal{E}(G)$ by training a binary domain classifier $G:x\rightarrow \left[0,1\right]$ (e.g., a linear SVM, a multi-layer perceptron) to distinguish between synthetic and real inputs (i.e.~we do not assume access to labels), where the error of the classifier $\mathcal{E}$ can be computed as:
\begin{equation}
\mathcal{E}\left(G\right)=1-\frac{1}{n_\synth+m_\real}\sum_{x_i\in \dataset_\synth,\space\samples_\real}\left|G\left(x_i\right)-\mathbb{I}\left(x_i\in \dataset_\synth\right)\right|   
\end{equation}
A higher classification error implies the datasets are not easily separable, indicating lower divergence and thus higher synthetic data quality.
For consistency with other divergence metrics, use $-\pad$ as the \ournameshort\ score.

\paragraph{\mauve}


As an established generative baseline to compare against our representation-based proxies, we evaluate \mauve~\citep{pillutla_mauve_2021}, a widely used metric that quantifies the divergence between \textit{text} distributions.
\mauve\ summarizes both Type I and Type II errors. Given a synthetic data distribution $Q$ and a real data distribution $P$, Type I error means $Q$ generates text that is unlikely under $P$ (unrealistic samples), while Type II error means $Q$ fails to generate text that is plausible under $P$ (lacks diversity).
\mauve\ captures both error types in a single score, which approaches 1 as the distributions become more similar.
Since a higher score indicates better alignment with real data, we utilize the \mauve\ score directly as a baseline dataset ranking metric to compare against our proposed \ournameshort\ proxies.

\subsection{LLM-Evaluated Normalized Score (\ourmethodnameshort)}
The representation-based proxies described so far rely on high-quality continuous representations of inputs.
In low-resource settings where such representations are unavailable, or in long-horizon settings where it is intractable to compress a long sequence of observations and states into a compact, fixed-size representation, these representation-based proxies may fall short.
To address this, we introduce \ourmethodname\ (\ourmethodnameshort), a novel method that leverages LLMs as zero-shot discriminators.
\ourmethodnameshort\ first derives a language \textbf{rubric} describing the similarities and differences between samples of unannotated real data $\samples_\real$ and inputs from the synthetic dataset $\dataset_\synth$.
A subsequent (potentially smaller) LLM then scores how likely each synthetic example is to belong to the real dataset, guided by the rubric.
The average score across the synthetic dataset is used as the final \ournameshort\ score.
The intuition is similar to that behind $\pad$: a higher classification error by the rubric-guided scorer implies higher synthetic data quality.
We now detail how \ourmethodnameshort\ is computed.

\paragraph{Rubric compilation}

Given real input samples $\samples_\real$, we collect an equal number of samples $\samples_\synth$ from the synthetic dataset $\dataset_\synth$.
Both collections are given to a reasoning LLM (e.g.~\texttt{DeepSeek R1} or \texttt{o4-mini}) to generate three sets of \textbf{characteristic descriptions}: commonalities ($\characterization$), differences of real from synthetic ($\characterization_{\real,\synth}$), and differences of synthetic from real ($\characterization_{\synth, \real}$).
Listing~\ref{lst:prompt_rubric} shows a simplified rubric compilation prompt template.
Our design is backed by the principled idea of approximating domain divergence through discriminator error~\citep{ben-david_analysis_2006}: \ourmethodnameshort's scoring is motivated by $\pad$, where the error of a classifier (here, the LLM-based scorer) reflects the distance between distributions.
Unlike $\pad$, however, \ourmethodnameshort\ does not require a pretrained encoder to map samples into fixed-length representations; instead, it operates directly on the native data format and characterizes differences using language rubrics.

\paragraph{Principled Debiasing and Scoring}

We now describe how to compute the score of a synthetic dataset.
A key challenge of scoring is in mitigating LLM biases.
We identified three primary sources:

\begin{enumerate}
    \item \textbf{Order Bias:} The set of differences an LLM derives when comparing A to B can differ significantly from when comparing B to A.
    \item \textbf{Label Bias:} When asked how likely an example $x$ belongs to A or B, an LLM may score both as ``very likely'', a contradiction.
    \item \textbf{Score Bias:} LLMs may have an inherent preference for certain score values (e.g., ``likely'') regardless of the input.
\end{enumerate}

To address these systematically, we employ a \textit{minimal design} involving four scoring permutations for each sample.
Specifically, we evaluate the likelihood of each sample belonging to the target dataset (real or synthetic) under both difference rubrics ($\characterization_{\synth, \real}$ and $\characterization_{\real, \synth}$). This yields four distinct scores \textit{per sample} $x$: $g_{\real \mid \characterization_{\synth,\real}}(x)$, $g_{\synth \mid \characterization_{\synth,\real}}(x)$, $g_{\real \mid \characterization_{\real, \synth}}(x)$, $g_{\synth \mid \characterization_{\real, \synth}}(x)$.

We formalize the LLM scoring function as $g_{\dataset \mid \characterization}:\mathcal{X}\rightarrow \left\{0,1,2,3,4 \right\}$, where $x \in \mathcal{X}$ denotes a sample from the input space. The output score (0-4) indicates how likely the example $x$ belongs to dataset $\dataset$ given characteristic descriptions $\characterization$ in natural language.

First, to mitigate \textbf{score bias}, we compute baseline scores by averaging the LLM's judgments on real inputs $x \in \samples_\real$ for each of the four permutations.
For instance, the baseline for scoring an example as real, given the description of how synthetic differs from real, is:
\begin{equation}
    z_{\real \mid \characterization_{\synth,\real}} = \expect{g_{\real \mid \characterization_{\synth,\real}}(x)} \approx \frac{1}{n_\real} \sum_{i=1}^{n_\real} g_{\real \mid \characterization_{\synth,\real}}(x_i)
\end{equation}
We then compute a \textit{score-debiased} score $h$ for each synthetic sample by normalizing its raw score against this baseline:
\begin{equation}
    h_{\real \mid \characterization_{\synth,\real}} (x) = \frac{g_{\real \mid \characterization_{\synth,\real}}(x)}{\mathrm{max}(\epsilon, z_{\real \mid \characterization_{\synth,\real}})}
\end{equation}
Here, $\epsilon$ is a small constant to avoid division by zero.
Intuitively, this scores-debiased score expresses how much the LLM scores the example compared to how it usually scores real examples. Similarly, the score-debiased score for the synthetic label is computed as:
\begin{equation}
    h_{\synth \mid \characterization_{\synth,\real}} (x) = \frac{g_{\synth \mid \characterization_{\synth,\real}}(x)}{\mathrm{max}(\epsilon, z_{\synth \mid \characterization_{\synth,\real}})}
\end{equation}
where $z_{\synth \mid \characterization_{\synth,\real}} = \frac{1}{n_\real} \sum_{i=1}^{n_\real} g_{\synth \mid \characterization_{\synth,\real}}(x_i)$.
Next, to compute a \textbf{label-debiased} score, we normalize the LLM's preference for the ``real'' label over the ``synthetic'' label for each synthetic example:
\begin{equation}
p_{\real \mid \characterization_{\synth,\real}} (x) = 
    \frac{
        h_{\real \mid \characterization_{\synth,\real}} (x)
    }{
        h_{\real \mid  \characterization_{\synth,\real}} (x)
        +
        h_{\synth \mid \characterization_{\synth,\real}} (x)
        +
        \epsilon
    }
\end{equation}
Finally, to create an \textbf{order-debiased} error score for a synthetic sample $x \in \dataset_\synth$, we average the label-debiased scores indicating a "real" prediction using both sets of difference descriptions ($\characterization_{\synth,\real}$ and $\characterization_{\real,\synth}$):
\begin{equation}
\hat{p}_\synth(x) = \frac{1}{2} \left[ p_{\real \mid \characterization_{\synth,\real}}(x) + p_{\real \mid \characterization_{\real,\synth}}(x) \right]
\end{equation}

Conversely, we compute a symmetrical error score for each real sample $x \in \samples_\real$, representing the likelihood it is misclassified as synthetic. Using corresponding baselines normalized against the synthetic dataset, this is calculated as:
\begin{equation}
\hat{p}_\real(x) = \frac{1}{2} \left[ p_{\synth \mid \characterization_{\synth,\real}}(x) + p_{\synth \mid \characterization_{\real,\synth}}(x) \right]
\end{equation}

The final \ourmethodnameshort\ score is the combined, weighted empirical mean of these classification errors across both datasets, capturing the overall indistinguishability between the real and synthetic samples:
\begin{equation}
\mathrm{\ourmethodnameshort}(\dataset_\synth) = \frac{1}{n_\synth + n_\real} \left( \sum_{i=1}^{n_\synth} \hat{p}_\synth(x_i) + \sum_{j=1}^{n_\real} \hat{p}_\real(x_j) \right)
\end{equation}
\section{Experiments}
\label{sec:experiment}

We choose four diverse tasks spanning different machine learning domains to examine how well each candidate proxy metric extrapolates to real data performance on tasks with varying complexities and modalities.
For \ourmethodnameshort, we incorporate \texttt{Deepseek-R1} to generate $10$ points about similar and different characteristics $\characterization_{\synth,\real}$ between synthetic data samples $\samples_\synth$ and real data samples $\samples_\real$.
We then use \texttt{Qwen2.5-32B-Instruct}~\citep{qwen_qwen25_2025} and \texttt{8B} to score synthetic examples according to the rubric.
For image domain, we use OpenAI \texttt{o4-mini} to compile rubrics and \texttt{Qwen2.5-VL-32B-Instruct} to score.
For representation-based metrics $\pad, \mdm,\text{and }\mmd$, we use state-of-the-art \texttt{qte-Qwen2-7B-Instruct} to embed text inputs and \texttt{E5-V}~\citep{jiang_e5-v_2024} to embed image inputs for proxy scoring.
We use XGBoost ~\citep{chen_xgboost_2016} to compute $\pad$ and  polynomial kernel for $\mmd$ ~\citep{gretton_kernel_2012}.
We include additional kernel ablations in Appendix~\ref{tab:mmd_kernel_ablation}.
We use the official release\footnote{\url{https://pypi.org/project/mauve-text/}} from \mauve, with the default hyperparameter setting for \mauve\ calculation.

We also include an experiment with perplexity-based metric as additional baseline in Text2SQL (see ~\cref{tab:text2sql_perplexity}).
\perplexity, inspired by scaling law methods~\citep{magnusson_datadecide_2025, liu_regmix_2025}, fine-tunes a model on each synthetic data set and measures its perplexity on the unannotated real data subset; a lower perplexity is expected to indicate higher quality.

We use Pearson~\citep{pearson_vii_1997} and Spearman rank~\citep{spearman_proof_1904} correlation coefficients to measure how strongly task performance and proxy scores are related.
Pearson focuses on \textit{predictability} by capturing linear relationships, while Spearman focuses on \textit{trend} by evaluating whether the relationship between variables is consistently increasing or decreasing (i.e., monotonic), regardless of the exact shape.
To reduce variance in correlation analysis across different sample subsets, for all tasks, we construct subsets $\samples_\real$ by sampling with five different seeds.
Final correlation scores are averaged across seeds.

\setlength{\fboxsep}{2pt} 
\begin{table}[t]
\small
\centering
\renewcommand{\arraystretch}{1.2}
\caption{Top-3 task performance for all candidate proxies of \ournameshort. Top-3 task performance is computed by averaging task performance of synthetic datasets chosen using each proxy metric. Improvements are calculated based on increase over average performance of all synthetic datasets.}
\vspace{-0.1 in}
\scalebox{0.9}{
\begin{tabular}{lcccccccccc}
\toprule
&&\multicolumn{8}{c}{\textbf{Top-3 Ranked Average Task Performance across Proxy Metrics}} \\
Tasks & Test & \multicolumn{2}{c}{\ourmethodnameshort\ 7B} & \multicolumn{2}{c}{\ourmethodnameshort\ 32B} & $\pad$ & $\mmd$ & $\mdm$ & \textbf{Mauve} \\
& mean & debiased & biased & debiased & biased & & & & \\ \midrule
\textbf{Sentiment} &
49.6 & \entry{50.5}{+0.8} & \entry{51.2}{+1.6} & \entry{52.0}{+2.4} & \entry{51.0}{+1.4} & \entry{55.3}{+5.7} & \entry{54.7}{+5.1} & \entry{54.2}{+4.6} & \entry{54.6}{+4.9} \\ \midrule

\textbf{Text2SQL} & & & & & & & & & \\
Computer &
45.8 & \entry{46.6}{+0.7} & \entry{46.4}{+0.6} & \entry{48.3}{+2.5} & \entry{46.3}{+0.5} & \entry{48.3}{+2.5} & \entry{47.4}{+1.6} & \entry{48.2}{+2.3} & \entry{48.3}{+2.5} \\

Apps &
30.4 & \entry{34.7}{+4.4} & \entry{36.3}{+5.9} & \entry{33.8}{+3.4} & \entry{35.2}{+4.9} & \entry{33.5}{+3.2} & \entry{38.4}{+8.1} & \entry{33.9}{+3.5} & \entry{38.4}{+8.1} \\

Movies &
37.3 & \entry{41.0}{+3.7} & \entry{41.4}{+4.1} & \entry{43.8}{+6.5} & \entry{44.7}{+7.4} & \entry{44.2}{+6.9} & \entry{43.0}{+5.7} & \entry{46.9}{+9.6} & \entry{44.6}{+7.3} \\

\textit{Average} &
37.8 & \entrya{40.8}{+2.9} & \entrya{41.4}{+3.5} & \entrya{42.0}{+4.1} & \entrya{42.1}{+4.3} & \entrya{42.0}{+4.2} & \entrya{42.9}{+5.1}  & \entrya{43.0}{+5.2} & \entrya{43.8}{+6.0} \\ \midrule

\textbf{Image} & & & & & & & & & \\
Split 1 &
57.2 & \entry{57.3}{+0.2} & \entry{56.7}{-0.4} & \entry{56.4}{-0.8} & \entry{53.4}{-3.7} & \entry{55.9}{-1.3} & \entry{57.3}{+0.1} & \entry{57.0}{-0.1} & \entry{56.0}{-1.1} \\
Split 2 &
55.8 & \entry{55.3}{-0.4} & \entry{55.8}{+0.0} & \entry{56.2}{+0.4} & \entry{56.0}{+0.2} & \entry{55.4}{-0.4} & \entry{54.5}{-1.3} & \entry{54.8}{-1.0} & \entry{56.3}{+0.5} \\
Split 3 &
57.7 & \entry{59.1}{+1.4} & \entry{58.7}{+1.0} & \entry{60.2}{+2.5} & \entry{57.0}{-0.7} & \entry{58.2}{+0.6} & \entry{64.1}{+6.5} & \entry{52.2}{-5.5} & \entry{58.4}{+0.7} \\
\textit{Average} &
56.9 & \entrya{57.2}{+0.4} & \entrya{57.0}{+0.2} & \entrya{57.6}{+0.7} & \entrya{55.5}{-1.4} & \entrya{56.5}{-0.4} & \entrya{58.6}{+1.8} & \entrya{54.7}{-2.2} & \entrya{56.9}{+0} \\
\midrule
\textbf{WebNav} &
25.8 & \entry{26.5}{+0.7} & \entry{26.3}{+0.5} & \entry{26.3}{+0.5} & \entry{26.0}{+0.2} & \entry{25.7}{-0.1} & \entry{26.5}{+0.7} & \entry{25.8}{-0.1} & \entry{26.3}{+0.5} \\ \bottomrule

\end{tabular}
}
\label{tab:top3_all}
\vspace{-0.2in}
\end{table}
\renewcommand{\arraystretch}{1}
\setlength{\fboxsep}{3pt} 
\setlength{\tabcolsep}{6pt} 

\paragraph{Sentiment Analysis}
Recent work shows LLMs overfit widely-used datasets due to data contamination~\citep{balloccu_leak_2024, sainz_nlp_2023, oren_proving_2023}.
To mitigate this, we evaluate on a domain-specific financial tweets sentiment dataset\footnote{\url{https://huggingface.co/datasets/zeroshot/twitter-financial-news-sentiment}}.
We create 32 synthetic class-balanced datasets (998 samples each) using eight prompt types: zero-shot, zero-shot with background knowledge, with train-time or test-time stock ticker info, and few-shot variants. We use \texttt{Qwen2.5-7B-Instruct}, \texttt{Qwen2.5-32B-Instruct}, \texttt{Llama3.1-8B-Instruct}, and \texttt{Llama3.3-70B-Instruct}~\citep{grattafiori_llama_2024} for each prompt type.
Background knowledge uses detailed guideline instructions for better task alignment.
Stock tickers are sampled one at a time for synthesis.
Details are in~\cref{app:experiment}.
We train task models using XGBoost and evaluate F1 score on a 2,388-item test set.
Rubrics are compiled by randomly sampling 200 points from each real and synthetic dataset.

\paragraph{Text2SQL}
We evaluate \ournameshort\ on Text2SQL using three DBs from the BIRD benchmark (Movies, App Store, Computer Students --- we denote the last two as Apps and Computers)~\citep{li_can_2023}.
We synthesize 1,000 data points with 4 prompt types: zero-shot with background knowledge (guidelines and schema), zero-shot with test-time info (random table rows), and few-shot (three examples). \texttt{Qwen2.5-7B-Instruct} and \texttt{Llama3.1-8B-Instruct} models are used for dataset generation.
Task models are finetuned from \texttt{Qwen2.5-Coder-1.5B-Instruct} following CodeS\footnote{\url{https://github.com/RUCKBReasoning/codes}} and evaluated using execution accuracy on the real test set.
For rubrics, we sample 30 points per synthetic and real dataset.
Real data sizes are 60, 69, and 164 for Apps, Computers, and Movies respectively.

\paragraph{Image Classification}
In addition to text-only settings, we evaluate \ournameshort\ on image classification using synthetic datasets curated from \texttt{unmet-promise}\footnote{\url{https://huggingface.co/datasets/scottgeng00/unmet-promise}}.
These datasets are created with different prompts using Stable Diffusion 1.1 and 1.5: label, label plus physical relation, and label plus background description~\citep{geng_unmet_2025}. 
Images are mapped to ImageNet classes~\citep{deng_imagenet_2009} via caption analysis~\ref{app:experiment}, then filtered with a vision-language model to remove noisy labels.
Data cleaning details are provided in Appendix~\cref{app:experiment}.
The final set includes 15 classes with 300 images each.
Due to limited samples, the 15-class task is split into three 5-class tasks (\cref{tab:imagenet-splits}).
We train ResNet-50~\citep{he_deep_2015} from scratch for 50 epochs, early stopping on 10\% validation data.
Evaluation uses mean reciprocal rank (MRR) for finer performance measurement.
Rubrics are constructed from 100 sampled images per real and synthetic data, consistent across \ournameshort\ methods.

\renewcommand{\arraystretch}{0}
\setlength{\fboxsep}{1pt} 
\setlength{\tabcolsep}{0pt}
\begin{table}[t]
\small
\centering
\renewcommand{\arraystretch}{1.2}
\caption{Spearman (left) and Pearson (right) correlation scores of \ournameshort\ proxy metrics. \ourmethodnameshort\ uses a fraction of samples for rubric compilation except for Web Navigation tasks.}
\vspace{-0.1 in}
\begin{tabularx}{\linewidth}{XXXXXXXXX}
\toprule
Tasks &
\multicolumn{2}{c}{\ourmethodnameshort\ 7B} &
\multicolumn{2}{c}{\ourmethodnameshort\ 32B} &
$\pad$ &
$\mmd$ &
$\mdm$ &
Mauve \\
& debiased & biased & debiased & biased & & & & \\ \midrule
\textbf{Sentiment}  &  \entrydualg{ .25}{ .33}  &  \entrydualg{ .26}{ .17}  &  \entrydualg{ .38}{ .26}  &  \entrydualg{ .24}{ .23}  &  \entrydualg{ .53}{ .65}  &  \entrydualg{ .45}{ .67}  &  \entrydualg{ .68}{ .85} & \entrydualg{ .53}{ .57}\\ \midrule
\textbf{Text2SQL} &&&&&&&& \\
Computer & \entrydual{ .19}{ .13}  & \entrydual{ .10}{ .10} & \entrydual{ .41}{ .45} & \entrydual{ .18}{ .23} & \entrydual{ .46}{ .69} & \entrydual{ .33}{ .85} & \entrydual{ .39}{ .63} & \entrydual{ .24}{ .78} \\
Apps  &  \entrydual{ .38}{ .37}  &  \entrydual{ .42}{ .49}  &  \entrydual{ .46}{ .40}  &  \entrydual{ .55}{ .61}  &  \entrydual{ .43}{ .42}  &  \entrydual{ .53}{ .79}  &  \entrydual{ .44}{ .56} & \entrydual{ .74}{ .52}\\
Movies  &  \entrydual{ .41}{ .50}  &  \entrydual{ .41}{ .26}  &  \entrydual{ .50}{ .46}  &  \entrydual{ .56}{ .47}  &  \entrydual{ .50}{ .64}  &  \entrydual{ .38}{ .46}  &  \entrydual{ .61}{ .41} & \entrydual{ .65}{ .68}\\
\textit{Average}  &  \entrydual{ .33}{ .33}  &  \entrydual{ .31}{ .28}  &  \entrydual{ .46}{ .43}  &  \entrydual{ .43}{ .44}  &  \entrydual{ .46}{ .58}  &  \entrydual{ .41}{ .70}  &  \entrydual{ .48}{ .53} & \entrydual{ .55}{ .66} \\ \midrule
\textbf{Image} &&&&&&&& \\
Split 1  &  \entrydual{-.18}{-.19}  &  \entrydual{-.30}{-.35}  &  \entrydual{-.28}{-.28}  &  \entrydual{-.68}{-.67} & \entrydual{-.06}{-.05}  &  \entrydual{ .66}{ .52}  &  \entrydual{-.37}{-.27} & \entrydual{-.04}{-.20} \\
Split 2  &  \entrydual{ .02}{-.04}  &  \entrydual{ .14}{ .05}  &  \entrydual{ .20}{ .05}  &  \entrydual{ .20}{ .05} & \entrydual{ .20}{ .31}  &  \entrydual{ .09}{ .17}  &  \entrydual{ .03}{-.32} & \entrydual{ .03}{ .13}\\
Split 3  &  \entrydual{-.10}{-.01}  &  \entrydual{-.15}{-.03}  &  \entrydual{ .31}{ .33}  &  \entrydual{ .37}{ .44} & \entrydual{ .02}{-.15}  &  \entrydual{ .26}{ .21}  &  \entrydual{-.54}{-.76} & \entrydual{ .46}{ .34}\\
\textit{Average}  &  \entrydual{-.09}{-.08}  &  \entrydual{-.10}{-.11}  &  \entrydual{ .08}{ .03}  &  \entrydual{-.04}{-.06} & \entrydual{ .05}{ .04}  &  \entrydual{ .33}{ .30}  &  \entrydual{-.30}{-.45} & \entrydual{ .15}{ .09}\\
\midrule
\textbf{WebNav}  &  \entrydual{ .15}{ .17}  &  \entrydual{ .11}{ .18}  &  \entrydual{ .15}{ .15}  &  \entrydual{ .08}{ .09}  &  \entrydual{ .11}{ .08}  &  \entrydual{-.02}{ .06}  &  \entrydual{-.11}{-.08} & \entrydual{-.09}{-.10}\\ \bottomrule
\end{tabularx}
\label{tab:correlation_all}
\vspace{-0.2in}
\end{table}

\renewcommand{\arraystretch}{1}
\setlength{\fboxsep}{3pt} 
\setlength{\tabcolsep}{6pt} 

\begin{wraptable}[10]{o}{2.8in}
    \vspace{-1.3em}
    \caption{Division of the 15 selected ImageNet classes into three 5-class splits for image classification tasks. Each row corresponds to one split used in our experiments.}
    \vspace{-0.5em}
    \small
    \begin{tabularx}{\linewidth}{cX}
    \toprule
    Splits & ImageNet classes \\ \midrule
    1  & bra, mask, lion, cloak, tank   \\
    2 & hammer, backpack, stage, throne, tray   \\
    3  & plate, desk, kimono, shield, church \\ \bottomrule
    \end{tabularx}
    \label{tab:imagenet-splits}
\end{wraptable}
\vspace{-0.5em}

\paragraph{Web Navigation}
Our fourth task evaluates \ournameshort\ on agentic web navigation planning using WebVoyager~\citep{he_webvoyager_2024} and synthetic data from NNetNav~\citep{murty_nnetnav_2025}.
Inputs include task objectives, current step observations (accessibility tree), and past actions; targets are actions leading to task success.
WebVoyager has 15 websites; we exclude Google Flights and Booking which are no longer feasible~\citep{zhou_proposer-agent-evaluatorpae_2024,murty_nnetnav_2025}, leaving 13 sites with 557 tasks.
Each site forms a test domain split into 5 synthetic subsets.
Models are fine-tuned with LoRA~\citep{hu_lora_2021} on \texttt{Qwen2.5-7B-Instruct}. We use all synthetic and 20 real samples per method.

\subsection{Results Analysis}

\paragraph{\ournameshort\ proxies correlate with task performance and improve selection.}
\cref{tab:correlation_all} shows that \ournameshort\ proxy metrics demonstrate moderate to strong correlation with downstream task performance.
To show the practical utility of these proxies, we simulate selecting the top 3 datasets based on each metric's score and compare their average task performance against the mean performance of all available synthetic datasets.
As shown in \cref{tab:top3_all}, nearly all proxy metrics significantly improve dataset selection over selecting synthetic datasets non-discriminately (i.e.~uniform selection).
This demonstrates that \ournameshort\ is an effective framework for maximizing real-data performance, despite not having access to labeled real data and only a limited sample of real data.
We also conduct an experiment with \perplexity\ on Text2SQL, to examine the effectiveness as a potential proxy. As shown in~\cref{tab:text2sql_perplexity}, it correlates poorly with even text data, therefore we conclude that scaling methods would not work under our setting, where no annotated data is available for evaluation to build a regression model that predicts the best data mixture.

\begin{wraptable}{r}{0.25\linewidth}
\vspace{-1.3em}
\renewcommand{\arraystretch}{0}
\setlength{\fboxsep}{1pt} 
\setlength{\tabcolsep}{3pt}
\centering
\renewcommand{\arraystretch}{1.2}
\small
\caption{Spearman (left) and Pearson (right) correlations with \perplexity\ scoring on the BIRD Text2SQL benchmark}
\vspace{-0.8em}
\begin{tabularx}{\linewidth}{Xc}
\toprule
&
\textbf{\perplexity} \\ \midrule

Computers         &
\entrydual{-.31}{-.33} \\

Apps                &
\entrydual{-.25}{-.29} \\

Movies           &
\entrydual{ .24}{ .31} \\

\textit{Average}                   &
\entrydual{-.11}{-.10} \\ \bottomrule
\end{tabularx}
\label{tab:text2sql_perplexity}
\vspace{-1em}
\renewcommand{\arraystretch}{1}
\setlength{\fboxsep}{3pt} 
\setlength{\tabcolsep}{6pt} 
\end{wraptable}

\paragraph{Performance on ambiguous image data shows high variance.}
The synthetic image classification data contains significant visual variability and label ambiguity, especially in Split 2 between classes like ``stage'' and ``throne'' (\cref{fig:img_stage}, \cref{fig:img_throne}).
This confuses most proxy metrics, resulting in inconsistent correlations across splits (\cref{tab:correlation_all}).
However, \cref{tab:top3_all} shows that when used for selection, several proxies (e.g. debiased \ourmethodnameshort\ 32B, $\mmd$) still improve average task performance.

\renewcommand{\arraystretch}{1.1}
\setlength{\fboxsep}{1pt} 
\setlength{\tabcolsep}{2pt} 
\begin{table}[t]
\centering
\renewcommand{\arraystretch}{1.2}
\caption{Spearman (left) and Pearson (right) correlations with different number of real samples for scoring Text2SQL. \ourmethodnameshort\ uses debiased 32B scoring.
Note that results in~\cref{tab:correlation_all} use 30 real samples. ``Comp.'' denotes Computers.
}
\vspace{-0.1in}
\resizebox{\linewidth}{!}{%
\begin{tabularx}{1.05\linewidth}{Xccccccccccc}
\toprule
&
\multicolumn{5}{c}{\textbf{$\left|\samples_\real\right|=25$}} &
&
\multicolumn{5}{c}{\textbf{$\left|\samples_\real\right|=50$}} \\ 
&
\textbf{\ourmethodnameshort} &
\textbf{PAD} & \textbf{$\mmd$} & \textbf{$\mdm$} & \textbf{Mauve} &
\hspace*{0.1in} &
\textbf{\ourmethodnameshort} &
\textbf{$\pad$} &
\textbf{$\mmd$} &
\textbf{$\mdm$} &
\textbf{Mauve}
\\ \midrule

Comp.         &
\entrydual{.22}{.22} &
\entrydual{ .36}{ .65} &
\entrydual{.28}{.80} &
\entrydual{ .39}{ .65} &
\entrydual{ .19}{ .65} &
&
\entrydual{.64}{.38} &
\entrydual{.51}{.78} &
\entrydual{.34}{.87} &
\entrydual{ .39}{ .62} & 
\entrydual{ .87}{ .78} \\

Apps                &
\entrydual{.29}{.48} &
\entrydual{-.20}{-.23} &
\entrydual{.40}{.77} &
\entrydual{ .44}{ .56} &
\entrydual{ .65}{ .70} &
&
\entrydual{.64}{.66} &
\entrydual{.68}{.76} &
\entrydual{.57}{.80} &
\entrydual{ .44}{ .56} & 
\entrydual{ .32}{ .65} \\

Movies           &
\entrydual{.33}{.48} &
\entrydual{ .33}{ .36} &
\entrydual{.52}{.57} &
\entrydual{ .60}{ .40} &
\entrydual{-.14}{-.35} &
&
\entrydual{.43}{.49} &
\entrydual{.57}{.37} &
\entrydual{.81}{.56} &
\entrydual{ .67}{ .45} &
\entrydual{ .81}{.60}
\\

\textit{Average}                   &
\entrydual{.28}{.39} &
\entrydual{ .16}{ .26} &
\entrydual{.40}{.71} &
\entrydual{ .47}{ .54} &
\entrydual{ .23}{ .33} &
&
\entrydual{.57}{.51} &
\entrydual{.59}{.64} &
\entrydual{.57}{.74} &
\entrydual{ .50}{ .55} &
\entrydual{ .67}{ .68} \\ \bottomrule
\end{tabularx}
}
\label{tab:text2sql_num_samples_ablation}
\vspace*{-0.2in}
\end{table}
\renewcommand{\arraystretch}{1}
\setlength{\fboxsep}{3pt} 
\setlength{\tabcolsep}{6pt} 

\paragraph{Using more real samples improves correlation.}
As shown in \cref{tab:text2sql_num_samples_ablation}, increasing the number of unannotated real samples $m_\real$ consistently leads to stronger correlations for all proxy metrics.
This indicates that even a modest increase in available real-world data can significantly improve the reliability of synthetic data quality estimation.

\paragraph{\ourmethodnameshort\ excels on complex, long-horizon tasks.}
As shown in both tables, the 32B debiased \ourmethodnameshort\ is the only proxy that consistently achieves positive correlation and improves top-3 task performance across all tasks and splits.
Its advantage is particularly pronounced in web navigation, a complex planning task where representation-based metrics struggle.
\ourmethodnameshort\ leverages LLM reasoning over rich, structured inputs like accessibility trees to generate interpretable rubrics.
For instance, an example characteristic point for the website ``Wolfram Alpha'' is: ``Dataset B tasks focus on data retrieval (e.g.~temperature anomalies, moon phases) while Dataset A emphasizes applied computational problem-solving''.
These specific nuances in long text (e.g.~instructions, state observations) are difficult to capture using general-purpose dense vector embedders, which explains why \ourmethodnameshort\ outperforms representation-based proxy metrics on complex, abstract tasks like web navigation.
This method does exhibit weaker correlation in image classification.
We hypothesize that this stems from the inability of VLMs to capture meaning characteristics descriptions in batches of images during rubric generation, and VLM rubric generation will likely improve as VLMs improve in quality.

\subsection{Ablation Studies and Cost Analysis}

\paragraph{Cost-Effectiveness of \ournameshort\ Proxies}
A key motivation for \ournameshort\ is efficiency.
Our representation-based proxies require a one-time embedding computation, after which scoring all datasets is nearly instantaneous (e.g., 19 seconds for MMD on 32 datasets).
\ourmethodnameshort, using modern LLM serving frameworks, is also highly efficient, taking $\sim15$ seconds per dataset with a 32B model on a H200 GPU.
In contrast, perplexity-based data selection, inspired by scaling-law studies~\citep{liu_regmix_2025}, require training many (e.g., 512 1M models used in their experiment) small models on the mixture of all synthetic datasets, a significantly more costly procedure, yet yield weaker correlations (\cref{tab:correlation_all}).
This highlights the practical advantage of the \ournameshort\ framework.

\paragraph{Larger scorers lead to stronger correlations}
We find that larger scoring models yield stronger correlations between \ourmethodnameshort\ and task performance, as shown in~\cref{tab:correlation_all}.
Intuitively, this is expected because larger models generally possess more robust instruction-following and reasoning capabilities, enabling them to better assess data quality and align proxy scores with downstream performance.

\paragraph{\ourmethodnameshort\ is robust to preferential bias in LLM training data.}
To address concerns that an LLM evaluator might favor data it generated, we tested \ourmethodnameshort\ with different scoring models on data generated by \texttt{Qwen2.5}.
The results, detailed in Appendix \cref{tab:lens_bias_ablation}, show that performance is consistent across evaluators, including those distinct from models used to generate the synthetic data, indicating minimal preferential bias.

\paragraph{Principled debiasing and rubrics are critical for \ourmethodnameshort.}
As illustrated in ~\cref{tab:correlation_all}, the correlations between \ourmethodnameshort\ and task performance consistently increase when debiasing is applied, indicating that raw scores may be systematically biased and do not reliably reflect true data quality.
Once debiasing is introduced, the correlation becomes strongly positive and consistently outperforms the biased scores.
This demonstrates that debiasing effectively corrects for these systematic errors and aligns \ourmethodnameshort\ scores with actual task performance.
Further ablations show that using a rubric consistently improves correlation over a zero-shot baseline (Appendix~\cref{tab:lens_rubric_ablation_existence}), and that 10 rubric points generally offer the best trade-off between specificity and generality (Appendix~\cref{tab:lens_rubric_ablation_size}).

\section{Practitioner Guidelines}
Based on our empirical evaluation including ablation studies across diverse domains and experiment setups, we offer the following guidelines for practitioners utilizing the SYNQUE framework to select synthetic data.

\begin{itemize}
    \item \textbf{Simple, One-shot NLP Tasks (e.g., Sentiment Analysis):}
    Representation-based metrics like $\pad$ and $\mdm$ serve as effective and \textit{computationally efficient} defaults. These metrics require only a one-time embedding computation and demonstrate remarkable robustness, maintaining strong correlations with task performance even with as few as 100 real samples.
    \item \textbf{Complex, Long-horizon Tasks (e.g., Agentic, Large-Scale Datasets):}
    We highly recommend utilizing \ourmethodnameshort. General-purpose dense vector embedders often struggle to capture the specific nuances and sequential logic present in long texts like state observations; \ourmethodnameshort{} overcomes this by leveraging LLM reasoning to generate interpretable rubrics over structured inputs.
    \item \textbf{Vision Tasks:}
    Selecting high-quality data is particularly challenging due to significant visual variability and label ambiguity. In these settings, distribution matching via $\mmd$ provides a solid baseline that can still improve dataset selection, though it should be paired with explicit data cleaning pipelines -- such as utilizing vision-language models to filter out noisy or hallucinated images.
    \item \textbf{Safeguarding Against Generative Issues:}
    To safeguard against common generative issues like mode collapse, $\mdm$ should be utilized to measure dataset diversity. If a synthetic dataset lacks diversity, its empirical support is artificially restricted; therefore, prioritizing higher MDM scores ensures broader coverage in the embedding space and better generalization to the real-world target distribution.
\end{itemize}

\section{Conclusions, Limitations and Future Work}
\label{sec:conclusion_limitation}
We formalized the \ournameshort\  problem of ranking synthetic datasets by their impact on real-world task performance using limited unannotated real data.
Our comprehensive evaluation established that various proxies can reliably predict downstream performance, offering a cost-effective alternative to full model training.
We proposed \ourmethodnameshort, a novel proxy leveraging LLM reasoning and principled debiasing, which consistently outperforms others on challenging, long-horizon tasks.
Overall, \ournameshort\ offers a robust framework for synthetic data selection when labeled real data is scarce.

\ournameshort\ assumes that real data is scarce, a setting not all deployments face.
While \ourmethodnameshort\ performs well on the complex tasks studied, its effectiveness should be validated on more diverse tasks.
Additionally, we experiment with limited-size LLMs due to resource constraints.
Future work should explore 1) scaling \ourmethodnameshort\ to larger sizes and different architectures, especially strong VLMs, to assess generality and improvements.
2) using rubric feedback to guide LLMs in synthesizing more realistic data, and
3) developing fine-grained, example-level proxy use to directly improve task model training.

\section{Broader Impact Statement}
While utilizing synthetic datasets offers a promising solution to data scarcity, it could introduce ethical considerations regarding the potential amplification of biases. It is crucial to clarify that our \ournameshort{} framework does not select datasets by directly optimizing for downstream test performance, as such labels are strictly inaccessible during the selection phase. Instead, proxies such as $\pad$, $\mmd$, and $\mauve$ evaluate synthetic data based strictly on its \textbf{unsupervised distributional similarity} to a small real-world reference set; test performance is utilized solely as a post-hoc oracle to evaluate the success of these proxies. Consequently, the risk of amplifying known biases depends entirely on the composition of this real-world anchor set. If the reference data contains historical or representational biases, similarity-based selection will inherently prioritize synthetic datasets that reflect and perpetuate those same biases. To mitigate this risk, we strongly advise practitioners to rigorously audit their real-world reference samples for representational fairness and equity prior to applying \ournameshort, ensuring that the anchor data aligns with desired ethical standards.


\subsubsection*{Acknowledgments}

We thank the Vector Institute for providing the compute resources that made this work possible. We are grateful to Shai Ben-David for insightful discussions that helped shape the theoretical grounding of this work.

\bibliography{references}

@article{chapman_dataset_2020,
    title = {Dataset search: a survey},
    volume = {29},
    issn = {0949-877X},
    shorttitle = {Dataset search},
    url = {https://doi.org/10.1007/s00778-019-00564-x},
    doi = {10.1007/s00778-019-00564-x},
    language = {en},
    number = {1},
    urldate = {2026-03-27},
    journal = {The VLDB Journal},
    author = {Chapman, Adriane and Simperl, Elena and Koesten, Laura and Konstantinidis, George and Ibáñez, Luis-Daniel and Kacprzak, Emilia and Groth, Paul},
    month = jan,
    year = {2020},
    pages = {251--272},
}

@misc{kaushal_learning_2018,
	title = {Learning {From} {Less} {Data}: {Diversified} {Subset} {Selection} and {Active} {Learning} in {Image} {Classification} {Tasks}},
	shorttitle = {Learning {From} {Less} {Data}},
	url = {http://arxiv.org/abs/1805.11191},
	doi = {10.48550/arXiv.1805.11191},
	urldate = {2026-03-25},
	publisher = {arXiv},
	author = {Kaushal, Vishal and Sahoo, Anurag and Doctor, Khoshrav and Raju, Narasimha and Shetty, Suyash and Singh, Pankaj and Iyer, Rishabh and Ramakrishnan, Ganesh},
	month = may,
	year = {2018},
	note = {arXiv:1805.11191 [cs]},
}

@inproceedings{kirsch_batchbald_2019,
	title = {{BatchBALD}: {Efficient} and {Diverse} {Batch} {Acquisition} for {Deep} {Bayesian} {Active} {Learning}},
	volume = {32},
	shorttitle = {{BatchBALD}},
	url = {https://proceedings.neurips.cc/paper_files/paper/2019/hash/95323660ed2124450caaac2c46b5ed90-Abstract.html},
	urldate = {2026-03-25},
	booktitle = {Advances in {Neural} {Information} {Processing} {Systems}},
	publisher = {Curran Associates, Inc.},
	author = {Kirsch, Andreas and van Amersfoort, Joost and Gal, Yarin},
	year = {2019},
}

@misc{kuo_not_2025,
	title = {Not {All} {LLM}-{Generated} {Data} {Are} {Equal}: {Rethinking} {Data} {Weighting} in {Text} {Classification}},
	shorttitle = {Not {All} {LLM}-{Generated} {Data} {Are} {Equal}},
	url = {http://arxiv.org/abs/2410.21526},
	doi = {10.48550/arXiv.2410.21526},
	urldate = {2025-12-03},
	publisher = {arXiv},
	author = {Kuo, Hsun-Yu and Liao, Yin-Hsiang and Chao, Yu-Chieh and Ma, Wei-Yun and Cheng, Pu-Jen},
	month = mar,
	year = {2025},
	note = {arXiv:2410.21526 [cs]},
}

@misc{zhang_harnessing_2024,
	title = {Harnessing {Diversity} for {Important} {Data} {Selection} in {Pretraining} {Large} {Language} {Models}},
	url = {http://arxiv.org/abs/2409.16986},
	doi = {10.48550/arXiv.2409.16986},
	urldate = {2025-12-03},
	publisher = {arXiv},
	author = {Zhang, Chi and Zhong, Huaping and Zhang, Kuan and Chai, Chengliang and Wang, Rui and Zhuang, Xinlin and Bai, Tianyi and Qiu, Jiantao and Cao, Lei and Fan, Ju and Yuan, Ye and Wang, Guoren and He, Conghui},
	month = oct,
	year = {2024},
	note = {arXiv:2409.16986 [cs]},
}

@misc{hu_most_2025,
	title = {Most {Influential} {Subset} {Selection}: {Challenges}, {Promises}, and {Beyond}},
	shorttitle = {Most {Influential} {Subset} {Selection}},
	url = {http://arxiv.org/abs/2409.18153},
	doi = {10.48550/arXiv.2409.18153},
	urldate = {2025-12-03},
	publisher = {arXiv},
	author = {Hu, Yuzheng and Hu, Pingbang and Zhao, Han and Ma, Jiaqi W.},
	month = jan,
	year = {2025},
	note = {arXiv:2409.18153 [cs]},
}

@article{nadas_synthetic_2025,
	title = {Synthetic {Data} {Generation} {Using} {Large} {Language} {Models}: {Advances} in {Text} and {Code}},
	volume = {13},
	issn = {2169-3536},
	shorttitle = {Synthetic {Data} {Generation} {Using} {Large} {Language} {Models}},
	url = {http://arxiv.org/abs/2503.14023},
	doi = {10.1109/ACCESS.2025.3589503},
	urldate = {2025-10-16},
	journal = {IEEE Access},
	author = {Nadas, Mihai and Diosan, Laura and Tomescu, Andreea},
	year = {2025},
	note = {arXiv:2503.14023 [cs]},
	pages = {134615--134633},
}

@misc{pillutla_mauve_2021,
	title = {{MAUVE}: {Measuring} the {Gap} {Between} {Neural} {Text} and {Human} {Text} using {Divergence} {Frontiers}},
	shorttitle = {{MAUVE}},
	url = {http://arxiv.org/abs/2102.01454},
	doi = {10.48550/arXiv.2102.01454},
	urldate = {2025-09-20},
	publisher = {arXiv},
	author = {Pillutla, Krishna and Swayamdipta, Swabha and Zellers, Rowan and Thickstun, John and Welleck, Sean and Choi, Yejin and Harchaoui, Zaid},
	month = nov,
	year = {2021},
	note = {arXiv:2102.01454 [cs]},
}

@misc{liu_regmix_2025,
	title = {{RegMix}: {Data} {Mixture} as {Regression} for {Language} {Model} {Pre}-training},
	shorttitle = {{RegMix}},
	url = {http://arxiv.org/abs/2407.01492},
	doi = {10.48550/arXiv.2407.01492},
	urldate = {2025-09-20},
	publisher = {arXiv},
	author = {Liu, Qian and Zheng, Xiaosen and Muennighoff, Niklas and Zeng, Guangtao and Dou, Longxu and Pang, Tianyu and Jiang, Jing and Lin, Min},
	month = jan,
	year = {2025},
	note = {arXiv:2407.01492 [cs]},
}

@misc{magnusson_datadecide_2025,
	title = {{DataDecide}: {How} to {Predict} {Best} {Pretraining} {Data} with {Small} {Experiments}},
	shorttitle = {{DataDecide}},
	url = {http://arxiv.org/abs/2504.11393},
	doi = {10.48550/arXiv.2504.11393},
	urldate = {2025-09-20},
	publisher = {arXiv},
	author = {Magnusson, Ian and Tai, Nguyen and Bogin, Ben and Heineman, David and Hwang, Jena D. and Soldaini, Luca and Bhagia, Akshita and Liu, Jiacheng and Groeneveld, Dirk and Tafjord, Oyvind and Smith, Noah A. and Koh, Pang Wei and Dodge, Jesse},
	month = jul,
	year = {2025},
	note = {arXiv:2504.11393 [cs]},
}

@article{li_synthetic_2023,
	title = {Synthetic {Data} {Generation} with {Large} {Language} {Models} for {Text} {Classification}: {Potential} and {Limitations}},
	copyright = {arXiv.org perpetual, non-exclusive license},
	shorttitle = {Synthetic {Data} {Generation} with {Large} {Language} {Models} for {Text} {Classification}},
	url = {https://arxiv.org/abs/2310.07849},
	doi = {10.48550/ARXIV.2310.07849},
	urldate = {2025-05-13},
	publisher = {arXiv},
	author = {Li, Zhuoyan and Zhu, Hangxiao and Lu, Zhuoran and Yin, Ming},
	year = {2023},
	note = {Version Number: 2},
}

@misc{he_deep_2015,
	title = {Deep {Residual} {Learning} for {Image} {Recognition}},
	url = {http://arxiv.org/abs/1512.03385},
	doi = {10.48550/arXiv.1512.03385},
	urldate = {2025-05-15},
	publisher = {arXiv},
	author = {He, Kaiming and Zhang, Xiangyu and Ren, Shaoqing and Sun, Jian},
	month = dec,
	year = {2015},
	note = {arXiv:1512.03385 [cs]},
}

@misc{grattafiori_llama_2024,
	title = {The {Llama} 3 {Herd} of {Models}},
	url = {http://arxiv.org/abs/2407.21783},
	doi = {10.48550/arXiv.2407.21783},
	urldate = {2025-05-15},
	publisher = {arXiv},
	author = {Grattafiori, Aaron and Dubey, Abhimanyu and Jauhri, Abhinav and Pandey, Abhinav and Kadian, Abhishek and Al-Dahle, Ahmad and Letman, Aiesha and Mathur, Akhil and Schelten, Alan and Vaughan, Alex and Yang, Amy and Fan, Angela and Goyal, Anirudh and Hartshorn, Anthony and Yang, Aobo and Mitra, Archi and Sravankumar, Archie and Korenev, Artem and Hinsvark, Arthur and Rao, Arun and Zhang, Aston and Rodriguez, Aurelien and Gregerson, Austen and Spataru, Ava and Roziere, Baptiste and Biron, Bethany and Tang, Binh and Chern, Bobbie and Caucheteux, Charlotte and Nayak, Chaya and Bi, Chloe and Marra, Chris and McConnell, Chris and Keller, Christian and Touret, Christophe and Wu, Chunyang and Wong, Corinne and Ferrer, Cristian Canton and Nikolaidis, Cyrus and Allonsius, Damien and Song, Daniel and Pintz, Danielle and Livshits, Danny and Wyatt, Danny and Esiobu, David and Choudhary, Dhruv and Mahajan, Dhruv and Garcia-Olano, Diego and Perino, Diego and Hupkes, Dieuwke and Lakomkin, Egor and AlBadawy, Ehab and Lobanova, Elina and Dinan, Emily and Smith, Eric Michael and Radenovic, Filip and Guzmán, Francisco and Zhang, Frank and Synnaeve, Gabriel and Lee, Gabrielle and Anderson, Georgia Lewis and Thattai, Govind and Nail, Graeme and Mialon, Gregoire and Pang, Guan and Cucurell, Guillem and Nguyen, Hailey and Korevaar, Hannah and Xu, Hu and Touvron, Hugo and Zarov, Iliyan and Ibarra, Imanol Arrieta and Kloumann, Isabel and Misra, Ishan and Evtimov, Ivan and Zhang, Jack and Copet, Jade and Lee, Jaewon and Geffert, Jan and Vranes, Jana and Park, Jason and Mahadeokar, Jay and Shah, Jeet and Linde, Jelmer van der and Billock, Jennifer and Hong, Jenny and Lee, Jenya and Fu, Jeremy and Chi, Jianfeng and Huang, Jianyu and Liu, Jiawen and Wang, Jie and Yu, Jiecao and Bitton, Joanna and Spisak, Joe and Park, Jongsoo and Rocca, Joseph and Johnstun, Joshua and Saxe, Joshua and Jia, Junteng and Alwala, Kalyan Vasuden and Prasad, Karthik and Upasani, Kartikeya and Plawiak, Kate and Li, Ke and Heafield, Kenneth and Stone, Kevin and El-Arini, Khalid and Iyer, Krithika and Malik, Kshitiz and Chiu, Kuenley and Bhalla, Kunal and Lakhotia, Kushal and Rantala-Yeary, Lauren and Maaten, Laurens van der and Chen, Lawrence and Tan, Liang and Jenkins, Liz and Martin, Louis and Madaan, Lovish and Malo, Lubo and Blecher, Lukas and Landzaat, Lukas and Oliveira, Luke de and Muzzi, Madeline and Pasupuleti, Mahesh and Singh, Mannat and Paluri, Manohar and Kardas, Marcin and Tsimpoukelli, Maria and Oldham, Mathew and Rita, Mathieu and Pavlova, Maya and Kambadur, Melanie and Lewis, Mike and Si, Min and Singh, Mitesh Kumar and Hassan, Mona and Goyal, Naman and Torabi, Narjes and Bashlykov, Nikolay and Bogoychev, Nikolay and Chatterji, Niladri and Zhang, Ning and Duchenne, Olivier and Çelebi, Onur and Alrassy, Patrick and Zhang, Pengchuan and Li, Pengwei and Vasic, Petar and Weng, Peter and Bhargava, Prajjwal and Dubal, Pratik and Krishnan, Praveen and Koura, Punit Singh and Xu, Puxin and He, Qing and Dong, Qingxiao and Srinivasan, Ragavan and Ganapathy, Raj and Calderer, Ramon and Cabral, Ricardo Silveira and Stojnic, Robert and Raileanu, Roberta and Maheswari, Rohan and Girdhar, Rohit and Patel, Rohit and Sauvestre, Romain and Polidoro, Ronnie and Sumbaly, Roshan and Taylor, Ross and Silva, Ruan and Hou, Rui and Wang, Rui and Hosseini, Saghar and Chennabasappa, Sahana and Singh, Sanjay and Bell, Sean and Kim, Seohyun Sonia and Edunov, Sergey and Nie, Shaoliang and Narang, Sharan and Raparthy, Sharath and Shen, Sheng and Wan, Shengye and Bhosale, Shruti and Zhang, Shun and Vandenhende, Simon and Batra, Soumya and Whitman, Spencer and Sootla, Sten and Collot, Stephane and Gururangan, Suchin and Borodinsky, Sydney and Herman, Tamar and Fowler, Tara and Sheasha, Tarek and Georgiou, Thomas and Scialom, Thomas and Speckbacher, Tobias and Mihaylov, Todor and Xiao, Tong and Karn, Ujjwal and Goswami, Vedanuj and Gupta, Vibhor and Ramanathan, Vignesh and Kerkez, Viktor and Gonguet, Vincent and Do, Virginie and Vogeti, Vish and Albiero, Vítor and Petrovic, Vladan and Chu, Weiwei and Xiong, Wenhan and Fu, Wenyin and Meers, Whitney and Martinet, Xavier and Wang, Xiaodong and Wang, Xiaofang and Tan, Xiaoqing Ellen and Xia, Xide and Xie, Xinfeng and Jia, Xuchao and Wang, Xuewei and Goldschlag, Yaelle and Gaur, Yashesh and Babaei, Yasmine and Wen, Yi and Song, Yiwen and Zhang, Yuchen and Li, Yue and Mao, Yuning and Coudert, Zacharie Delpierre and Yan, Zheng and Chen, Zhengxing and Papakipos, Zoe and Singh, Aaditya and Srivastava, Aayushi and Jain, Abha and Kelsey, Adam and Shajnfeld, Adam and Gangidi, Adithya and Victoria, Adolfo and Goldstand, Ahuva and Menon, Ajay and Sharma, Ajay and Boesenberg, Alex and Baevski, Alexei and Feinstein, Allie and Kallet, Amanda and Sangani, Amit and Teo, Amos and Yunus, Anam and Lupu, Andrei and Alvarado, Andres and Caples, Andrew and Gu, Andrew and Ho, Andrew and Poulton, Andrew and Ryan, Andrew and Ramchandani, Ankit and Dong, Annie and Franco, Annie and Goyal, Anuj and Saraf, Aparajita and Chowdhury, Arkabandhu and Gabriel, Ashley and Bharambe, Ashwin and Eisenman, Assaf and Yazdan, Azadeh and James, Beau and Maurer, Ben and Leonhardi, Benjamin and Huang, Bernie and Loyd, Beth and Paola, Beto De and Paranjape, Bhargavi and Liu, Bing and Wu, Bo and Ni, Boyu and Hancock, Braden and Wasti, Bram and Spence, Brandon and Stojkovic, Brani and Gamido, Brian and Montalvo, Britt and Parker, Carl and Burton, Carly and Mejia, Catalina and Liu, Ce and Wang, Changhan and Kim, Changkyu and Zhou, Chao and Hu, Chester and Chu, Ching-Hsiang and Cai, Chris and Tindal, Chris and Feichtenhofer, Christoph and Gao, Cynthia and Civin, Damon and Beaty, Dana and Kreymer, Daniel and Li, Daniel and Adkins, David and Xu, David and Testuggine, Davide and David, Delia and Parikh, Devi and Liskovich, Diana and Foss, Didem and Wang, Dingkang and Le, Duc and Holland, Dustin and Dowling, Edward and Jamil, Eissa and Montgomery, Elaine and Presani, Eleonora and Hahn, Emily and Wood, Emily and Le, Eric-Tuan and Brinkman, Erik and Arcaute, Esteban and Dunbar, Evan and Smothers, Evan and Sun, Fei and Kreuk, Felix and Tian, Feng and Kokkinos, Filippos and Ozgenel, Firat and Caggioni, Francesco and Kanayet, Frank and Seide, Frank and Florez, Gabriela Medina and Schwarz, Gabriella and Badeer, Gada and Swee, Georgia and Halpern, Gil and Herman, Grant and Sizov, Grigory and Guangyi and Zhang and Lakshminarayanan, Guna and Inan, Hakan and Shojanazeri, Hamid and Zou, Han and Wang, Hannah and Zha, Hanwen and Habeeb, Haroun and Rudolph, Harrison and Suk, Helen and Aspegren, Henry and Goldman, Hunter and Zhan, Hongyuan and Damlaj, Ibrahim and Molybog, Igor and Tufanov, Igor and Leontiadis, Ilias and Veliche, Irina-Elena and Gat, Itai and Weissman, Jake and Geboski, James and Kohli, James and Lam, Janice and Asher, Japhet and Gaya, Jean-Baptiste and Marcus, Jeff and Tang, Jeff and Chan, Jennifer and Zhen, Jenny and Reizenstein, Jeremy and Teboul, Jeremy and Zhong, Jessica and Jin, Jian and Yang, Jingyi and Cummings, Joe and Carvill, Jon and Shepard, Jon and McPhie, Jonathan and Torres, Jonathan and Ginsburg, Josh and Wang, Junjie and Wu, Kai and U, Kam Hou and Saxena, Karan and Khandelwal, Kartikay and Zand, Katayoun and Matosich, Kathy and Veeraraghavan, Kaushik and Michelena, Kelly and Li, Keqian and Jagadeesh, Kiran and Huang, Kun and Chawla, Kunal and Huang, Kyle and Chen, Lailin and Garg, Lakshya and A, Lavender and Silva, Leandro and Bell, Lee and Zhang, Lei and Guo, Liangpeng and Yu, Licheng and Moshkovich, Liron and Wehrstedt, Luca and Khabsa, Madian and Avalani, Manav and Bhatt, Manish and Mankus, Martynas and Hasson, Matan and Lennie, Matthew and Reso, Matthias and Groshev, Maxim and Naumov, Maxim and Lathi, Maya and Keneally, Meghan and Liu, Miao and Seltzer, Michael L. and Valko, Michal and Restrepo, Michelle and Patel, Mihir and Vyatskov, Mik and Samvelyan, Mikayel and Clark, Mike and Macey, Mike and Wang, Mike and Hermoso, Miquel Jubert and Metanat, Mo and Rastegari, Mohammad and Bansal, Munish and Santhanam, Nandhini and Parks, Natascha and White, Natasha and Bawa, Navyata and Singhal, Nayan and Egebo, Nick and Usunier, Nicolas and Mehta, Nikhil and Laptev, Nikolay Pavlovich and Dong, Ning and Cheng, Norman and Chernoguz, Oleg and Hart, Olivia and Salpekar, Omkar and Kalinli, Ozlem and Kent, Parkin and Parekh, Parth and Saab, Paul and Balaji, Pavan and Rittner, Pedro and Bontrager, Philip and Roux, Pierre and Dollar, Piotr and Zvyagina, Polina and Ratanchandani, Prashant and Yuvraj, Pritish and Liang, Qian and Alao, Rachad and Rodriguez, Rachel and Ayub, Rafi and Murthy, Raghotham and Nayani, Raghu and Mitra, Rahul and Parthasarathy, Rangaprabhu and Li, Raymond and Hogan, Rebekkah and Battey, Robin and Wang, Rocky and Howes, Russ and Rinott, Ruty and Mehta, Sachin and Siby, Sachin and Bondu, Sai Jayesh and Datta, Samyak and Chugh, Sara and Hunt, Sara and Dhillon, Sargun and Sidorov, Sasha and Pan, Satadru and Mahajan, Saurabh and Verma, Saurabh and Yamamoto, Seiji and Ramaswamy, Sharadh and Lindsay, Shaun and Lindsay, Shaun and Feng, Sheng and Lin, Shenghao and Zha, Shengxin Cindy and Patil, Shishir and Shankar, Shiva and Zhang, Shuqiang and Zhang, Shuqiang and Wang, Sinong and Agarwal, Sneha and Sajuyigbe, Soji and Chintala, Soumith and Max, Stephanie and Chen, Stephen and Kehoe, Steve and Satterfield, Steve and Govindaprasad, Sudarshan and Gupta, Sumit and Deng, Summer and Cho, Sungmin and Virk, Sunny and Subramanian, Suraj and Choudhury, Sy and Goldman, Sydney and Remez, Tal and Glaser, Tamar and Best, Tamara and Koehler, Thilo and Robinson, Thomas and Li, Tianhe and Zhang, Tianjun and Matthews, Tim and Chou, Timothy and Shaked, Tzook and Vontimitta, Varun and Ajayi, Victoria and Montanez, Victoria and Mohan, Vijai and Kumar, Vinay Satish and Mangla, Vishal and Ionescu, Vlad and Poenaru, Vlad and Mihailescu, Vlad Tiberiu and Ivanov, Vladimir and Li, Wei and Wang, Wenchen and Jiang, Wenwen and Bouaziz, Wes and Constable, Will and Tang, Xiaocheng and Wu, Xiaojian and Wang, Xiaolan and Wu, Xilun and Gao, Xinbo and Kleinman, Yaniv and Chen, Yanjun and Hu, Ye and Jia, Ye and Qi, Ye and Li, Yenda and Zhang, Yilin and Zhang, Ying and Adi, Yossi and Nam, Youngjin and Yu and Wang and Zhao, Yu and Hao, Yuchen and Qian, Yundi and Li, Yunlu and He, Yuzi and Rait, Zach and DeVito, Zachary and Rosnbrick, Zef and Wen, Zhaoduo and Yang, Zhenyu and Zhao, Zhiwei and Ma, Zhiyu},
	month = nov,
	year = {2024},
	note = {arXiv:2407.21783 [cs]},
}

@misc{jiang_e5-v_2024,
	title = {E5-{V}: {Universal} {Embeddings} with {Multimodal} {Large} {Language} {Models}},
	shorttitle = {E5-{V}},
	url = {http://arxiv.org/abs/2407.12580},
	doi = {10.48550/arXiv.2407.12580},
	urldate = {2025-05-15},
	publisher = {arXiv},
	author = {Jiang, Ting and Song, Minghui and Zhang, Zihan and Huang, Haizhen and Deng, Weiwei and Sun, Feng and Zhang, Qi and Wang, Deqing and Zhuang, Fuzhen},
	month = jul,
	year = {2024},
	note = {arXiv:2407.12580 [cs]},
}

@misc{touvron_llama_2023,
	title = {{LLaMA}: {Open} and {Efficient} {Foundation} {Language} {Models}},
	shorttitle = {{LLaMA}},
	url = {http://arxiv.org/abs/2302.13971},
	doi = {10.48550/arXiv.2302.13971},
	urldate = {2025-05-15},
	publisher = {arXiv},
	author = {Touvron, Hugo and Lavril, Thibaut and Izacard, Gautier and Martinet, Xavier and Lachaux, Marie-Anne and Lacroix, Timothée and Rozière, Baptiste and Goyal, Naman and Hambro, Eric and Azhar, Faisal and Rodriguez, Aurelien and Joulin, Armand and Grave, Edouard and Lample, Guillaume},
	month = feb,
	year = {2023},
	note = {arXiv:2302.13971 [cs]},
}

@misc{qwen_qwen25_2025,
	title = {Qwen2.5 {Technical} {Report}},
	url = {http://arxiv.org/abs/2412.15115},
	doi = {10.48550/arXiv.2412.15115},
	urldate = {2025-05-15},
	publisher = {arXiv},
	author = {Qwen and Yang, An and Yang, Baosong and Zhang, Beichen and Hui, Binyuan and Zheng, Bo and Yu, Bowen and Li, Chengyuan and Liu, Dayiheng and Huang, Fei and Wei, Haoran and Lin, Huan and Yang, Jian and Tu, Jianhong and Zhang, Jianwei and Yang, Jianxin and Yang, Jiaxi and Zhou, Jingren and Lin, Junyang and Dang, Kai and Lu, Keming and Bao, Keqin and Yang, Kexin and Yu, Le and Li, Mei and Xue, Mingfeng and Zhang, Pei and Zhu, Qin and Men, Rui and Lin, Runji and Li, Tianhao and Tang, Tianyi and Xia, Tingyu and Ren, Xingzhang and Ren, Xuancheng and Fan, Yang and Su, Yang and Zhang, Yichang and Wan, Yu and Liu, Yuqiong and Cui, Zeyu and Zhang, Zhenru and Qiu, Zihan},
	month = jan,
	year = {2025},
	note = {arXiv:2412.15115 [cs]},
}

@inproceedings{schubert_faster_2019,
	address = {Cham},
	title = {Faster k-{Medoids} {Clustering}: {Improving} the {PAM}, {CLARA}, and {CLARANS} {Algorithms}},
	isbn = {978-3-030-32047-8},
	shorttitle = {Faster k-{Medoids} {Clustering}},
	doi = {10.1007/978-3-030-32047-8_16},
	language = {en},
	booktitle = {Similarity {Search} and {Applications}},
	publisher = {Springer International Publishing},
	author = {Schubert, Erich and Rousseeuw, Peter J.},
	editor = {Amato, Giuseppe and Gennaro, Claudio and Oria, Vincent and Radovanović, Miloš},
	year = {2019},
	pages = {171--187},
}

@article{pearson_vii_1997,
	title = {{VII}. {Note} on regression and inheritance in the case of two parents},
	volume = {58},
	url = {https://royalsocietypublishing.org/doi/abs/10.1098/rspl.1895.0041},
	doi = {10.1098/rspl.1895.0041},
	number = {347-352},
	urldate = {2025-05-15},
	journal = {Proceedings of the Royal Society of London},
	publisher = {Royal Society},
	author = {Pearson, Karl and Galton, Francis},
	month = jan,
	year = {1997},
	pages = {240--242},
}

@article{spearman_proof_1904,
	title = {The {Proof} and {Measurement} of {Association} between {Two} {Things}},
	volume = {15},
	issn = {00029556},
	url = {https://www.jstor.org/stable/1412159?origin=crossref},
	doi = {10.2307/1412159},
	language = {en},
	number = {1},
	urldate = {2025-05-15},
	journal = {The American Journal of Psychology},
	author = {Spearman, C.},
	month = jan,
	year = {1904},
	pages = {72},
}

@inproceedings{li_generative_2015,
	title = {Generative {Moment} {Matching} {Networks}},
	url = {https://www.semanticscholar.org/paper/2904a9932f4cd0f0886121dc1f2d4aaac0455176},
	urldate = {2025-05-14},
	author = {Li, Yujia and Swersky, Kevin and Zemel, R.},
	month = feb,
	year = {2015},
}

@article{lu_improved_2022,
	title = {An {Improved} {Algorithm} of {Drift} {Compensation} for {Olfactory} {Sensors}},
	volume = {12},
	copyright = {https://creativecommons.org/licenses/by/4.0/},
	issn = {2076-3417},
	url = {https://www.mdpi.com/2076-3417/12/19/9529},
	doi = {10.3390/app12199529},
	language = {en},
	number = {19},
	urldate = {2025-05-14},
	journal = {Applied Sciences},
	author = {Lu, Siyu and Guo, Jialiang and Liu, Shan and Yang, Bo and Liu, Mingzhe and Yin, Lirong and Zheng, Wenfeng},
	month = sep,
	year = {2022},
	pages = {9529},
}

@article{borgwardt_integrating_2006,
	title = {Integrating structured biological data by {Kernel} {Maximum} {Mean} {Discrepancy}},
	volume = {22},
	issn = {1367-4811, 1367-4803},
	url = {https://academic.oup.com/bioinformatics/article/22/14/e49/228383},
	doi = {10.1093/bioinformatics/btl242},
	language = {en},
	number = {14},
	urldate = {2025-05-14},
	journal = {Bioinformatics},
	author = {Borgwardt, Karsten M. and Gretton, Arthur and Rasch, Malte J. and Kriegel, Hans-Peter and Schölkopf, Bernhard and Smola, Alex J.},
	month = jul,
	year = {2006},
	pages = {e49--e57},
}

@inproceedings{moller_parrot_2023,
	title = {The {Parrot} {Dilemma}: {Human}-{Labeled} vs. {LLM}-augmented {Data} in {Classification} {Tasks}},
	shorttitle = {The {Parrot} {Dilemma}},
	url = {https://www.semanticscholar.org/paper/The-Parrot-Dilemma%3A-Human-Labeled-vs.-LLM-augmented-M%C3%B8ller-Dalsgaard/73051b7b25ef972c15ea8e7a221f4361991facbe},
	urldate = {2025-05-14},
	author = {Møller, Anders Giovanni and Dalsgaard, Jacob Aarup and Pera, Arianna and Aiello, L.},
	month = apr,
	year = {2023},
}

@article{setlur_rl_2024,
	title = {{RL} on {Incorrect} {Synthetic} {Data} {Scales} the {Efficiency} of {LLM} {Math} {Reasoning} by {Eight}-{Fold}},
	copyright = {Creative Commons Attribution 4.0 International},
	url = {https://arxiv.org/abs/2406.14532},
	doi = {10.48550/ARXIV.2406.14532},
	urldate = {2025-05-14},
	publisher = {arXiv},
	author = {Setlur, Amrith and Garg, Saurabh and Geng, Xinyang and Garg, Naman and Smith, Virginia and Kumar, Aviral},
	year = {2024},
	note = {Version Number: 1},
}

@inproceedings{gao_self-guided_2022,
	title = {Self-{Guided} {Noise}-{Free} {Data} {Generation} for {Efficient} {Zero}-{Shot} {Learning}},
	url = {https://www.semanticscholar.org/paper/Self-Guided-Noise-Free-Data-Generation-for-Learning-Gao-Pi/6f7e03e4ccd26c762090e25dc5d2eb1e1f8c641d},
	urldate = {2025-05-14},
	author = {Gao, Jiahui and Pi, Renjie and Lin, Yong and Xu, Hang and Ye, Jiacheng and Wu, Zhiyong and Zhang, Weizhong and Liang, Xiaodan and Li, Zhenguo and Kong, Lingpeng},
	month = may,
	year = {2022},
}

@article{briesch_large_2023,
	title = {Large {Language} {Models} {Suffer} {From} {Their} {Own} {Output}: {An} {Analysis} of the {Self}-{Consuming} {Training} {Loop}},
	copyright = {arXiv.org perpetual, non-exclusive license},
	shorttitle = {Large {Language} {Models} {Suffer} {From} {Their} {Own} {Output}},
	url = {https://arxiv.org/abs/2311.16822},
	doi = {10.48550/ARXIV.2311.16822},
	urldate = {2025-05-13},
	publisher = {arXiv},
	author = {Briesch, Martin and Sobania, Dominik and Rothlauf, Franz},
	year = {2023},
	note = {Version Number: 2},
}

@article{hataya_will_2023,
	title = {Will {Large}-scale {Generative} {Models} {Corrupt} {Future} {Datasets}?},
	copyright = {https://doi.org/10.15223/policy-029},
	url = {https://ieeexplore.ieee.org/document/10376575/},
	doi = {10.1109/ICCV51070.2023.01879},
	urldate = {2025-05-13},
	journal = {2023 IEEE/CVF International Conference on Computer Vision (ICCV)},
	publisher = {IEEE},
	author = {Hataya, Ryuichiro and Bao, Han and Arai, Hiromi},
	month = oct,
	year = {2023},
	note = {Conference Name: 2023 IEEE/CVF International Conference on Computer Vision (ICCV)
ISBN: 9798350307184
Place: Paris, France},
	pages = {20498--20508},
}

@article{casco-rodriguez_self-consuming_2023,
	title = {Self-{Consuming} {Generative} {Models} go {MAD}},
	url = {https://research.latinxinai.org/papers/neurips/2023/pdf/Josue_CascoRodriguez.pdf},
	doi = {10.52591/lxai202312101},
	urldate = {2025-05-13},
	journal = {LatinX in AI at Neural Information Processing Systems Conference 2023},
	publisher = {Journal of LatinX in AI Research},
	author = {Casco-Rodriguez, Josue and Alemohammad, Sina and Luzi, Lorenzo and Imtiaz, Ahmed and Babaei, Hossein and LeJeune, Daniel and Siahkoohi, Ali and Baraniuk, Richard},
	month = dec,
	year = {2023},
	note = {Conference Name: LatinX in AI at Neural Information Processing Systems Conference 2023},
}

@inproceedings{gunjal_detecting_2023,
	title = {Detecting and {Preventing} {Hallucinations} in {Large} {Vision} {Language} {Models}},
	copyright = {Creative Commons Attribution 4.0 International},
	url = {https://arxiv.org/abs/2308.06394},
	doi = {10.48550/ARXIV.2308.06394},
	urldate = {2025-05-13},
	publisher = {arXiv},
	author = {Gunjal, Anisha and Yin, Jihan and Bas, Erhan},
	year = {2023},
	note = {Version Number: 3},
}

@article{huang_survey_2025,
	title = {A {Survey} on {Hallucination} in {Large} {Language} {Models}: {Principles}, {Taxonomy}, {Challenges}, and {Open} {Questions}},
	volume = {43},
	issn = {1046-8188, 1558-2868},
	shorttitle = {A {Survey} on {Hallucination} in {Large} {Language} {Models}},
	url = {https://dl.acm.org/doi/10.1145/3703155},
	doi = {10.1145/3703155},
	language = {en},
	number = {2},
	urldate = {2025-05-13},
	journal = {ACM Transactions on Information Systems},
	author = {Huang, Lei and Yu, Weijiang and Ma, Weitao and Zhong, Weihong and Feng, Zhangyin and Wang, Haotian and Chen, Qianglong and Peng, Weihua and Feng, Xiaocheng and Qin, Bing and Liu, Ting},
	month = mar,
	year = {2025},
	pages = {1--55},
}

@inproceedings{li_evaluating_2023,
	title = {Evaluating {Object} {Hallucination} in {Large} {Vision}-{Language} {Models}},
	copyright = {arXiv.org perpetual, non-exclusive license},
	url = {https://arxiv.org/abs/2305.10355},
	doi = {10.48550/ARXIV.2305.10355},
	urldate = {2025-05-13},
	publisher = {arXiv},
	author = {Li, Yifan and Du, Yifan and Zhou, Kun and Wang, Jinpeng and Zhao, Wayne Xin and Wen, Ji-Rong},
	year = {2023},
	note = {Version Number: 3},
}

@misc{xie_osworld_2024,
	title = {{OSWorld}: {Benchmarking} {Multimodal} {Agents} for {Open}-{Ended} {Tasks} in {Real} {Computer} {Environments}},
	shorttitle = {{OSWorld}},
	url = {http://arxiv.org/abs/2404.07972},
	doi = {10.48550/arXiv.2404.07972},
	urldate = {2025-05-13},
	publisher = {arXiv},
	author = {Xie, Tianbao and Zhang, Danyang and Chen, Jixuan and Li, Xiaochuan and Zhao, Siheng and Cao, Ruisheng and Hua, Toh Jing and Cheng, Zhoujun and Shin, Dongchan and Lei, Fangyu and Liu, Yitao and Xu, Yiheng and Zhou, Shuyan and Savarese, Silvio and Xiong, Caiming and Zhong, Victor and Yu, Tao},
	month = may,
	year = {2024},
	note = {arXiv:2404.07972 [cs]},
}

@article{ben-david_theory_2010,
	title = {A theory of learning from different domains},
	volume = {79},
	issn = {1573-0565},
	url = {https://doi.org/10.1007/s10994-009-5152-4},
	doi = {10.1007/s10994-009-5152-4},
	language = {en},
	number = {1},
	urldate = {2025-05-13},
	journal = {Machine Learning},
	author = {Ben-David, Shai and Blitzer, John and Crammer, Koby and Kulesza, Alex and Pereira, Fernando and Vaughan, Jennifer Wortman},
	month = may,
	year = {2010},
	pages = {151--175},
}

@book{quinonero-candela_dataset_2022,
	title = {Dataset {Shift} in {Machine} {Learning}},
	isbn = {978-0-262-54587-7},
	language = {en},
	publisher = {MIT Press},
	author = {Quinonero-Candela, Joaquin and Sugiyama, Masashi and Schwaighofer, Anton and Lawrence, Neil D.},
	month = jun,
	year = {2022},
	note = {Google-Books-ID: MBZuEAAAQBAJ},
}

@article{shumailov_ai_2024,
	title = {{AI} models collapse when trained on recursively generated data},
	volume = {631},
	issn = {0028-0836, 1476-4687},
	url = {https://www.nature.com/articles/s41586-024-07566-y},
	doi = {10.1038/s41586-024-07566-y},
	language = {en},
	number = {8022},
	urldate = {2025-05-13},
	journal = {Nature},
	author = {Shumailov, Ilia and Shumaylov, Zakhar and Zhao, Yiren and Papernot, Nicolas and Anderson, Ross and Gal, Yarin},
	month = jul,
	year = {2024},
	pages = {755--759},
}

@article{yu_large_2023,
	title = {Large {Language} {Model} as {Attributed} {Training} {Data} {Generator}: {A} {Tale} of {Diversity} and {Bias}},
	copyright = {arXiv.org perpetual, non-exclusive license},
	shorttitle = {Large {Language} {Model} as {Attributed} {Training} {Data} {Generator}},
	url = {https://arxiv.org/abs/2306.15895},
	doi = {10.48550/ARXIV.2306.15895},
	urldate = {2025-05-13},
	publisher = {arXiv},
	author = {Yu, Yue and Zhuang, Yuchen and Zhang, Jieyu and Meng, Yu and Ratner, Alexander and Krishna, Ranjay and Shen, Jiaming and Zhang, Chao},
	year = {2023},
	note = {Version Number: 2},
}

@inproceedings{mishra_cross-task_2022,
	address = {Dublin, Ireland},
	title = {Cross-{Task} {Generalization} via {Natural} {Language} {Crowdsourcing} {Instructions}},
	url = {https://aclanthology.org/2022.acl-long.244},
	doi = {10.18653/v1/2022.acl-long.244},
	language = {en},
	urldate = {2025-05-13},
	booktitle = {Proceedings of the 60th {Annual} {Meeting} of the {Association} for {Computational} {Linguistics} ({Volume} 1: {Long} {Papers})},
	publisher = {Association for Computational Linguistics},
	author = {Mishra, Swaroop and Khashabi, Daniel and Baral, Chitta and Hajishirzi, Hannaneh},
	year = {2022},
	pages = {3470--3487},
}

@misc{ouyang_training_2022,
	title = {Training language models to follow instructions with human feedback},
	url = {http://arxiv.org/abs/2203.02155},
	doi = {10.48550/arXiv.2203.02155},
	urldate = {2025-02-24},
	publisher = {arXiv},
	author = {Ouyang, Long and Wu, Jeff and Jiang, Xu and Almeida, Diogo and Wainwright, Carroll L. and Mishkin, Pamela and Zhang, Chong and Agarwal, Sandhini and Slama, Katarina and Ray, Alex and Schulman, John and Hilton, Jacob and Kelton, Fraser and Miller, Luke and Simens, Maddie and Askell, Amanda and Welinder, Peter and Christiano, Paul and Leike, Jan and Lowe, Ryan},
	month = mar,
	year = {2022},
	note = {arXiv:2203.02155 [cs]},
}

@misc{honovich_unnatural_2022,
	title = {Unnatural {Instructions}: {Tuning} {Language} {Models} with ({Almost}) {No} {Human} {Labor}},
	shorttitle = {Unnatural {Instructions}},
	url = {http://arxiv.org/abs/2212.09689},
	doi = {10.48550/arXiv.2212.09689},
	urldate = {2025-05-13},
	publisher = {arXiv},
	author = {Honovich, Or and Scialom, Thomas and Levy, Omer and Schick, Timo},
	month = dec,
	year = {2022},
	note = {arXiv:2212.09689 [cs]},
}

@inproceedings{deng_imagenet_2009,
	title = {{ImageNet}: {A} large-scale hierarchical image database},
	issn = {1063-6919},
	shorttitle = {{ImageNet}},
	url = {https://ieeexplore.ieee.org/document/5206848},
	doi = {10.1109/CVPR.2009.5206848},
	urldate = {2025-05-13},
	booktitle = {2009 {IEEE} {Conference} on {Computer} {Vision} and {Pattern} {Recognition}},
	author = {Deng, Jia and Dong, Wei and Socher, Richard and Li, Li-Jia and Li, Kai and Fei-Fei, Li},
	month = jun,
	year = {2009},
	pages = {248--255},
}

@article{balloccu_leak_2024,
	title = {Leak, {Cheat}, {Repeat}: {Data} {Contamination} and {Evaluation} {Malpractices} in {Closed}-{Source} {LLMs}},
	copyright = {Creative Commons Attribution 4.0 International},
	shorttitle = {Leak, {Cheat}, {Repeat}},
	url = {https://arxiv.org/abs/2402.03927},
	doi = {10.48550/ARXIV.2402.03927},
	urldate = {2025-05-13},
	publisher = {arXiv},
	author = {Balloccu, Simone and Schmidtová, Patrícia and Lango, Mateusz and Dušek, Ondřej},
	year = {2024},
	note = {Version Number: 2},
}

@inproceedings{sainz_nlp_2023,
	title = {{NLP} {Evaluation} in trouble: {On} the {Need} to {Measure} {LLM} {Data} {Contamination} for each {Benchmark}},
	copyright = {Creative Commons Attribution Share Alike 4.0 International},
	shorttitle = {{NLP} {Evaluation} in trouble},
	url = {https://arxiv.org/abs/2310.18018},
	doi = {10.48550/ARXIV.2310.18018},
	urldate = {2025-05-13},
	publisher = {arXiv},
	author = {Sainz, Oscar and Campos, Jon Ander and García-Ferrero, Iker and Etxaniz, Julen and de Lacalle, Oier Lopez and Agirre, Eneko},
	year = {2023},
	note = {Version Number: 1},
}

@article{oren_proving_2023,
	title = {Proving {Test} {Set} {Contamination} in {Black} {Box} {Language} {Models}},
	copyright = {Creative Commons Attribution 4.0 International},
	url = {https://arxiv.org/abs/2310.17623},
	doi = {10.48550/ARXIV.2310.17623},
	urldate = {2025-05-13},
	publisher = {arXiv},
	author = {Oren, Yonatan and Meister, Nicole and Chatterji, Niladri and Ladhak, Faisal and Hashimoto, Tatsunori B.},
	year = {2023},
	note = {Version Number: 2},
}

@misc{yang_synthesizing_2024,
	title = {Synthesizing {Text}-to-{SQL} {Data} from {Weak} and {Strong} {LLMs}},
	url = {http://arxiv.org/abs/2408.03256},
	doi = {10.48550/arXiv.2408.03256},
	urldate = {2024-12-24},
	publisher = {arXiv},
	author = {Yang, Jiaxi and Hui, Binyuan and Yang, Min and Yang, Jian and Lin, Junyang and Zhou, Chang},
	month = aug,
	year = {2024},
	note = {arXiv:2408.03256 [cs]},
}

@misc{liu_apigen_2024,
	title = {{APIGen}: {Automated} {Pipeline} for {Generating} {Verifiable} and {Diverse} {Function}-{Calling} {Datasets}},
	shorttitle = {{APIGen}},
	url = {http://arxiv.org/abs/2406.18518},
	doi = {10.48550/arXiv.2406.18518},
	urldate = {2024-11-26},
	publisher = {arXiv},
	author = {Liu, Zuxin and Hoang, Thai and Zhang, Jianguo and Zhu, Ming and Lan, Tian and Kokane, Shirley and Tan, Juntao and Yao, Weiran and Liu, Zhiwei and Feng, Yihao and Murthy, Rithesh and Yang, Liangwei and Savarese, Silvio and Niebles, Juan Carlos and Wang, Huan and Heinecke, Shelby and Xiong, Caiming},
	month = jun,
	year = {2024},
	note = {arXiv:2406.18518},
}

@misc{he_webvoyager_2024,
	title = {{WebVoyager}: {Building} an {End}-to-{End} {Web} {Agent} with {Large} {Multimodal} {Models}},
	shorttitle = {{WebVoyager}},
	url = {http://arxiv.org/abs/2401.13919},
	doi = {10.48550/arXiv.2401.13919},
	urldate = {2025-03-21},
	publisher = {arXiv},
	author = {He, Hongliang and Yao, Wenlin and Ma, Kaixin and Yu, Wenhao and Dai, Yong and Zhang, Hongming and Lan, Zhenzhong and Yu, Dong},
	month = jun,
	year = {2024},
	note = {arXiv:2401.13919 [cs]},
}

@misc{kwon_efficient_2023,
	title = {Efficient {Memory} {Management} for {Large} {Language} {Model} {Serving} with {PagedAttention}},
	url = {http://arxiv.org/abs/2309.06180},
	doi = {10.48550/arXiv.2309.06180},
	urldate = {2025-03-01},
	publisher = {arXiv},
	author = {Kwon, Woosuk and Li, Zhuohan and Zhuang, Siyuan and Sheng, Ying and Zheng, Lianmin and Yu, Cody Hao and Gonzalez, Joseph E. and Zhang, Hao and Stoica, Ion},
	month = sep,
	year = {2023},
	note = {arXiv:2309.06180 [cs]},
}

@misc{hu_lora_2021,
	title = {{LoRA}: {Low}-{Rank} {Adaptation} of {Large} {Language} {Models}},
	shorttitle = {{LoRA}},
	url = {http://arxiv.org/abs/2106.09685},
	doi = {10.48550/arXiv.2106.09685},
	urldate = {2024-12-10},
	publisher = {arXiv},
	author = {Hu, Edward J. and Shen, Yelong and Wallis, Phillip and Allen-Zhu, Zeyuan and Li, Yuanzhi and Wang, Shean and Wang, Lu and Chen, Weizhu},
	month = oct,
	year = {2021},
	note = {arXiv:2106.09685 [cs]},
}

@inproceedings{chen_xgboost_2016,
	title = {{XGBoost}: {A} {Scalable} {Tree} {Boosting} {System}},
	shorttitle = {{XGBoost}},
	url = {http://arxiv.org/abs/1603.02754},
	doi = {10.1145/2939672.2939785},
	urldate = {2024-12-08},
	booktitle = {Proceedings of the 22nd {ACM} {SIGKDD} {International} {Conference} on {Knowledge} {Discovery} and {Data} {Mining}},
	author = {Chen, Tianqi and Guestrin, Carlos},
	month = aug,
	year = {2016},
	note = {arXiv:1603.02754 [cs]},
	pages = {785--794},
}

@misc{zhou_proposer-agent-evaluatorpae_2024,
	title = {Proposer-{Agent}-{Evaluator}({PAE}): {Autonomous} {Skill} {Discovery} {For} {Foundation} {Model} {Internet} {Agents}},
	shorttitle = {Proposer-{Agent}-{Evaluator}({PAE})},
	url = {http://arxiv.org/abs/2412.13194},
	doi = {10.48550/arXiv.2412.13194},
	urldate = {2025-05-12},
	publisher = {arXiv},
	author = {Zhou, Yifei and Yang, Qianlan and Lin, Kaixiang and Bai, Min and Zhou, Xiong and Wang, Yu-Xiong and Levine, Sergey and Li, Erran},
	month = dec,
	year = {2024},
	note = {arXiv:2412.13194 [cs]},
}

@misc{li_can_2023,
	title = {Can {LLM} {Already} {Serve} as {A} {Database} {Interface}? {A} {BIg} {Bench} for {Large}-{Scale} {Database} {Grounded} {Text}-to-{SQLs}},
	shorttitle = {Can {LLM} {Already} {Serve} as {A} {Database} {Interface}?},
	url = {http://arxiv.org/abs/2305.03111},
	doi = {10.48550/arXiv.2305.03111},
	urldate = {2025-05-12},
	publisher = {arXiv},
	author = {Li, Jinyang and Hui, Binyuan and Qu, Ge and Yang, Jiaxi and Li, Binhua and Li, Bowen and Wang, Bailin and Qin, Bowen and Cao, Rongyu and Geng, Ruiying and Huo, Nan and Zhou, Xuanhe and Ma, Chenhao and Li, Guoliang and Chang, Kevin C. C. and Huang, Fei and Cheng, Reynold and Li, Yongbin},
	month = nov,
	year = {2023},
	note = {arXiv:2305.03111 [cs]},
}

@inproceedings{risi_how_2009,
	address = {New York, NY, USA},
	series = {{GECCO} '09},
	title = {How novelty search escapes the deceptive trap of learning to learn},
	isbn = {978-1-60558-325-9},
	url = {https://dl.acm.org/doi/10.1145/1569901.1569923},
	doi = {10.1145/1569901.1569923},
	urldate = {2025-05-10},
	booktitle = {Proceedings of the 11th {Annual} conference on {Genetic} and evolutionary computation},
	publisher = {Association for Computing Machinery},
	author = {Risi, Sebastian and Vanderbleek, Sandy D. and Hughes, Charles E. and Stanley, Kenneth O.},
	month = jul,
	year = {2009},
	pages = {153--160},
}

@article{lehman_abandoning_2011,
	title = {Abandoning {Objectives}: {Evolution} {Through} the {Search} for {Novelty} {Alone}},
	volume = {19},
	issn = {1063-6560},
	shorttitle = {Abandoning {Objectives}},
	url = {https://ieeexplore.ieee.org/abstract/document/6793380},
	doi = {10.1162/EVCO_a_00025},
	number = {2},
	urldate = {2025-05-10},
	journal = {Evolutionary Computation},
	author = {Lehman, Joel and Stanley, Kenneth O.},
	month = jun,
	year = {2011},
	pages = {189--223},
}

@article{gretton_kernel_2012,
	title = {A {Kernel} {Two}-{Sample} {Test}},
	volume = {13},
	issn = {1533-7928},
	url = {http://jmlr.org/papers/v13/gretton12a.html},
	number = {25},
	urldate = {2025-05-10},
	journal = {Journal of Machine Learning Research},
	author = {Gretton, Arthur and Borgwardt, Karsten M. and Rasch, Malte J. and Schölkopf, Bernhard and Smola, Alexander},
	year = {2012},
	pages = {723--773},
}

@article{laliberte_distance-based_2010,
	title = {A distance-based framework for measuring functional diversity from multiple traits},
	volume = {91},
	copyright = {© 2010 by the Ecological Society of America},
	issn = {1939-9170},
	url = {https://onlinelibrary.wiley.com/doi/abs/10.1890/08-2244.1},
	doi = {10.1890/08-2244.1},
	language = {en},
	number = {1},
	urldate = {2025-05-10},
	journal = {Ecology},
	author = {Laliberté, Etienne and Legendre, Pierre},
	year = {2010},
	note = {\_eprint: https://onlinelibrary.wiley.com/doi/pdf/10.1890/08-2244.1},
	pages = {299--305},
}

@inproceedings{cox_directed_2021,
	address = {Yokohama Japan},
	title = {Directed {Diversity}: {Leveraging} {Language} {Embedding} {Distances} for {Collective} {Creativity} in {Crowd} {Ideation}},
	isbn = {978-1-4503-8096-6},
	shorttitle = {Directed {Diversity}},
	url = {https://dl.acm.org/doi/10.1145/3411764.3445782},
	doi = {10.1145/3411764.3445782},
	language = {en},
	urldate = {2025-05-09},
	booktitle = {Proceedings of the 2021 {CHI} {Conference} on {Human} {Factors} in {Computing} {Systems}},
	publisher = {ACM},
	author = {Cox, Samuel Rhys and Wang, Yunlong and Abdul, Ashraf and Von Der Weth, Christian and Y. Lim, Brian},
	month = may,
	year = {2021},
	pages = {1--35},
}

@misc{zheng_judging_2023,
	title = {Judging {LLM}-as-a-{Judge} with {MT}-{Bench} and {Chatbot} {Arena}},
	url = {http://arxiv.org/abs/2306.05685},
	doi = {10.48550/arXiv.2306.05685},
	urldate = {2025-04-14},
	publisher = {arXiv},
	author = {Zheng, Lianmin and Chiang, Wei-Lin and Sheng, Ying and Zhuang, Siyuan and Wu, Zhanghao and Zhuang, Yonghao and Lin, Zi and Li, Zhuohan and Li, Dacheng and Xing, Eric P. and Zhang, Hao and Gonzalez, Joseph E. and Stoica, Ion},
	month = dec,
	year = {2023},
	note = {arXiv:2306.05685 [cs]},
}

@misc{gu_survey_2025,
	title = {A {Survey} on {LLM}-as-a-{Judge}},
	url = {http://arxiv.org/abs/2411.15594},
	doi = {10.48550/arXiv.2411.15594},
	urldate = {2025-04-14},
	publisher = {arXiv},
	author = {Gu, Jiawei and Jiang, Xuhui and Shi, Zhichao and Tan, Hexiang and Zhai, Xuehao and Xu, Chengjin and Li, Wei and Shen, Yinghan and Ma, Shengjie and Liu, Honghao and Wang, Saizhuo and Zhang, Kun and Wang, Yuanzhuo and Gao, Wen and Ni, Lionel and Guo, Jian},
	month = mar,
	year = {2025},
	note = {arXiv:2411.15594 [cs]},
}

@inproceedings{li_bigdatasetgan_2022,
	address = {New Orleans, LA, USA},
	title = {{BigDatasetGAN}: {Synthesizing} {ImageNet} with {Pixel}-wise {Annotations}},
	copyright = {https://doi.org/10.15223/policy-029},
	isbn = {978-1-6654-6946-3},
	shorttitle = {{BigDatasetGAN}},
	url = {https://ieeexplore.ieee.org/document/9878775/},
	doi = {10.1109/CVPR52688.2022.02064},
	language = {en},
	urldate = {2025-04-10},
	booktitle = {2022 {IEEE}/{CVF} {Conference} on {Computer} {Vision} and {Pattern} {Recognition} ({CVPR})},
	publisher = {IEEE},
	author = {Li, Daiqing and Ling, Huan and Kim, Seung Wook and Kreis, Karsten and Fidler, Sanja and Torralba, Antonio},
	month = jun,
	year = {2022},
	pages = {21298--21308},
}

@misc{geng_unmet_2025,
	title = {The {Unmet} {Promise} of {Synthetic} {Training} {Images}: {Using} {Retrieved} {Real} {Images} {Performs} {Better}},
	shorttitle = {The {Unmet} {Promise} of {Synthetic} {Training} {Images}},
	url = {http://arxiv.org/abs/2406.05184},
	doi = {10.48550/arXiv.2406.05184},
	urldate = {2025-04-08},
	publisher = {arXiv},
	author = {Geng, Scott and Hsieh, Cheng-Yu and Ramanujan, Vivek and Wallingford, Matthew and Li, Chun-Liang and Koh, Pang Wei and Krishna, Ranjay},
	month = jan,
	year = {2025},
	note = {arXiv:2406.05184 [cs]},
}

@misc{murty_nnetnav_2025,
	title = {{NNetNav}: {Unsupervised} {Learning} of {Browser} {Agents} {Through} {Environment} {Interaction} in the {Wild}},
	shorttitle = {{NNetNav}},
	url = {http://arxiv.org/abs/2410.02907},
	doi = {10.48550/arXiv.2410.02907},
	urldate = {2025-03-13},
	publisher = {arXiv},
	author = {Murty, Shikhar and Zhu, Hao and Bahdanau, Dzmitry and Manning, Christopher D.},
	month = feb,
	year = {2025},
	note = {arXiv:2410.02907 [cs]},
}

@misc{krumdick_no_2025,
	title = {No {Free} {Labels}: {Limitations} of {LLM}-as-a-{Judge} {Without} {Human} {Grounding}},
	shorttitle = {No {Free} {Labels}},
	url = {http://arxiv.org/abs/2503.05061},
	doi = {10.48550/arXiv.2503.05061},
	urldate = {2025-03-10},
	publisher = {arXiv},
	author = {Krumdick, Michael and Lovering, Charles and Reddy, Varshini and Ebner, Seth and Tanner, Chris},
	month = mar,
	year = {2025},
	note = {arXiv:2503.05061 [cs]},
}

@inproceedings{goodfellow_generative_2014,
	title = {Generative {Adversarial} {Nets}},
	volume = {27},
	url = {https://proceedings.neurips.cc/paper_files/paper/2014/hash/5ca3e9b122f61f8f06494c97b1afccf3-Abstract.html},
	urldate = {2025-02-24},
	booktitle = {Advances in {Neural} {Information} {Processing} {Systems}},
	publisher = {Curran Associates, Inc.},
	author = {Goodfellow, Ian and Pouget-Abadie, Jean and Mirza, Mehdi and Xu, Bing and Warde-Farley, David and Ozair, Sherjil and Courville, Aaron and Bengio, Yoshua},
	year = {2014},
}

@misc{durall_combating_2020,
	title = {Combating {Mode} {Collapse} in {GAN} training: {An} {Empirical} {Analysis} using {Hessian} {Eigenvalues}},
	shorttitle = {Combating {Mode} {Collapse} in {GAN} training},
	url = {http://arxiv.org/abs/2012.09673},
	doi = {10.48550/arXiv.2012.09673},
	urldate = {2025-02-24},
	publisher = {arXiv},
	author = {Durall, Ricard and Chatzimichailidis, Avraam and Labus, Peter and Keuper, Janis},
	month = dec,
	year = {2020},
	note = {arXiv:2012.09673 [cs]},
}

@article{jordan_machine_2015,
	title = {Machine learning: {Trends}, perspectives, and prospects},
	volume = {349},
	shorttitle = {Machine learning},
	url = {https://www.science.org/doi/10.1126/science.aaa8415},
	doi = {10.1126/science.aaa8415},
	number = {6245},
	urldate = {2025-02-24},
	journal = {Science},
	publisher = {American Association for the Advancement of Science},
	author = {Jordan, M. I. and Mitchell, T. M.},
	month = jul,
	year = {2015},
	pages = {255--260},
}

@misc{zhu_measuring_2025,
	title = {Measuring {Diversity} in {Synthetic} {Datasets}},
	url = {http://arxiv.org/abs/2502.08512},
	doi = {10.48550/arXiv.2502.08512},
	urldate = {2025-02-14},
	publisher = {arXiv},
	author = {Zhu, Yuchang and Zhang, Huizhe and Wu, Bingzhe and Li, Jintang and Zheng, Zibin and Zhao, Peilin and Chen, Liang and Bian, Yatao},
	month = feb,
	year = {2025},
	note = {arXiv:2502.08512 [cs]},
}

@misc{long_llms-driven_2024,
	title = {On {LLMs}-{Driven} {Synthetic} {Data} {Generation}, {Curation}, and {Evaluation}: {A} {Survey}},
	shorttitle = {On {LLMs}-{Driven} {Synthetic} {Data} {Generation}, {Curation}, and {Evaluation}},
	url = {http://arxiv.org/abs/2406.15126},
	doi = {10.48550/arXiv.2406.15126},
	urldate = {2025-02-12},
	publisher = {arXiv},
	author = {Long, Lin and Wang, Rui and Xiao, Ruixuan and Zhao, Junbo and Ding, Xiao and Chen, Gang and Wang, Haobo},
	month = jun,
	year = {2024},
	note = {arXiv:2406.15126 [cs]},
}

@inproceedings{elsahar_annotate_2019,
	address = {Hong Kong, China},
	title = {To {Annotate} or {Not}? {Predicting} {Performance} {Drop} under {Domain} {Shift}},
	shorttitle = {To {Annotate} or {Not}?},
	url = {https://aclanthology.org/D19-1222/},
	doi = {10.18653/v1/D19-1222},
	urldate = {2025-01-17},
	booktitle = {Proceedings of the 2019 {Conference} on {Empirical} {Methods} in {Natural} {Language} {Processing} and the 9th {International} {Joint} {Conference} on {Natural} {Language} {Processing} ({EMNLP}-{IJCNLP})},
	publisher = {Association for Computational Linguistics},
	author = {Elsahar, Hady and Gallé, Matthias},
	editor = {Inui, Kentaro and Jiang, Jing and Ng, Vincent and Wan, Xiaojun},
	month = nov,
	year = {2019},
	pages = {2163--2173},
}

@inproceedings{ben-david_analysis_2006,
	title = {Analysis of {Representations} for {Domain} {Adaptation}},
	volume = {19},
	url = {https://proceedings.neurips.cc/paper_files/paper/2006/hash/b1b0432ceafb0ce714426e9114852ac7-Abstract.html},
	urldate = {2025-01-17},
	booktitle = {Advances in {Neural} {Information} {Processing} {Systems}},
	publisher = {MIT Press},
	author = {Ben-David, Shai and Blitzer, John and Crammer, Koby and Pereira, Fernando},
	year = {2006},
}

@misc{hu_agentgen_2024,
	title = {{AgentGen}: {Enhancing} {Planning} {Abilities} for {Large} {Language} {Model} based {Agent} via {Environment} and {Task} {Generation}},
	shorttitle = {{AgentGen}},
	url = {http://arxiv.org/abs/2408.00764},
	doi = {10.48550/arXiv.2408.00764},
	urldate = {2025-01-09},
	publisher = {arXiv},
	author = {Hu, Mengkang and Zhao, Pu and Xu, Can and Sun, Qingfeng and Lou, Jianguang and Lin, Qingwei and Luo, Ping and Rajmohan, Saravan},
	month = nov,
	year = {2024},
	note = {arXiv:2408.00764 [cs]},
}

@misc{sun_os-genesis_2024,
	title = {{OS}-{Genesis}: {Automating} {GUI} {Agent} {Trajectory} {Construction} via {Reverse} {Task} {Synthesis}},
	shorttitle = {{OS}-{Genesis}},
	url = {http://arxiv.org/abs/2412.19723},
	doi = {10.48550/arXiv.2412.19723},
	urldate = {2025-01-01},
	publisher = {arXiv},
	author = {Sun, Qiushi and Cheng, Kanzhi and Ding, Zichen and Jin, Chuanyang and Wang, Yian and Xu, Fangzhi and Wu, Zhenyu and Jia, Chengyou and Chen, Liheng and Liu, Zhoumianze and Kao, Ben and Li, Guohao and He, Junxian and Qiao, Yu and Wu, Zhiyong},
	month = dec,
	year = {2024},
	note = {arXiv:2412.19723 [cs]},
}

@misc{lei_spider_2024,
	title = {Spider 2.0: {Evaluating} {Language} {Models} on {Real}-{World} {Enterprise} {Text}-to-{SQL} {Workflows}},
	shorttitle = {Spider 2.0},
	url = {http://arxiv.org/abs/2411.07763},
	doi = {10.48550/arXiv.2411.07763},
	urldate = {2024-12-03},
	publisher = {arXiv},
	author = {Lei, Fangyu and Chen, Jixuan and Ye, Yuxiao and Cao, Ruisheng and Shin, Dongchan and Su, Hongjin and Suo, Zhaoqing and Gao, Hongcheng and Hu, Wenjing and Yin, Pengcheng and Zhong, Victor and Xiong, Caiming and Sun, Ruoxi and Liu, Qian and Wang, Sida and Yu, Tao},
	month = nov,
	year = {2024},
	note = {arXiv:2411.07763 [cs]},
}

@misc{xu_agenttrek_2024,
	title = {{AgentTrek}: {Agent} {Trajectory} {Synthesis} via {Guiding} {Replay} with {Web} {Tutorials}},
	shorttitle = {{AgentTrek}},
	url = {http://arxiv.org/abs/2412.09605},
	doi = {10.48550/arXiv.2412.09605},
	urldate = {2024-12-13},
	publisher = {arXiv},
	author = {Xu, Yiheng and Lu, Dunjie and Shen, Zhennan and Wang, Junli and Wang, Zekun and Mao, Yuchen and Xiong, Caiming and Yu, Tao},
	month = dec,
	year = {2024},
	note = {arXiv:2412.09605 [cs]},
}

@misc{tan_large_2024,
	title = {Large {Language} {Models} for {Data} {Annotation} and {Synthesis}: {A} {Survey}},
	shorttitle = {Large {Language} {Models} for {Data} {Annotation} and {Synthesis}},
	url = {http://arxiv.org/abs/2402.13446},
	doi = {10.48550/arXiv.2402.13446},
	urldate = {2024-12-08},
	publisher = {arXiv},
	author = {Tan, Zhen and Li, Dawei and Wang, Song and Beigi, Alimohammad and Jiang, Bohan and Bhattacharjee, Amrita and Karami, Mansooreh and Li, Jundong and Cheng, Lu and Liu, Huan},
	month = dec,
	year = {2024},
	note = {arXiv:2402.13446 [cs]},
}

@inproceedings{ye_zerogen_2022,
	address = {Abu Dhabi, United Arab Emirates},
	title = {{ZeroGen}: {Efficient} {Zero}-shot {Learning} via {Dataset} {Generation}},
	shorttitle = {{ZeroGen}},
	url = {https://aclanthology.org/2022.emnlp-main.801},
	doi = {10.18653/v1/2022.emnlp-main.801},
	urldate = {2024-11-30},
	booktitle = {Proceedings of the 2022 {Conference} on {Empirical} {Methods} in {Natural} {Language} {Processing}},
	publisher = {Association for Computational Linguistics},
	author = {Ye, Jiacheng and Gao, Jiahui and Li, Qintong and Xu, Hang and Feng, Jiangtao and Wu, Zhiyong and Yu, Tao and Kong, Lingpeng},
	editor = {Goldberg, Yoav and Kozareva, Zornitsa and Zhang, Yue},
	month = dec,
	year = {2022},
	pages = {11653--11669},
}

@misc{li_codes_2024,
	title = {{CodeS}: {Towards} {Building} {Open}-source {Language} {Models} for {Text}-to-{SQL}},
	shorttitle = {{CodeS}},
	url = {http://arxiv.org/abs/2402.16347},
	doi = {10.48550/arXiv.2402.16347},
	urldate = {2024-11-29},
	publisher = {arXiv},
	author = {Li, Haoyang and Zhang, Jing and Liu, Hanbing and Fan, Ju and Zhang, Xiaokang and Zhu, Jun and Wei, Renjie and Pan, Hongyan and Li, Cuiping and Chen, Hong},
	month = feb,
	year = {2024},
	note = {arXiv:2402.16347},
}
\bibliographystyle{tmlr}
\clearpage

\appendix
\section{Visualization of ambiguous image}
\begin{figure}[hb]
\centering
\begin{subfigure}[a]{0.96\textwidth}
   \includegraphics[width=1\linewidth]{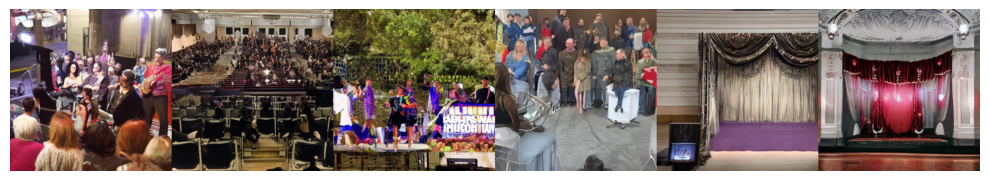}
   \caption{Sample synthetic images of class "stage"}
   \label{fig:img_stage} 
\end{subfigure}
\begin{subfigure}[b]{0.96\textwidth}
   \includegraphics[width=1\linewidth]{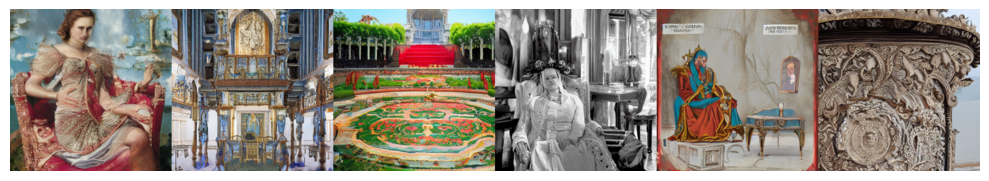}
   \caption{Sample synthetic images of class "throne"}
   \label{fig:img_throne}
\end{subfigure}
\caption[Image Visualization]{Visualization of synthetic images from second split for classes (a) "stage" and (b) "throne"}
\vspace{-0.1in}
\end{figure}

\section{Additional Ablations}

\subsection{Ablation Study on \ourmethodnameshort\ Preferential Bias}

\begin{table}[h]
\centering
\renewcommand{\arraystretch}{1.2}
\caption{
    We use two non-Qwen scoring models on tasks where synthetic data was generated by \texttt{Qwen2.5} models.
    The consistent positive correlations of the debiased scores demonstrate that \ourmethodnameshort\ is robust to this potential bias.
}
\vspace{0.1 in}
\begin{tabular}{lcccc}
\toprule
\textbf{LENS (\texttt{Granite-8B})} & \multicolumn{2}{c}{Debiased} & \multicolumn{2}{c}{Biased} \\
Task & Spearman & Pearson & Spearman & Pearson \\ \midrule
Sentiment & .21 & .30 & -.04 & -.02 \\
Text2SQL & .22 & .20 & .19 & .23 \\ \midrule
\textbf{LENS (\texttt{Ministral-8B})} & \multicolumn{2}{c}{Debiased} & \multicolumn{2}{c}{Biased} \\
Task & Spearman & Pearson & Spearman & Pearson \\
\midrule Sentiment & .28 & .39 & .52 & .31 \\ Text2SQL & .32 & .29 & .16 & .16 \\
\bottomrule
\end{tabular}
\label{tab:lens_bias_ablation}
\end{table}

\subsection{\ourmethodnameshort{} with Different Backbone Models and Sizes}
As shown in~\cref{fig:more_models}, \ourmethodnameshort{} demonstrates robust ranking capabilities across a diverse range of backbone models and sizes. However, the impact of the debiasing step varies significantly by task complexity and model capability:\begin{itemize}\item \textbf{Text2SQL (Complex Generation):} Debiasing consistently improves or maintains correlation strength across nearly all models. For instance, with \texttt{Ministral-8B}, debiasing doubles the Spearman correlation (.16 $\rightarrow$ .32), and it yields the highest overall performance with \texttt{Qwen2.5-32B} (.46). This suggests that for complex structural tasks, removing distributional priors remains beneficial even for capable models.\item \textbf{Sentiment Analysis (Simple Classification):} The results are more nuanced. For models with severe initial bias, such as \texttt{Granite-8B-Instruct}, debiasing is critical, correcting a negative correlation (-.04) to a positive one (.21). However, for highly capable or larger models (e.g., \texttt{GPT-4.1}, \texttt{Granite-4.0 (32B)}, \texttt{Gemma3-27B}), debiasing tends to degrade performance. We hypothesize that these stronger models are already well-calibrated for simple classification tasks; consequently, our analytical debiasing term may introduce unnecessary noise that overrides the model's accurate intrinsic priors.\end{itemize}

\begin{table}[h]
\centering
\renewcommand{\arraystretch}{1.2}
\caption{Comparison of original and debiased \ourmethodnameshort{} scores for ranking synthetic datasets across various backbone models and sizes. For GPT-4.1, sentiment analysis results are reported only due to cost constraints.}
\begin{tabular}{lcccc}
\hline
\textbf{LENS (Qwen2.5-7B-Instruct)}       & \multicolumn{2}{c}{Debiased} & \multicolumn{2}{c}{Original} \\
Task                                      & Spearman      & Pearson      & Spearman      & Pearson      \\ \hline
Sentiment                                 & .25           & .33          & .26           & .17          \\
Text2SQL                                  & .33           & .33          & .31           & .28          \\ \hline
\textbf{LENS (Qwen2.5-32B-Instruct)}      & \multicolumn{2}{c}{Debiased} & \multicolumn{2}{c}{Original} \\
Task                                      & Spearman      & Pearson      & Spearman      & Pearson      \\ \hline
Sentiment                                 & .38           & .26          & .24           & .23          \\
Text2SQL                                  & .46           & .43          & .43           & .44          \\ \hline
\textbf{LENS (Ministral-8B-Instruct)}     & \multicolumn{2}{c}{Debiased} & \multicolumn{2}{c}{Original} \\
Task                                      & Spearman      & Pearson      & Spearman      & Pearson      \\ \hline
Sentiment                                 & .28           & .39          & .52           & .31          \\
Text2SQL                                  & .32           & .29          & .16           & .16          \\ \hline
\textbf{LENS (Granite-8B-Instruct)}       & \multicolumn{2}{c}{Debiased} & \multicolumn{2}{c}{Original} \\
Task                                      & Spearman      & Pearson      & Spearman      & Pearson      \\ \hline
Sentiment                                 & .21           & .30          & -.04          & -.02         \\
Text2SQL                                  & .22           & .20          & .19           & .23          \\ \hline
\textbf{LENS (Granite-4.0-h-small) (32B)} & \multicolumn{2}{c}{Debiased} & \multicolumn{2}{c}{Original} \\
Task                                      & Spearman      & Pearson      & Spearman      & Pearson      \\ \hline
Sentiment                                 & .16           & .39          & .50           & .44          \\
Text2SQL                                  & .32           & .36          & .40           & .48          \\ \hline
\textbf{LENS (Gemma3-12B-Instruct)}       & \multicolumn{2}{c}{Debiased} & \multicolumn{2}{c}{Original} \\
Task                                      & Spearman      & Pearson      & Spearman      & Pearson      \\ \hline
Sentiment                                 & .10           & .13          & .30           & .18          \\
Text2SQL                                  & .30           & .34          & .25           & .25          \\ \hline
\textbf{LENS (Gemma3-27B-Instruct)}       & \multicolumn{2}{c}{Debiased} & \multicolumn{2}{c}{Original} \\
Task                                      & Spearman      & Pearson      & Spearman      & Pearson      \\ \hline
Sentiment                                 & .17           & .33          & .37           & .31          \\
Text2SQL                                  & .31           & .38          & .31           & .27          \\ \hline
\textbf{LENS (GPT-4.1-2025-04-14)}        & \multicolumn{2}{c}{Debiased} & \multicolumn{2}{c}{Original} \\
Task                                      & Spearman      & Pearson      & Spearman      & Pearson      \\ \hline
Sentiment                                 & .07           & .30          & .28           & .37          \\ \hline
\end{tabular}
\label{fig:more_models}
\end{table}

\subsection{Ablation Study on \ourmethodnameshort\ Rubric Design}

\begin{table}[h]
\small
\centering
\renewcommand{\arraystretch}{1.2}
\caption{
     LENS with vs. without a rubric on sentiment analysis, showing the rubric's effectiveness.
     We use \texttt{Qwen2.5-7B-Instruct} for scoring.
}
\vspace{0.1 in}
\begin{tabular}{lccc}
\toprule
\textbf{LENS Model} & \textbf{w/ Rubric} & \textbf{Spearman} & \textbf{Pearson} \\ \midrule
Qwen2.5-7B-Instruct & NO & 0.23 & 0.21 \\
Qwen2.5-32B-Instruct & NO & 0.32 & 0.13 \\
Qwen2.5-7B-Instruct & YES & 0.25 & 0.33 \\
Qwen2.5-32B-Instruct & YES & 0.38 & 0.26 \\
\bottomrule
\end{tabular}
\label{tab:lens_rubric_ablation_existence}
\end{table}

\begin{table}[h]
\small
\centering
\renewcommand{\arraystretch}{1.2}
\caption{
    Bottom: Varying the number of rubric points, where 10 points provides a good balance.
}
\vspace{0.1 in}
\begin{tabular}{lcccc}
\toprule
\textbf{\# Rubric Points} & \multicolumn{2}{c}{\textbf{Sentiment Analysis}} & \multicolumn{2}{c}{\textbf{Text2SQL (Avg)}} \\
(Debiased) & Spearman & Pearson & Spearman & Pearson \\ \midrule
5 & 0.14 & 0.19 & 0.38 & 0.30 \\
10 & 0.25 & 0.33 & 0.33 & 0.33 \\ 
15 & -0.06 & 0.05 & 0.23 & 0.21 \\
\bottomrule
\end{tabular}
\label{tab:lens_rubric_ablation_size}
\end{table}

Our ablation results from~\cref{tab:lens_rubric_ablation_existence} demonstrate the effectiveness of the rubric component of \ourmethodnameshort{}. Table~\ref{tab:lens_rubric_ablation_size} demonstrates that generating too few or too many points will degrade the performance of \ourmethodnameshort{}. 
\subsection{Ablation Study on $\mmd$ Kernel Functions}

\begin{table}[H]
\centering
\renewcommand{\arraystretch}{1.2}
\caption{
Pearson / Spearman correlation coefficients of $\mmd$ with different kernel functions.
}
\vspace{0.1 in}
\begin{tabular}{lccccc}
\toprule
\textbf{Task} & \textbf{Polynomial} & \textbf{RBF} & \textbf{Laplacian} & \textbf{Linear} & \textbf{Sigmoid} \\
\midrule Sentiment & .67 / .45 & .67 / .45 & \textbf{.78} / \textbf{.60} & .67 / .45 & .67 / .46 \\
Text2SQL (avg) & .51 / .43 & .52 / .44 & \textbf{.70} / \textbf{.51} & .52 / .44 & .53 / .44 \\
Image (avg) & .22 / .30 & .22 / .30 & .21 / .30 & .22 / .30 & .22 / .30 \\
WebNav & .06 / -.02 & .06 / -.01 & .05 / -.04 & .06 / -.02 & .05 / -.03 \\
\bottomrule
\end{tabular}
\label{tab:mmd_kernel_ablation}
\end{table}

Based on the ablation results from~\cref{tab:mmd_kernel_ablation}, the Laplacian kernel performs best on text tasks, but all kernels show limited effectiveness on more complex domains.

\subsection{Ablation Study on $\mdm$ with Different Number of Clusters}
\begin{table}[H]
\centering
\renewcommand{\arraystretch}{1.2}
\caption{Pearson / Spearman coefficients of $\mdm$ with different number of clusters (K=3, 5, 10, 20) on sentiment analysis.}
\begin{tabular}{cccc}
\hline
\textbf{K=3} & \textbf{K=5} & \textbf{K=10} & \textbf{K=20} \\ \hline
.85/ .68     & .86/ .68     & .86/ .69      & .86/ .69      \\ \hline
\end{tabular}
\label{tab:ablation_mdm_cluster}
\end{table}
The table with varying number of medoid clusters for $\mdm$ shows that changing the number of clusters has negligible effect on the correlation between $\mdm$ and F1 score.

\subsection{Ablation Study on Using Different Encoder for Representation-Based Metrics}
The results with BGE-M3\footnote{\url{https://huggingface.co/BAAI/bge-m3}} in~\cref{tab:ablation_encoder} demonstrate that representation-based metrics are measuring fundamental distribution shifts rather than artifacts of a specific encoder, therefore indicating the effectiveness of representation-based metrics.
\begin{table}[H]
\centering
\renewcommand{\arraystretch}{1.2}

\caption{Pearson / Spearman correlation coefficients of proxies on sentiment analysis with BGE-M3 as the encoder. }
\begin{tabular}{clclclcl}
\hline
\multicolumn{2}{c}{\textbf{$\mauve$}} & \multicolumn{2}{c}{\textbf{$\pad$}} & \multicolumn{2}{c}{\textbf{$\mmd$}} & \multicolumn{2}{c}{\textbf{$\mdm$}} \\ \hline
\multicolumn{2}{c}{.60/ .65}       & \multicolumn{2}{c}{.75/ .65}     & \multicolumn{2}{c}{.82/ .67}     & \multicolumn{2}{c}{.86/ .74}     \\ \hline
\end{tabular}
\label{tab:ablation_encoder}
\end{table}

\subsection{Correlation of $\pad$ Predictions from Different Classifiers vs. F1 Scores}
\begin{table}[h]
\centering
\renewcommand{\arraystretch}{1.2}
\caption{Correlation Coefficients (Pearson / Spearman) of $\pad$ predictions from different classifiers vs. F1 scores.}
\begin{tabular}{cc}
\hline
\textbf{Classifier}          & \textbf{Correlation Coefficients (P/S)} \\ \hline
Logistic Regression & .65 / .53                      \\
Random Forest       & .75 / .64                      \\
2-layer MLP         & .56 / .43                      \\ \hline
\end{tabular}
\label{tab:ablation_pad_classifier}
\end{table}
In summary, according to~\cref{tab:ablation_pad_classifier}, classifier choice plays a critical role in $\pad$ prediction performance, with the random forest model demonstrating the highest correlation with F1 scores. This may be due to the ensemble nature of both the random forest and the sentiment analysis classifier (XGBoost), possibly enhancing alignment between predictions and F1 outcomes, while the MLP shows weaker correspondence.

\subsection{Trade-Off Analysis Between \ourmethodnameshort{} and Diversity Measure}
\begin{figure}[H]
    \centering
    \includegraphics[width=0.8\linewidth]{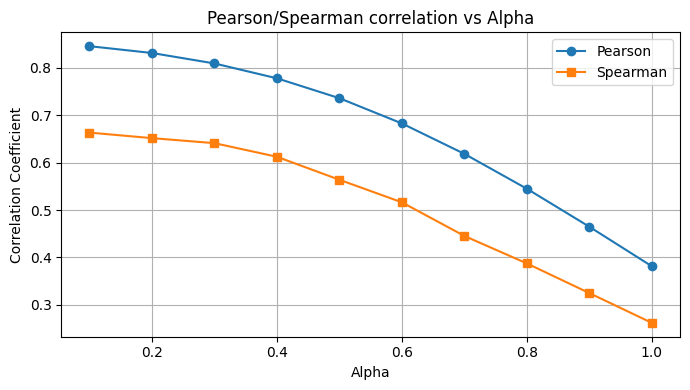}
    \caption{Pearson/Spearman correlation coefficients of the hybrid score vs. alpha. Both correlation coefficients are monotonically decreasing as alpha increases (more \ourmethodnameshort{} impact and less $\mdm$ impact).}
    \label{fig:lens_mdm_trade_off}
\end{figure}

To further understand the interplay between \ourmethodnameshort{} and the diversity measure, we conducted an experiment using a "hybrid" score, defined as $\alpha \cdot \ourmethodnameshort{} + (1-\alpha) \cdot \mdm$. As shown in~\cref{fig:lens_mdm_trade_off}, our analysis reveals that increasing $\alpha$ -- thereby placing greater weight on LENS and less on diversity -- results in a monotonic decrease in both the Pearson and Spearman correlation coefficients. This trend suggests that the LENS and diversity measures do not demonstrate synergistic effects when combined.

\subsection{Analysis on Using Hybrid Proxies for Dataset Selection}

\begin{table}[h]
    \centering
    \caption{
        Top 10 hybrid proxies by Spearman ($\rho$) and Pearson ($r$) correlation for dataset selection in the Sentiment Analysis (upper block) and Text2SQL (lower block) domains. Each row details a specific weighted combination of $\pad$, $\mmd$, $\mdm$, $\ourmethodnameshort$, and $\mauve$, alongside its resulting correlation to downstream task performance. For baseline comparison, $\hat{\rho}$ and $\hat{r}$ denote the maximum correlation achieved by the best-performing individual metric from \cref{tab:correlation_all}, whereas $\rho$ and $r$ denote the correlation achieved by the hybrid combination.
    }
    \label{tab:hybrid_proxies_corr_all_domains}
    \renewcommand{\arraystretch}{1.2}
    \setlength{\tabcolsep}{2.8pt}
    \begin{tabular}{c c c c c c c | c c c c c c c}
        \hline
        \multicolumn{7}{c|}{\textbf{Spearman}} & \multicolumn{7}{c}{\textbf{Pearson}} \\ \hline
        \textbf{$\hat{\rho}$} & \textbf{$\rho$} & \textbf{$\pad$} & \textbf{$\mmd$} & \textbf{$\mdm$} & \textbf{$\ourmethodnameshort$} & \textbf{$\mauve$}
        & \textbf{$\hat{r}$} & \textbf{$r$} & \textbf{$\pad$} & \textbf{$\mmd$} & \textbf{$\mdm$} & \textbf{$\ourmethodnameshort$} & \textbf{$\mauve$} \\ \hline
        \multicolumn{14}{c}{\textbf{Sentiment Analysis}} \\ \hline
        0.68 & 0.72  & 0.0 & 0.4 & 0.8 & 0.0 & 0.2 & 0.85 & 0.85 & 0.0 & 0.0 & 0.2 & 0.0 & 0.0 \\
        0.68 & 0.72  & 0.0 & 0.6 & 1.0 & 0.0 & 0.4 & 0.85 & 0.85 & 0.0 & 0.0 & 0.4 & 0.0 & 0.0 \\
        0.68 & 0.72  & 0.0 & 0.4 & 0.6 & 0.0 & 0.2 & 0.85 & 0.85 & 0.0 & 0.0 & 0.6 & 0.0 & 0.0 \\
        0.68 & 0.71  & 0.0 & 0.6 & 0.8 & 0.0 & 0.4 & 0.85 & 0.85 & 0.0 & 0.0 & 0.8 & 0.0 & 0.0 \\
        0.68 & 0.71  & 0.2 & 0.6 & 1.0 & 0.0 & 0.6 & 0.85 & 0.85 & 0.0 & 0.0 & 1.0 & 0.0 & 0.0 \\
        0.68 & 0.71  & 0.0 & 0.6 & 1.0 & 0.0 & 0.2 & 0.85 & 0.85 & 0.2 & 0.0 & 1.0 & 0.0 & 0.2 \\
        0.68 & 0.71  & 0.0 & 0.8 & 1.0 & 0.0 & 0.6 & 0.85 & 0.85 & 0.2 & 0.0 & 0.8 & 0.0 & 0.2 \\
        0.68 & 0.71  & 0.2 & 0.4 & 0.8 & 0.0 & 0.4 & 0.85 & 0.85 & 0.0 & 0.0 & 1.0 & 0.0 & 0.2 \\
        0.68 & 0.71  & 0.0 & 0.8 & 1.0 & 0.0 & 0.4 & 0.85 & 0.85 & 0.0 & 0.2 & 1.0 & 0.0 & 0.2 \\
        0.68 & 0.71  & 0.0 & 0.2 & 1.0 & 0.0 & 0.0 & 0.85 & 0.84 & 0.0 & 0.0 & 0.8 & 0.0 & 0.2 \\
        \hline
        \multicolumn{14}{c}{\textbf{Text2SQL}} \\ \hline
        0.65 & 0.89 & 0.6 & 0.8 & 0.4 & 0.6 & 0.2 & 0.68 & 0.79 & 0.0 & 1.0 & 0.2 & 0.8 & 0.2 \\
        0.65 & 0.89 & 0.6 & 0.6 & 0.4 & 0.6 & 0.2 & 0.68 & 0.79 & 0.0 & 0.8 & 0.2 & 0.8 & 0.2 \\
        0.65 & 0.89 & 0.2 & 0.2 & 0.6 & 0.8 & 0.2 & 0.68 & 0.79 & 0.0 & 0.6 & 0.2 & 0.8 & 0.2 \\
        0.65 & 0.89 & 0.0 & 0.2 & 0.6 & 0.8 & 0.2 & 0.68 & 0.79 & 0.0 & 0.4 & 0.2 & 0.8 & 0.2 \\
        0.65 & 0.89 & 0.2 & 0.4 & 0.4 & 0.6 & 0.2 & 0.68 & 0.79 & 0.0 & 0.2 & 0.2 & 0.8 & 0.2 \\
        0.65 & 0.89 & 0.4 & 1.0 & 0.4 & 0.6 & 0.2 & 0.68 & 0.79 & 0.0 & 0.0 & 0.2 & 0.8 & 0.2 \\
        0.65 & 0.89 & 0.2 & 0.2 & 0.4 & 0.6 & 0.2 & 0.68 & 0.79 & 0.0 & 1.0 & 0.2 & 1.0 & 0.2 \\
        0.65 & 0.89 & 0.4 & 0.6 & 0.4 & 0.6 & 0.2 & 0.68 & 0.79 & 0.0 & 0.8 & 0.2 & 1.0 & 0.2 \\
        0.65 & 0.89 & 0.4 & 0.4 & 0.4 & 0.6 & 0.2 & 0.68 & 0.79 & 0.0 & 0.6 & 0.2 & 1.0 & 0.2 \\
        0.65 & 0.89 & 0.6 & 0.2 & 0.4 & 0.6 & 0.2 & 0.68 & 0.79 & 0.0 & 0.4 & 0.2 & 1.0 & 0.2 \\
        \hline
    \end{tabular}
\end{table}

To understand the synergistic potential of different synthetic data metrics, we evaluated over 7,700 linear combinations (6 settings for each proxy, ranging from 0.0 to 1.0 with a step size of 0.2, resulting in $6^5$ combinations) of our five primary proxies. We expanded this experiment to include both the Sentiment Analysis and Text2SQL ``movie\_platform'' domains (as this domain provides the most instances and is therefore a more stable Text2SQL domain) to analyze how hybrid performance scales with task complexity. Here are the core findings based on results from~\cref{tab:hybrid_proxies_corr_all_domains}:

\begin{itemize}
    \item \textbf{Accurate Dataset Ranking Requires Synergy:} While individual metrics establish a baseline, combining them yields strict improvements, particularly for rank-ordering. In Sentiment Analysis, blending metrics increases the maximum single-proxy Spearman correlation ($\max \rho$) from 0.68 to 0.72. In the more complex Text2SQL domain, the synergistic gains are significant: the hybrid approach boosts the maximum single-proxy Spearman correlation ($\max \rho$) from 0.65 to 0.89, and the maximum Pearson correlation ($\max r$) from 0.68 to 0.79.
    \item \textbf{The Optimal Blend Varies from Task to Task:} The optimal hybrid composition shifts significantly depending on the nature of the task -- there is no single combination of proxies that performs the best across tasks. For simple stylistic tasks like Sentiment Analysis, representation-based metrics dominate the top combinations, with $\mdm$ acting as a primary anchor. However, for structurally complex reasoning tasks like Text2SQL, $\ourmethodnameshort$ becomes a critical component for both Pearson and Spearman correlations, consistently requiring heavy weights (between 0.6 and 1.0) to achieve optimal utility prediction.
    \item \textbf{Proxies Capture Orthogonal Blind Spots:} The necessity of this blend confirms that these metrics measure fundamentally distinct properties. A synthetic dataset might exhibit excellent diversity but drift out-of-domain, or it might perfectly mimic the real distribution's style while suffering from severe mode collapse. Furthermore, representation metrics may miss nuanced logical errors in complex queries that LLM-based reasoning ($\ourmethodnameshort$) catches. The top-performing hybrid proxies act as a counterbalance, successfully penalizing datasets that fail on any front.
\end{itemize}

Overall, if the goal is to select the absolute best synthetic dataset for downstream training, practitioners must use a holistic scoring approach that demands broad embedding coverage (i.e., diversity), strict distributional alignment, and -- particularly for complex tasks -- explicit reasoning-based evaluation.

\subsection{Ablation on the Number of Samples Used for Scoring on Simple Task}
\begin{table}[H]
\centering
\renewcommand{\arraystretch}{1.2}
\caption{Pearson / Spearman correlation between the \ourmethodnameshort{} score and the test F1 score for different metrics with differet number of real instances used for scoring (100 and 200).}
\begin{tabular}{cccccccccc}
\hline
\multicolumn{5}{c}{$|\samples_\real|=100$}   & \multicolumn{5}{c}{$|\samples_\real|=200$} \\ \hline
\ourmethodnameshort{}      & $\pad$     & $\mmd$     & $\mdm$     & $\mauve$   & \ourmethodnameshort{}    & $\pad$     & $\mmd$     & $\mdm$     & $\mauve$   \\
-.03/-.13 & .65/.47 & .68/.44 & .85/.70 & .54/.51 & .26/.38 & .65/.53 & .67/.45 & .85/.68 & .57/.53 \\ \hline
\end{tabular}
\label{tab:sa_scoring_ablation_num_samples}
\end{table}

We investigate the sensitivity of LENS proxies to the size of the unannotated real dataset $|\mathcal{U}_r|$ for the sentiment analysis task.As shown in~\cref{tab:sa_scoring_ablation_num_samples}, representation-based metrics ($\pad$, $\mmd$, $\mdm$, $\mauve$) demonstrate remarkable robustness in this domain, maintaining strong correlations (e.g., $\mdm$ Spearman $0.70$) even with as few as 100 samples. This aligns with our hypothesis that embedding spaces effectively capture distributional shifts in simpler, stylistic tasks. In contrast, \ourmethodnameshort{} requires a larger sample size to stabilize for this specific task 5, exhibiting weak correlations at $|\mathcal{U}_r|=100$ (Pearson $-0.03$, Spearman $-0.13$) before recovering significantly to a positive Spearman correlation of $0.38$ at $|\mathcal{U}_r|=200$. This behavior highlights a distinct complexity trade-off: while representation metrics are highly data-efficient for stylistic domains, they fail on complex, long-horizon tasks such as Web Navigation. Conversely, \ourmethodnameshort{} proves robust in those more complex domains even with very limited unlabled real data.

\subsection{Correlation Improvement with De-Hallucination Prompt}
\begin{table}[H]
\centering
\renewcommand{\arraystretch}{1.2}
\caption{Spearman and Pearson correlations for baseline and with de-hallucination prompt (w/ DHP), across different models.}
\label{tab:dehallucinated_corr_sa}
\begin{tabular}{lcccc}
\hline
 & \textbf{Baseline} & \textbf{w/ DHP (Delta)} \\
\hline
\multicolumn{3}{l}{\textbf{Qwen2.5-7B-Instruct}} \\
\hspace{1mm} Spearman & 0.25 & 0.15\ (\textminus0.10) \\
\hspace{1mm} Pearson  & 0.33 & 0.30\ (\textminus0.03) \\
\hline
\multicolumn{3}{l}{\textbf{Gemma-3-12B-It}} \\
\hspace{1mm} Spearman & 0.10 & 0.19\ (+0.09) \\
\hspace{1mm} Pearson  & 0.13 & 0.24\ (+0.11) \\
\hline
\multicolumn{3}{l}{\textbf{Ministral-8B-Instruct-2410}} \\
\hspace{1mm} Spearman & 0.28 & 0.53\ (+0.25) \\
\hspace{1mm} Pearson  & 0.39 & 0.49\ (+0.10) \\
\hline
\multicolumn{3}{l}{\textbf{Granite3.3-8B-Instruct}} \\
\hspace{1mm} Spearman & 0.21 & 0.23\ (+0.02) \\
\hspace{1mm} Pearson  & 0.30 & 0.32\ (+0.02) \\
\hline
\end{tabular}
\end{table}
As shown in Table~\ref{tab:dehallucinated_corr_sa}, incorporating an explicit de-hallucination prompt during the LLM-based scoring phase produces highly model-dependent effects -- four out of three models improve correlations.

To incorporate de-hallucination into the LENS scoring prompt, we add the following sentence:
\begin{tcolorbox}[colback=gray!10, colframe=gray!50!black, boxrule=0.5pt, arc=4pt, left=2pt, right=2pt, top=2pt, bottom=2pt]
\texttt{If you find the headline is hallucinated (i.e., not a real headline), answer "unsure".}
\end{tcolorbox}
We find that the impact of de-hallucination prompts on correlation metrics is highly model-dependent: while some models benefit noticeably with improved alignment (e.g., substantial gains in proxy reliability for certain models), others show little change or even reduced correlations. This indicates that prompt-based de-hallucination provides inconsistent results across models, suggesting that more robust, systematic approaches are needed to reliably reduce hallucinations during evaluation.

\subsection{Ablation on the Size of Synthetic Dataset for \ournameshort{} Proxies}
\begin{table}[h]
\centering
\renewcommand{\arraystretch}{1.2} 
\caption{
Scaling analysis of \ournameshort{} proxies for sentiment analysis.
Table entries show Pearson/Spearman correlation coefficients, in the form ``Pearson/Spearman'', for each proxy at different synthetic dataset size multiples.
We report results using debiased \ourmethodnameshort{}, evaluated with \texttt{Qwen2.5-(7B/32B)-Instruct} scorers.
A size multiple of $n$ indicates that $n$ synthetic datasets (each containing 999 samples) are merged, yielding $n \times 999$ total samples.
}
\label{tab:scaling_results_sa}
\begin{tabular}{@{}ccccccc@{}}
\toprule
\textbf{Size} & \textbf{LENS-7B} & \textbf{LENS-32B} & $\mdm$ & \textbf{$\mmd$} & $\pad$ & $\mauve$ \\ 
\midrule
1  & .33/.25 & .26/.38 & .85/.68 & .67/.45 & .65/.53 & .57/.53 \\
2  & .17/.21 & .11/.16 & .69/.52 & .55/.50 & .51/.43 & .35/.36 \\
4  & .17/.12 & .38/.30 & -.70/-.81 & -.27/-.42 & -.25/-.43 & -.31/-.32 \\
8  & .04/.04 & .49/.44 & -.51/-.20 & .62/.40 & .04/.40 & -.20/.24 \\
\bottomrule
\end{tabular}
\end{table}

We perform a scaling experiment on sentiment analysis to evaluate how well \ournameshort{} proxies maintain their effectiveness as the size of synthetic datasets increases. Sentiment analysis provides the most abundant synthetic training data (32 datasets), enabling us to systematically merge $n$ datasets together, $n \in \left\{1, 2, 4, 8\right\}$.

From the results in~\cref{tab:scaling_results_sa}, we observe that increasing the synthetic dataset size generally leads to performance degradation across all proxies. In particular, when the size multiple reaches 4, \ourmethodnameshort{} remains capable of achieving a positive correlation, suggesting greater robustness to growing dataset size compared to other methods. It is important to note that, throughout these experiments, the number of synthetic samples chosen for rubric compilation in \ourmethodnameshort{} is fixed at 200, which means that as dataset size increases, the relative proportion of data represented in the scoring rubric decreases, potentially limiting its ability to capture the full variance present in larger datasets.

\subsection{Ablation on Instance-Level Data Selection}
\begin{table}[h]
\centering
\renewcommand{\arraystretch}{1.2}
\caption{Instance-level downstream task performance (\%) on sentiment analysis for synthetic datasets selected by each \ournameshort{} proxy. Debiased \ourmethodnameshort{} uses  \texttt{Qwen2.5-7B-Instruct} as the backbone model for scoring}
\label{tab:instance_level_performance_sa}
\begin{tabular}{ccccccc}
\hline
\textbf{Proxy}  & \textbf{Average} & \ourmethodnameshort & $\pad$ & $\mmd$ & $\mdm$ & $\mauve$ \\
\textbf{Result} & 49.6             & 54.8          & 55.9         & 53.2                            & n/a          & 52.7           \\ \hline
\end{tabular}
\end{table}

We explore whether \ournameshort{} proxies can be leveraged for \textit{instance-level} data selection, as opposed to the default \textit{dataset-level} ranking. In this ablation, for sentiment analysis, we aggregate all deduplicated synthetic datasets into a single pool containing 29,235 instances. From this pool, for each class, we select the top 333 instances according to each proxy score, yielding a final synthetic dataset of 999 samples (333 per class).

The results in~\cref{tab:instance_level_performance_sa} reveal: \textit{discriminative} proxies -- namely $\pad$ and \ourmethodnameshort{} -- achieve higher downstream task performance when selecting at the instance level, compared to their performance with dataset-level selection. Specifically, incorporating comparisons to~\cref{tab:top3_all}, instance-level selection improves $\pad$'s performance (55.9\% vs.\ 55.3\%) and \ourmethodnameshort{}'s performance (54.8\% vs.\ 50.5\%) over the dataset-level setting.

In contrast, for the other proxies such as $\mmd$ and $\mauve$, switching from dataset-level to instance-level selection leads to reduced task performance, indicating that these methods are less suitable for fine-grained instance selection. We also note that $\mdm$ is not applicable at the instance level, since it measures diversity across a set of samples and cannot score individual ones -- hence it is marked as ``n/a'' in the table.

Overall, these results demonstrate that discriminative proxies can better exploit instance-level selection, whereas distributional measures suffer from a loss of signal in this setting.

\section{Additional Experimental Details}
\label{app:experiment}
\subsection{Data Curation}
\paragraph{Sentiment analysis}
For sentiment analysis, we used a cleaned version of the original validation split as the test set, removing URLs from the data. Synthetic samples are generated and validated to ensure each output is non-empty, alphabetic 'headline' and a 'sentiment' label restricted to the values '0', '1', or '2'. Only samples meeting these structural and content criteria were retained for downstream analysis.
\paragraph{Text2SQL}
During Text2SQL data synthesis, we validate generated question--SQL pairs by ensuring both fields are non-empty, contain alphanumeric content, and are formatted as a dictionary with 'question' and 'SQL' keys. For non-zero-shot generations, SQL queries are executed against the target database; pairs failing execution are discarded. This process enforces correct structure, meaningful content, and SQL executability.
\paragraph{Image classification}
We classify each image in the \texttt{unmet-promise} dataset into one of three categories: \textit{label\_only}, \textit{label\_relation}, or \textit{label\_background}, based on its caption. Captions are lowercased and stemmed. If a class-specific background keyword appears in the caption, the image is assigned to \textit{label\_background}. If a relation keyword is present, it is assigned to \textit{label\_relation}. Images not matching either are assigned to \textit{label\_only}. We balance the number of samples per category and class, and store the processed data for downstream tasks.
To further ensure fidelity of images, we use \texttt{Qwen2.5-VL-7B-Instruct} to filter noisy images. Prompt used for filtering is provided in listing ~\ref{lst:filter_prompt}.
For image classification, we create a test set for each split by selecting 150 samples that are balanced across all labels.

\paragraph{Web navigation}
We first group individual trajectories into tasks. Then we use different seeds to create 5 disjoint subsets with equal amount of tasks. We fine-tune using LoRA rank of 64 and the open-instruct~\footnote{\url{https://github.com/allenai/open-instruct}} fine-tuning codebase.

\subsection{\ournameshort\ Scoring}
For $\pad$, we reserve 20\% of all embeddings as a holdout test set to train the classifier, and compute the classification error on this set to obtain the final $\pad$ score. For $\mmd$, we generate the proxy score using a polynomial kernel~\footnote{\url{https://scikit-learn.org/stable/modules/generated/sklearn.metrics.pairwise.polynomial_kernel.html}} with degree 3 and Coef0 parameter 1. For $\mdm$, we use Fasterpam~\citep{schubert_faster_2019} to compute $k$ medoids in the embeddings, setting $k$ to 3 for sentiment analysis and 5 for other tasks, using Euclidean distance for clustering; $\mdm$ is then calculated by averaging the Euclidean distance from each data point to its corresponding medoid. For $\ourmethodnameshort$, the prompts used for rubric compilation and scoring are provided in~\cref{sec:rubric_compilation_prompt} and~\cref{sec:scoring_prompt}, respectively.
For \mauve\ calculation, we use the default hyperparameter, the same embedding model as the other representation-based proxies (\textit{qte-Qwen2-7B-Instruct}), and the scaling factor is 5.
For \perplexity\ calculation in Text2SQL, we first reformat each Text2SQL question as a string: \fcolorbox{gray!75!black}{gray!5!white}{\texttt{"question": <sample>}}. We then use \texttt{Qwen2.5-7B} to compute the perplexity, considering only the tokens corresponding to the question text (i.e., the <sample> portion), and excluding the prompt tokens (such as \texttt{"question":}).

\subsection{Dataset Size Descriptions}
\begin{table}[H]
\centering
\caption{Dataset sizes for synthetic ($\dataset_\synth$), rubric selection, real subset ($\samples_\real$), and full real dataset ($\dataset_\real$) for all tasks.}
\renewcommand{\arraystretch}{1.2}
\begin{tabular}{ccccc}
\hline
\textbf{Task}        & \textbf{Synthetic} & \textbf{LENS Rubric Compilation} & \textbf{Real Subset} & \textbf{Full Real} \\ \hline
Sentiment Analysis   & 999                & 400                                           & 200                  & 2388               \\ \hline
\multicolumn{5}{c}{\textit{Text-to-SQL}}                                                                                              \\
Movie Platform       & 1000               & 60                                            & 30                   & 164                \\
App Store            & 1000               & 60                                            & 30                   & 60                 \\
Computer Student     & 1000               & 60                                            & 30                   & 69                 \\ \hline
Image Classification & 1500               & 200                                           & 100                  & 150                \\ \hline
\multicolumn{5}{c}{\textit{Web Navigation}}                                                                                           \\
Allrecipes           & 79                 & 99                                            & 20                   & 45                 \\
Amazon               & 63                 & 83                                            & 20                   & 41                 \\
Apple                & 70                 & 90                                            & 20                   & 43                 \\
ArXiv                & 80                 & 100                                           & 20                   & 43                 \\
BBC                  & 69                 & 89                                            & 20                   & 42                 \\
Coursera             & 72                 & 92                                            & 20                   & 42                 \\
dictionary           & 54                 & 74                                            & 20                   & 43                 \\
ESPN                 & 62                 & 82                                            & 20                   & 44                 \\
GitHub               & 71                 & 91                                            & 20                   & 41                 \\
Google Maps          & 75                 & 95                                            & 20                   & 41                 \\
Google Search        & 72                 & 92                                            & 20                   & 43                 \\
Hugging Face         & 76                 & 96                                            & 20                   & 43                 \\
Wolfram Alpha        & 66                 & 86                                            & 20                   & 46                 \\ \hline
\end{tabular}
\end{table}

\begin{table}[H]
    \centering
    \renewcommand{\arraystretch}{1.2}
    \caption{Number of synthetic datasets available per task.}
    \label{tab:dataset_size_per_task}
    \begin{tabular}{cc}
        \hline
        \textbf{Task}        & \textbf{Number of Synthetic Datasets} \\ \hline
        Sentiment Analysis   & 32                                    \\
        Text-to-SQL          & 8                                     \\
        Image Classification & 6                                     \\
        Web Navigation       & 65                                    \\ \hline
    \end{tabular}
\end{table}

\subsection{Standard deviation of correlation coefficients}
\renewcommand{\arraystretch}{0}
\setlength{\fboxsep}{1pt} 
\setlength{\tabcolsep}{0pt}
\begin{table}[H]
\centering
\caption{Standard deviation of Spearman (left) and Pearson (right) correlation coefficients of \ournameshort\ proxy metrics across 5 seeds. $\mdm$ only scores on synthetic datasets therefore no change in input data.}
\vspace{0.1 in}
\renewcommand{\arraystretch}{1.2}
\begin{tabularx}{\linewidth}{XXXXXXXX}
\toprule
Tasks &
\multicolumn{2}{c}{\ourmethodnameshort\ 7B} &
\multicolumn{2}{c}{\ourmethodnameshort\ 32B} &
PAD &
MMD &
MDM \\
& debiased & biased & debiased & biased & & & \\ \midrule
\textbf{Sentiment}  &  \entrydualvar{ .23}{ .17}  &  \entrydualvar{ .12}{ .14}  &  \entrydualvar{ .09}{ .11}  &  \entrydualvar{ .07}{ .07}  &  \entrydualvar{ .02}{ .02}  &  \entrydualvar{ .02}{ .02}  &  \entrydualvar{ .00}{ .00} \\ \midrule
\textbf{Text2SQL} &&&&&&& \\
Computer & \entrydualvar{ .48}{ .39}  &  \entrydualvar{ .38}{ .44}  &  \entrydualvar{ .21}{ .21}  &  \entrydualvar{ .18}{ .14}
 & \entrydualvar{ .04}{ .08}  &  \entrydualvar{ .03}{ .08}  & \entrydualvar{ .00}{ .00} \\
Apps  &  \entrydualvar{ .38}{ .34}  &  \entrydualvar{ .25}{ .21}  &  \entrydualvar{ .22}{ .22}  &  \entrydualvar{ .22}{ .14} &  \entrydualvar{ .36}{ .48}  &  \entrydualvar{ .05}{ .15}  &  \entrydualvar{ .00}{ .00} \\
Movies  &  \entrydualvar{ .32}{ .32}  &  \entrydualvar{ .33}{ .24}  &  \entrydualvar{ .22}{ .24}  &  \entrydualvar{ .11}{ .14} &  \entrydualvar{ .07}{ .13}  &  \entrydualvar{ .05}{ .14}  &  \entrydualvar{ .00}{ .00} \\
\textit{Average}  &  \entrydualvar{ .39}{ .35}  &  \entrydualvar{ .32}{ .30}  &  \entrydualvar{ .22}{ .22}  &  \entrydualvar{ .17}{ .14} &  \entrydualvar{ .15}{ .23}  &  \entrydualvar{ .04}{ .12}  &  \entrydualvar{ .00}{ .00} \\ \midrule
\textbf{Image} &&&&&&& \\
Split 1  &  \entrydualvar{ .47}{ .48}  &  \entrydualvar{ .51}{ .51}  &  \entrydualvar{ .39}{ .38}  &  \entrydualvar{ .28}{ .22} &  \entrydualvar{ .08}{ .06}  &  \entrydualvar{ .00}{ .02}  &  \entrydualvar{ .00}{ .00} \\
Split 2  &  \entrydualvar{ .43}{ .47}  &  \entrydualvar{ .43}{ .42}  &  \entrydualvar{ .40}{ .41}  &  \entrydualvar{ .43}{ .53} &  \entrydualvar{ .11}{ .08}  &  \entrydualvar{ .00}{ .04}  &  \entrydualvar{ .00}{ .00}\\
Split 3  &  \entrydualvar{ .52}{ .39}  &  \entrydualvar{ .47}{ .39}  &  \entrydualvar{ .40}{ .42}  &  \entrydualvar{ .29}{ .22} &  \entrydualvar{ .07}{ .07}  &  \entrydualvar{ .00}{ .05}  &  \entrydualvar{ .00}{ .00}\\
Average  &  \entrydualvar{ .47}{ .44}  &  \entrydualvar{ .47}{ .44}  &  \entrydualvar{ .40}{ .40}  &  \entrydualvar{ .33}{ .32} &  \entrydualvar{ .09}{ .07}  &  \entrydualvar{ .00}{ .04}  &  \entrydualvar{ .00}{ .00}\\
\midrule

\textbf{WebNav}  &  \entrydualvar{.15}{.17}  &  \entrydualvar{.11}{ .13}  &  \entrydualvar{.09}{.11}  &  \entrydualvar{.15}{.19}  &  \entrydualvar{.04}{.02}  &  \entrydualvar{.02}{.02}  &  \entrydualvar{.00}{.00} \\ \bottomrule
\end{tabularx}
\label{tab:correlation_std_all}
\end{table}

\subsection{Scatter Plot of \ourmethodnameshort{} Scores and Task Utility}

\subsubsection{Sentiment Analysis}
Scatter plots of debiased \ournameshort{} scores and baselines vs. F1 utility on sentiment analysis datasets. Note that we use the raw scores of $\mmd$ and $\pad$ -- we negate them to have a positive correlation with task utility for dataset selection.
{
\begin{figure}[H]
    \centering
    \begin{tabular}{cc}
        \includegraphics[width=0.45\textwidth]{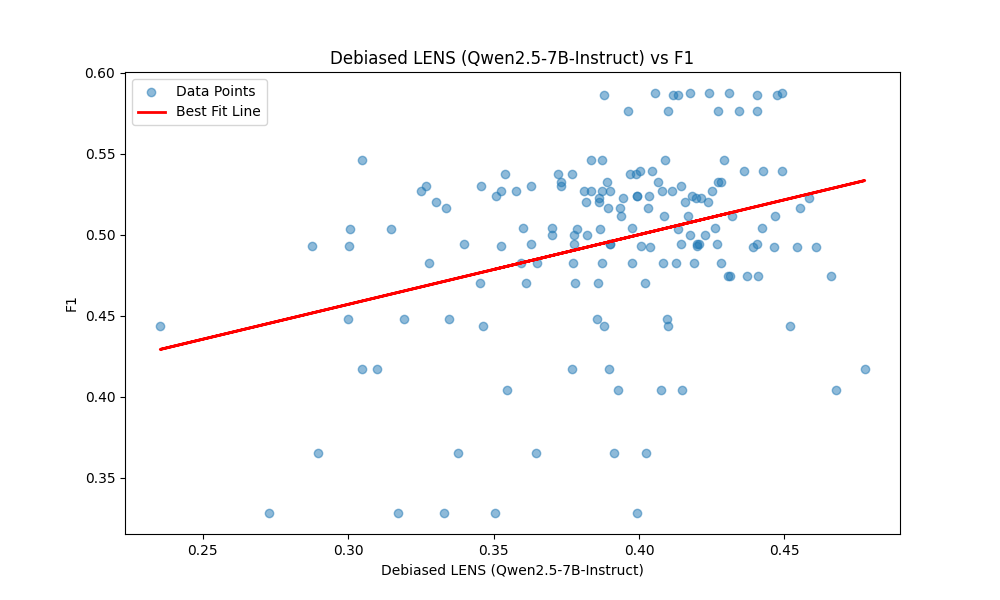} &
        \includegraphics[width=0.45\textwidth]{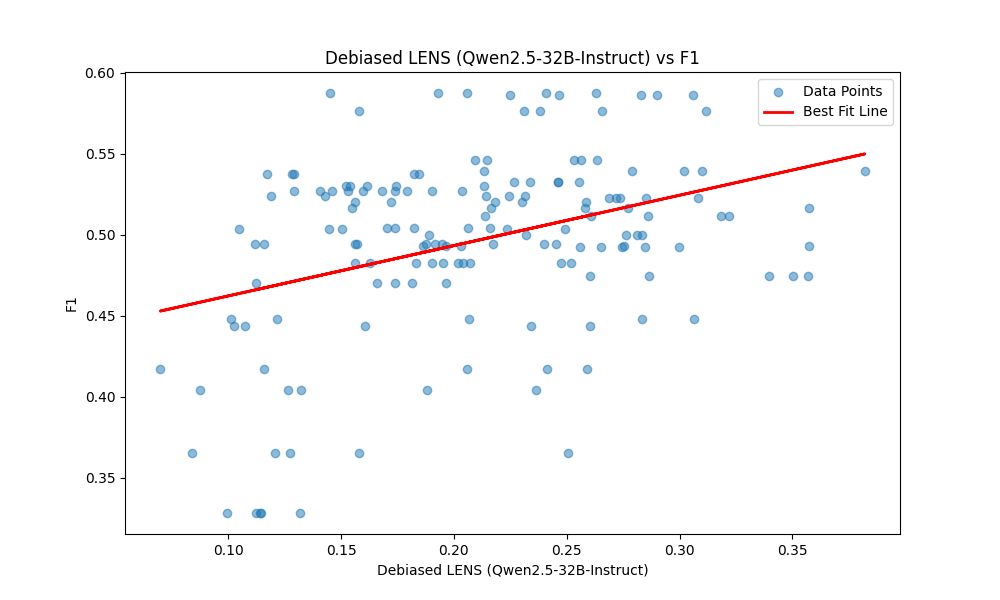} \\
        \small (a) Debiased \ourmethodnameshort{} (\texttt{Qwen2.5-7B-Instruct}) vs. F1 &
        \small (b) Debiased \ourmethodnameshort{} (\texttt{Qwen2.5-32B-Instruct}) vs. F1 \\[1.5em]
        \includegraphics[width=0.45\textwidth]{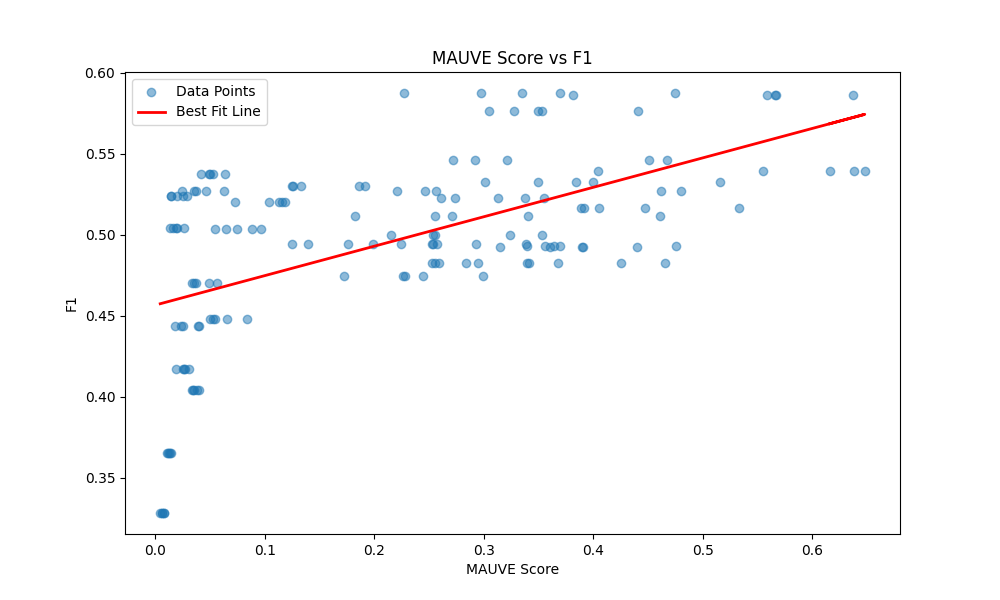} &
        \includegraphics[width=0.45\textwidth]{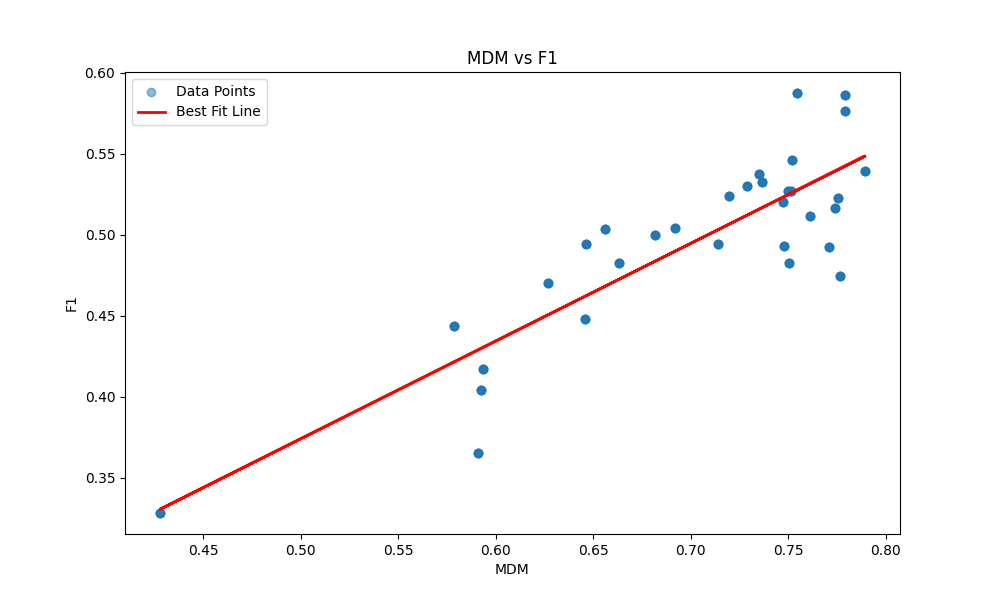} \\
        \small (c) $\mauve$ vs. F1 &
        \small (d) $\mdm$ vs. F1 \\[1.5em]
        \includegraphics[width=0.45\textwidth]{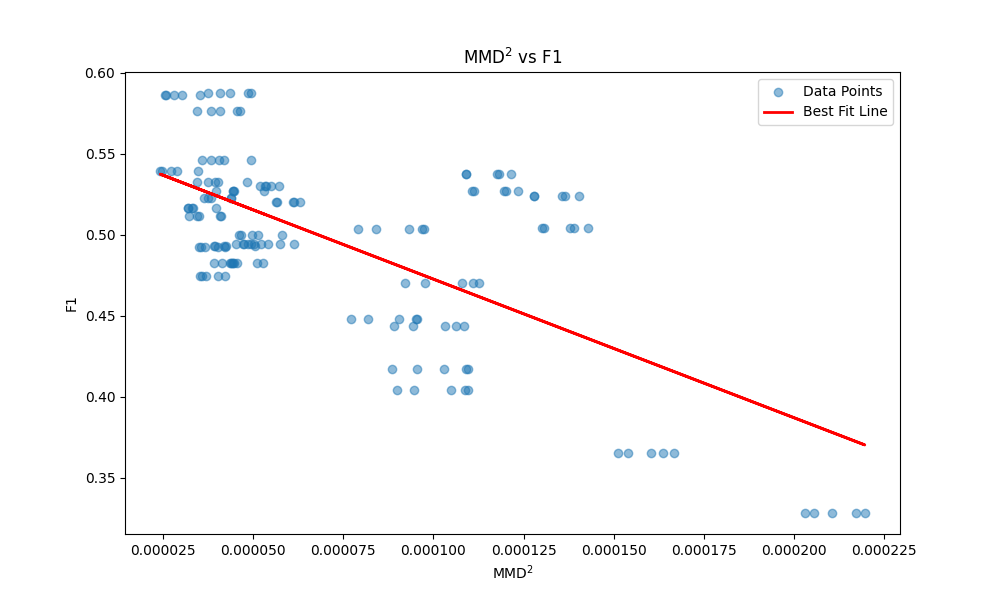} &
        \includegraphics[width=0.45\textwidth]{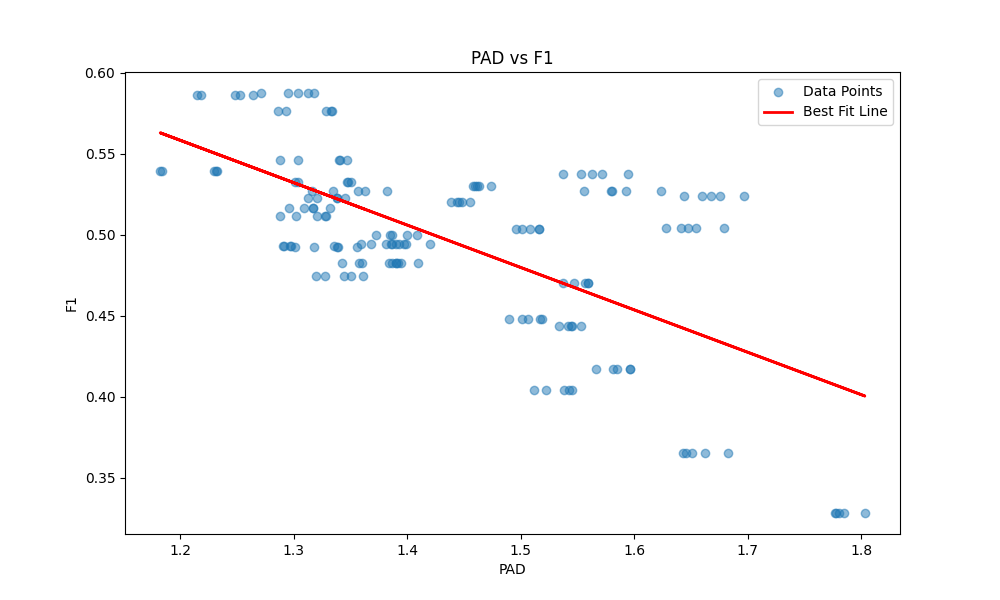} \\
        \small (e) $\mmd$ vs. F1 &
        \small (f) $\pad$ vs. F1 \\
    \end{tabular}
    \label{fig:scatter_sa_all_metrics}
\end{figure}
}
\clearpage
\subsubsection{Text-to-SQL}
Below are scatter plots of \ournameshort{} scores vs. accuracy utility on text-to-SQL datasets.
For $\mmd$ and $\pad$, we use raw scores and negate them for positive correlation with task utility for dataset selection.
{
\begin{figure}[H]
    \centering
    \begin{tabular}{c}
        \includegraphics[width=0.70\textwidth]{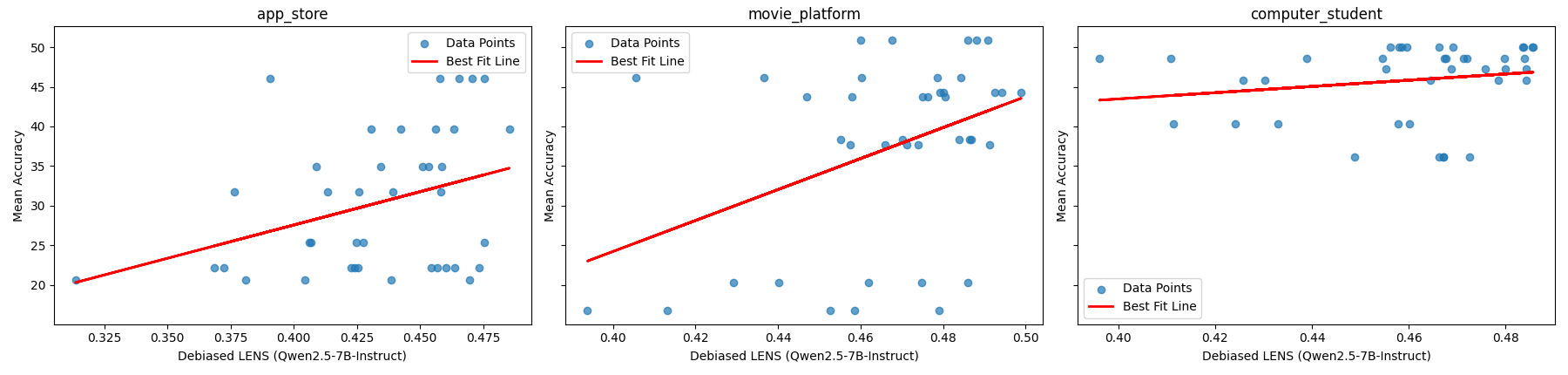} \\
        \small (a) Debiased \ourmethodnameshort{} (\texttt{Qwen2.5-7B-Instruct}) vs. Accuracy \\[1em]
        \includegraphics[width=0.70\textwidth]{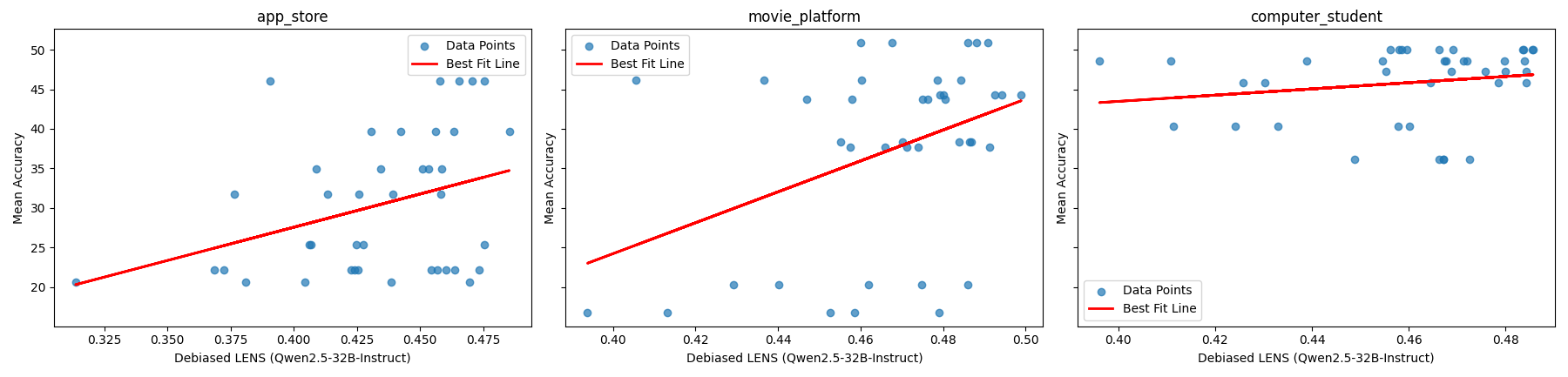} \\
        \small (b) Debiased \ourmethodnameshort{} (\texttt{Qwen2.5-32B-Instruct}) vs. Accuracy \\[1em]
        \includegraphics[width=0.70\textwidth]{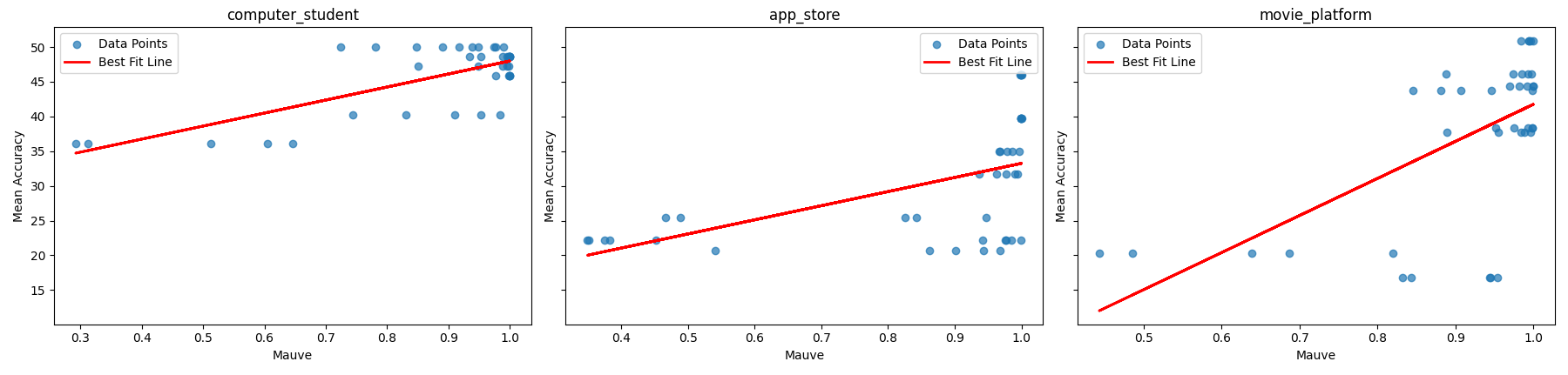} \\
        \small (c) $\mauve$ vs. Accuracy \\[1em]
        \includegraphics[width=0.70\textwidth]{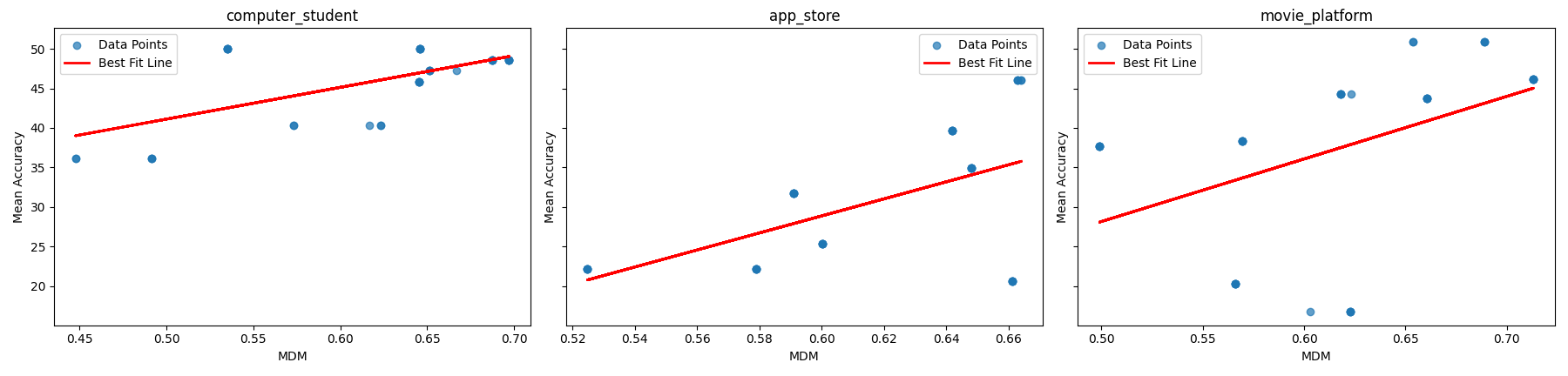} \\
        \small (d) $\mdm$ vs. Accuracy \\[1em]
        \includegraphics[width=0.70\textwidth]{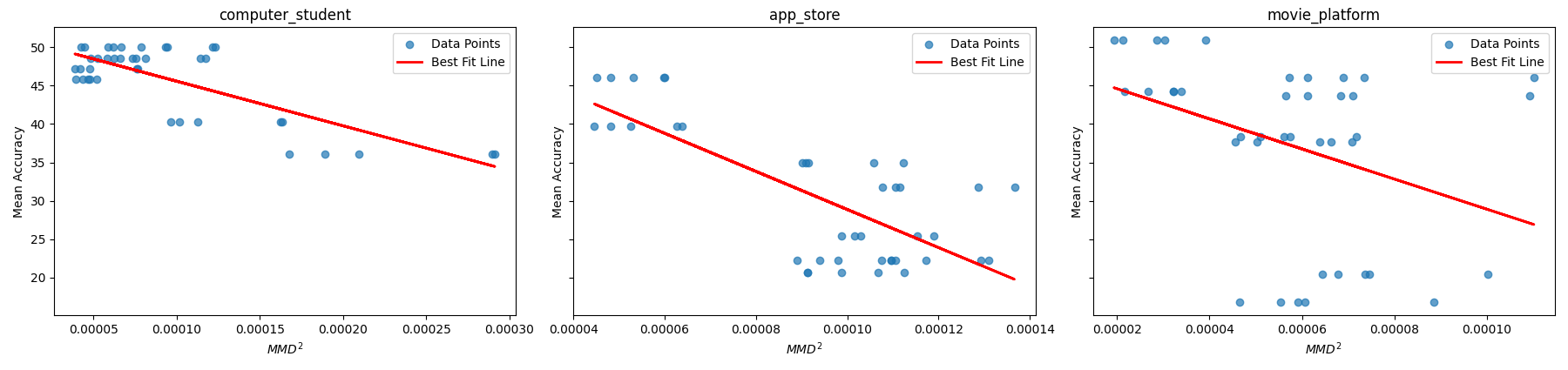} \\
        \small (e) $\mmd$ vs. Accuracy \\[1em]
        \includegraphics[width=0.70\textwidth]{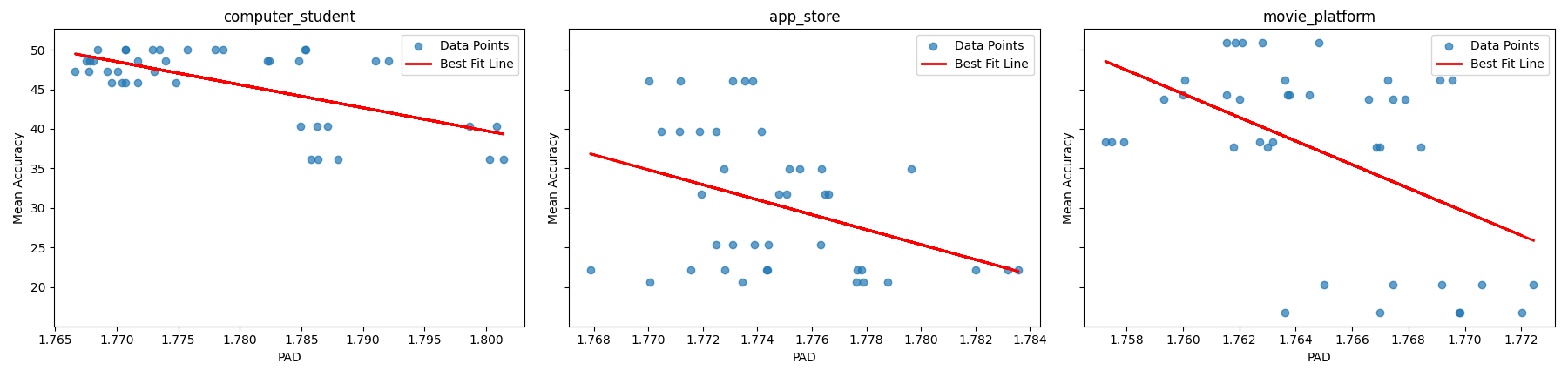} \\
        \small (f) $\pad$ vs. Accuracy \\
    \end{tabular}
    \label{fig:scatter_t2sql_all_metrics}
\end{figure}
}
\subsubsection{Image Classification}
Below are scatter plots \ournameshort{} scores vs. MRR utility on image classification datasets.
For $\mmd$ and $\pad$, we use raw scores and negate them for positive correlation with task utility for dataset selection.
{
\begin{figure}[H]
    \centering
    \begin{tabular}{c}
        \includegraphics[width=0.70\textwidth]{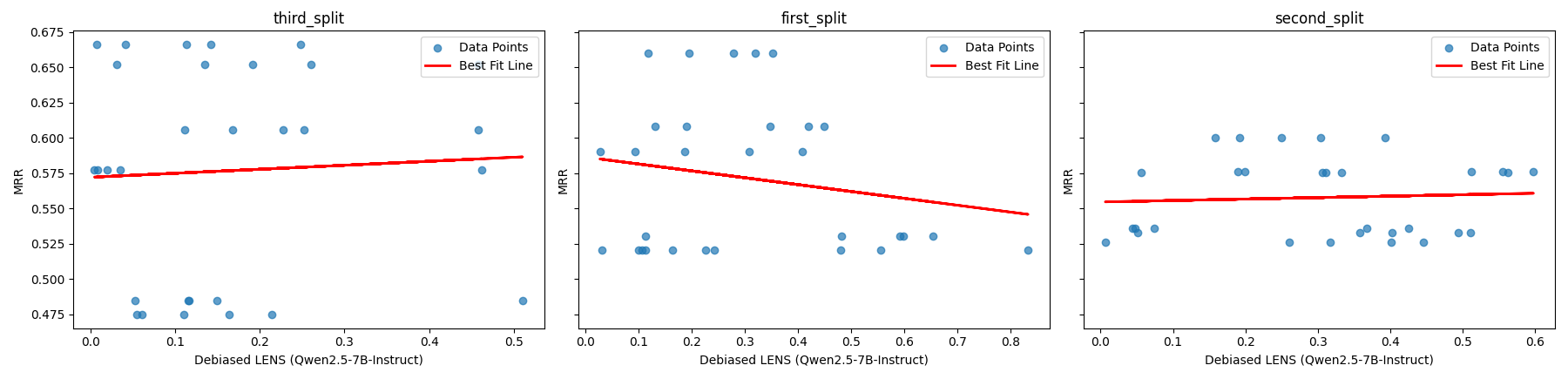} \\
        \small (a) Debiased \ourmethodnameshort{} (\texttt{Qwen2.5-7B-Instruct}) vs. MRR \\[1em]
        \includegraphics[width=0.70\textwidth]{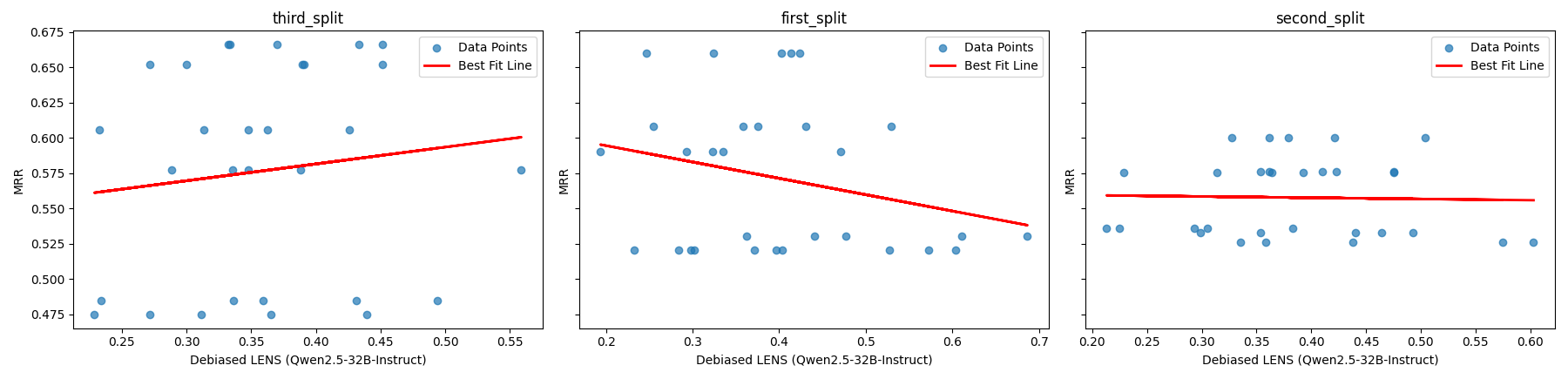} \\
        \small (b) Debiased \ourmethodnameshort{} (\texttt{Qwen2.5-32B-Instruct}) vs. MRR \\[1em]
        \includegraphics[width=0.70\textwidth]{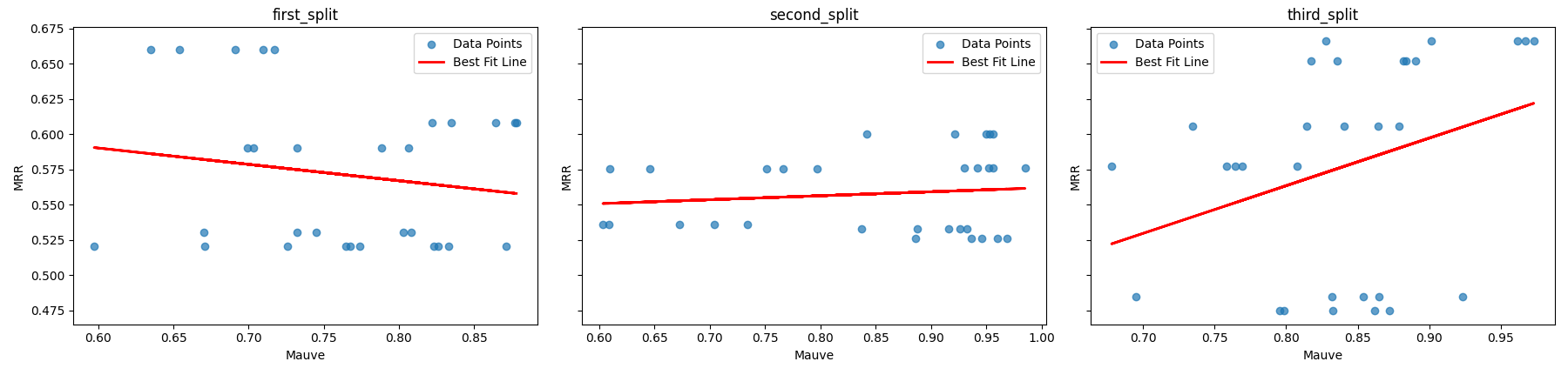} \\
        \small (c) $\mauve$ vs. MRR \\[1em]
        \includegraphics[width=0.70\textwidth]{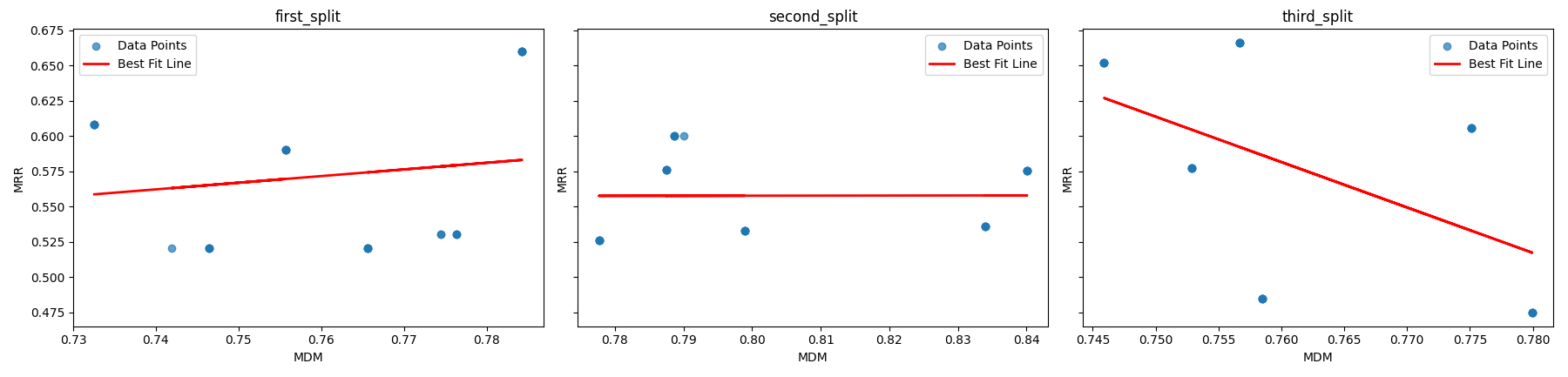} \\
        \small (d) $\mdm$ vs. MRR \\[1em]
        \includegraphics[width=0.70\textwidth]{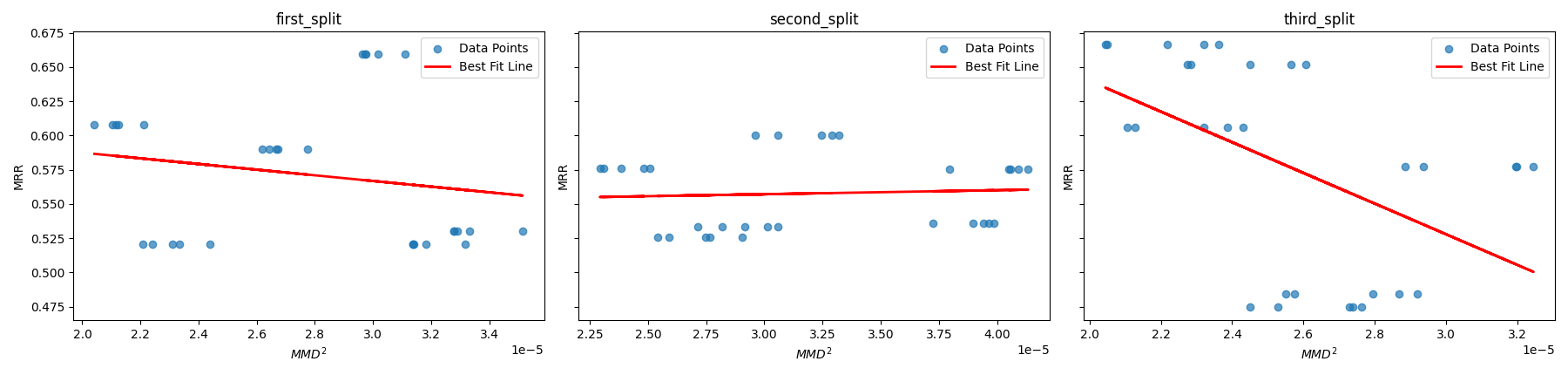} \\
        \small (e) $\mmd$ vs. MRR \\[1em]
        \includegraphics[width=0.70\textwidth]{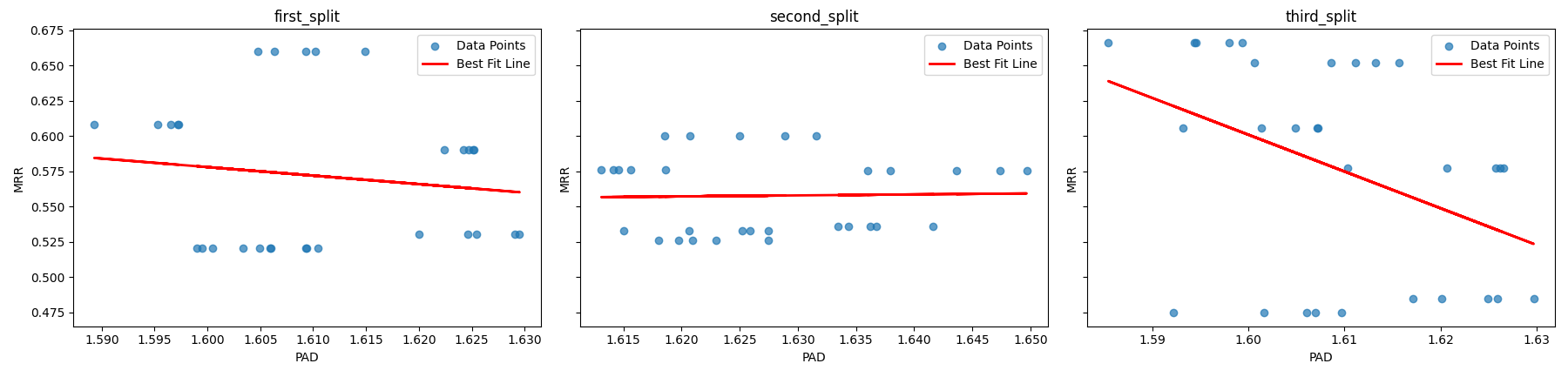} \\
        \small (f) $\pad$ vs. MRR \\
    \end{tabular}
    \label{fig:scatter_imgcls_all_metrics}
\end{figure}
}

\subsection{SynQuE proxy cost and efficiency analysis}
\begin{table}[h]
\centering
\renewcommand{\arraystretch}{1.2}
\setlength{\tabcolsep}{14pt} 
\caption{
Comparison of computational cost across different proxy methods in sentiment analysis.
Columns: \textbf{Emb.} (embedding latency), \textbf{Score} (scoring latency), \textbf{Total} (end-to-end latency), and \textbf{Tokens} (total tokens for \ourmethodnameshort, including input and output tokens).
Latencies reported in seconds per synthetic dataset (999 samples), with standard deviation in parentheses. \ourmethodnameshort{} uses Qwen2.5 models.}
\label{tab:method-cost}
\vspace{-0.6em}
\begin{tabular}{ccccc}
\hline
\textbf{Method}       & \textbf{Emb. (s/$\mathcal{D}$)} & \textbf{Score (s/$\mathcal{D}$)} & \textbf{Total (s/$\mathcal{D}$)} & \textbf{Num. of Tokens} \\ \hline
\ourmethodnameshort (7B)  & --         & 90.52 $\pm$ 4.84  & 90.52 $\pm$ 4.84  & 9264 $\pm$ 440      \\
\ourmethodnameshort (32B) & --         & 271.25 $\pm$ 20.54 & 271.25 $\pm$ 20.54 & 9264 $\pm$ 440      \\
$\pad$                   & 6.08 $\pm$ 0.66  & 1.99 $\pm$ 0.29   & 8.07 $\pm$ 0.72    & --                 \\
$\mdm$                   & 6.08 $\pm$ 0.66  & 0.24 $\pm$ 0.01   & 6.32 $\pm$ 0.66    & --                 \\
$\mmd$                   & 6.08 $\pm$ 0.66  & 0.04 $\pm$ 0.01   & 6.12 $\pm$ 0.66    & --                 \\
$\mauve$                 & 6.08 $\pm$ 0.66  & 106.43 $\pm$ 6.30 & 112.51 $\pm$ 6.33  & --                 \\ \hline
\end{tabular}
\end{table}
\vspace{-0.6em}

All computational cost analyses were performed on a system equipped with an Nvidia RTX Pro 6000 Blackwell Server Edition GPU\footnote{\url{https://www.nvidia.com/en-us/data-center/rtx-pro-6000-blackwell-server-edition/}} and a 12-core Intel(R) Xeon(R) Gold 5415+ CPU\footnote{\url{https://www.intel.com/content/www/us/en/products/sku/232373/intel-xeon-gold-5415-processor-22-5m-cache-2-90-ghz/specifications.html}}.

Across all sentiment analysis datasets, we observe that $\mmd$ offers the lowest end-to-end computational cost, closely followed by $\mdm$ and $\pad$, which also demonstrate strong efficiency. Notably, while \ourmethodnameshort{} relies on LLM-based scoring—which substantially increases latency compared to embedding-based proxies—its runtime in the 7B setting (90.52 s/$\mathcal{D}$) is still considerably lower than that of $\mauve$ (112.51 s/$\mathcal{D}$). This suggests that, although LLM scoring introduces additional overhead, \ourmethodnameshort{} can remain competitive in throughput relative to existing generative evaluation methods when using smaller-scale LLMs.

\subsection{Example Rubrics}
{
  \begin{lstlisting}[label=lst:rubric_example_sfr,caption=Example characteristic descriptions $\characterization_{\synth,\real}$,
    frame=tlrb,backgroundcolor=\color{light-gray}]
"Dataset B consistently specifies the analyst behind actions...",
"Dataset B maintains strict financial focus without political or
    entertainment tangents present in A...",
"Dataset B entries always directly connect stock movements to specific 
    analyst actions...",
"Dataset B shows more frequent price target amount disclosures...",
"Dataset B uses standardized financial terminology consistently...",
"Dataset B maintains neutral tone in earnings reports...",
"Dataset B focuses exclusively on institutional analyst
    perspectives...",
"Dataset B headlines strictly follow '[Analyst] [Action] on [Ticker]
    [Rationale]' structure...",
"Dataset B contains no social media tags/hashtags...",
"Dataset B shows higher frequency of ETF coverage..."
  \end{lstlisting}

  
  \begin{lstlisting}[label=lst:rubric_example_rfs,caption=Example characteristic descriptions $\characterization_{\real,\synth}$,frame=tlrb,backgroundcolor=\color{light-gray}]
"Dataset B includes headlines without stock tickers ...",
"Dataset B contains non-financial news ...",
"Dataset B incorporates social media-style commentary...",
"Dataset B includes international/non-English company names...",
"Dataset B references non-institutional analysts/sources...",
"Dataset B features headlines about dividends...",
"Dataset B includes legal/regulatory actions unrelated to markets...",
"Dataset B uses technical trading jargon...",
"Dataset B contains macroeconomic commentary without stock links...",
"Dataset B includes non-company-specific index/currency forecasts..."
\end{lstlisting}
\label{lst:rubric_example}
}

\subsection{LLM Usage}
\paragraph{Model serving}
We use vLLM \citep{kwon_efficient_2023} to serve open source models such as for data synthesis and dataset scoring. For \texttt{Llama3.3-70B-Instruct} model, we use Ollama \footnote{\url{https://ollama.com/}} Q4\_K quantized version to construct synthetic datasets for sentiment analysis. We use 2 * Nvidia A40 48GB GPUs for other models for synthesis and scoring. To improve LLMs' generation throughput, we use vLLM's batched inference feature and enable prefix-caching to further improve generation efficiency.

\paragraph{LLM hyperparameter}
For \ourmethodnameshort\ rubric compilation and scoring, we set temperature to 0 and top\_p to 0.95. 

\subsubsection{Data Synthesis Prompts}
\begin{lstlisting}[label=lst:sa_synthesis_prompt,caption={Zero-shot prompt used for sentiment analysis dataset generation.},
    frame=tlrb,backgroundcolor=\color{light-gray}, breaklines]
Generate three realistic financial news headlines for sentiment analysis.

Guidelines for Generating Headlines:

Sentiment Labeling:
Each headline must be assigned a sentiment label based on its tone:
    -   Bearish (0): Indicates negative sentiment about a stock or market trend.
    -   Bullish (1): Indicates positive sentiment about a stock or market trend.
    -   Neutral (2): Indicates neutral or informational tone.

Now, generate three new financial news headlines following these guidelines. Please use JSON format and generate one type of each sentiment label (0, 1, 2) in your response.
\end{lstlisting}

\begin{lstlisting}[label=lst:sa_synthesis_prompt_background,caption=Zero-shot with background knowledge prompt used for sentiment analysis dataset generation.,
    frame=tlrb,backgroundcolor=\color{light-gray}, breaklines]

Generate three realistic financial news headlines about stock tickers following real-world financial reporting for sentiment analysis.

Guidelines for Generating Headlines:

1. Format & Style:
    -	Headlines must be concise and mimic real financial news.
    -	Use sentence case formatting (capitalize only the first word and proper nouns).
    -	Some headlines should start with a stock ticker (e.g., $AAPL -), while others should begin with the company name or a broader market trend.

2. Ticker Inclusion:
    -	At least one headline should include a stock ticker (e.g., $TSLA - or $NVDA -).
    -	Some headlines should refer to companies by name instead of tickers (e.g., "Alphabet and Meta see price targets cut at Barclays").

3. Common Financial Themes:

Ensure headlines reflect realistic financial news topics,
including:
    -	Stock downgrades/upgrades
    -	Price target adjustments
    -	Market trends/economic outlook
    -	Company performance concerns
    -	Company news
    -	Company announcements
    -	Company events

4. Source Attribution:
    -	When relevant, mention an investment firm, analyst, or research group (e.g., Morgan Stanley, Barclays, Oppenheimer).
    -	Do not fabricate research firms-use only well-known institutions.

5. Sentiment Labeling:

Each headline must be assigned a sentiment label based on its tone:
    -	Bearish (0): Indicates negative sentiment about a stock or market trend.
    -	Bullish (1): Indicates positive sentiment about a stock or market trend.
    -	Neutral (2): Indicates neutral or informational tone.

Sentiment Labeling:
Each headline must be assigned a sentiment label based on its tone:
    - Bearish (0): Indicates negative sentiment about a stock or market trend.
    - Bullish (1): Indicates positive sentiment about a stock or market trend.
    - Neutral (2): Indicates neutral or informational tone.

Now, generate three new financial news headlines following these guidelines. Please use JSON format and generate one type of each sentiment label (0, 1, 2) in your response.
\end{lstlisting}

\begin{lstlisting}[label=lst:sa_synthesis_prompt_background_ticker,caption=Zero-shot with background and stock ticker information prompt used for sentiment analysis dataset generation.,
    frame=tlrb,backgroundcolor=\color{light-gray}, breaklines]
Generate three realistic financial news headlines about stock tickers following real-world financial reporting for sentiment analysis.

Guidelines for Generating Headlines:

1. Format & Style:
    -	Headlines must be concise and mimic real financial news.
    -	Use sentence case formatting (capitalize only the first word and proper nouns).
    -	Some headlines should start with a stock ticker (e.g., $AAPL -), while others should begin with the company name or a broader market trend.

2. Ticker Inclusion:
    -	At least one headline should include a stock ticker (e.g., $TSLA - or $NVDA -).
    -	Some headlines should refer to companies by name instead of tickers (e.g., "Alphabet and Meta see price targets cut at Barclays").

3. Common Financial Themes:

Ensure headlines reflect realistic financial news topics, including:
    -	Stock downgrades/upgrades
    -	Price target adjustments
    -	Market trends/economic outlook
    -	Company performance concerns
    -	Company news
    -	Company announcements
    -	Company events

4. Source Attribution:
    -	When relevant, mention an investment firm, analyst, or research group (e.g., Morgan Stanley, Barclays, Oppenheimer).
    -	Do not fabricate research firms-use only well-known institutions.

5. Sentiment Labeling:

Each headline must be assigned a sentiment label based on its tone:
    -	Bearish (0): Indicates negative sentiment about a stock or market trend.
    -	Bullish (1): Indicates positive sentiment about a stock or market trend.
    -	Neutral (2): Indicates neutral or informational tone.

Now, generate three new financial news headlines about stock tickers: {stock_ticker} following these guidelines. Please use JSON format and generate one type of each sentiment label (0, 1, 2) for diversity.
\end{lstlisting}


\subsubsection{Data Cleaning Prompts}
\begin{lstlisting}[label=lst:filter_prompt,caption=Prompt used for filtering noisy synthetic images,frame=tlrb,backgroundcolor=\color{light-gray}, breaklines]
You are a helpful assistant that filters out an image. You will be given an image and its corresponding text caption.

You should return true if the primary object in the image is not a ${label} in common sense. Return false otherwise.

Image:
{image}

Caption:
{caption}
\end{lstlisting}

\subsubsection{\ourmethodnameshort\ Rubric Compilation Prompts}
\label{sec:rubric_compilation_prompt}
\begin{lstlisting}[label=lst:sa_rubric_prompt_com,caption=Rubric compilation prompt used in sentiment analysis (commonalities), frame=tlrb,backgroundcolor=\color{light-gray}, breaklines]
You are a world class data analyst on financial news headline. You will be given some financial news headline samples from dataset A and dataset B. Based on the provided similar characteristics between them, list how B is similar to A. Return {num} points as a JSON list of strings. Please focus on specific and granular similarities between the two datasets, your generated characteristic points should apply to all the samples from the two datasets.

Samples from A:
{A}

Samples from B:
{B}
\end{lstlisting}

\begin{lstlisting}[label=lst:sa_rubric_prompt_diff,caption=Rubric compilation prompt used in sentiment analysis (differences), frame=tlrb,backgroundcolor=\color{light-gray}, breaklines]
You are a world class data analyst on financial news headlines. You will be given some financial news headline samples from dataset A and dataset B. Based on the provided similar characteristics between them, list how B is {feedback} A. Please focus on granular differences between the two datasets, your generated characteristic points should apply to all the samples from the corresponding dataset (A or B). Return {num} points as a JSON list of strings.

Similar characteristics between A and B:
{similar_points}

Samples from A:
{A}

Samples from B:
{B}
\end{lstlisting}

\begin{lstlisting}[label=lst:text2sql_rubric_prompt_com,caption=Rubric compilation prompt used in Text2SQL (commonalities), frame=tlrb,backgroundcolor=\color{light-gray}, breaklines]
You are a world class data analyst on database queries in natural language. Given samples from dataset A and dataset B, list how B is {feedback} A. Return {num} points as a JSON list of strings. Please focus on specific and granular similarities between the two datasets, your generated characteristic points should apply to all the samples.

Question samples from A:
{A}

Question samples from B:
{B}
\end{lstlisting}
\begin{lstlisting}[label=lst:text2sql_rubric_prompt_diff,caption=Rubric compilation prompt used in Text2SQL (differences), frame=tlrb,backgroundcolor=\color{light-gray}, breaklines]
You are a world class data analyst on database queries in natural language. Given query samples from dataset A and dataset B. Based on the provided similar characteristics between them, list how B is {feedback} A. Return {num} points as a JSON list of strings. Please focus on specific and granular differences between the two datasets, your generated characteristic points should apply to all the samples from the corresponding dataset (A or B).

Similar characteristics between A and B:
{similar_points}

Question samples from A:
{A}

Question samples from B:
{B}
\end{lstlisting}
\begin{lstlisting}[label=lst:image_rubric_prompt_com,caption=Rubric compilation prompt used in image classification (commonalities), frame=tlrb,backgroundcolor=\color{light-gray}, breaklines]
Below are {num_image} images from dataset B:
{B}

Below are {num_images} images from dataset A:
{A}

Given some samples from image classification dataset A and dataset B, list how dataset B is similar to dataset A. Return ${num_points} points that summarize the similar characteristics of the two datasets. Focus on the characteristics of the image in terms of how they are structured, styled, or captured (e.g., lighting, background, composition, etc.) rather than the image specifications such as resolution, size, etc. Your generated characteristic points should apply to all the samples from the corresponding dataset (A or B). Output should be a JSON list of strings.

Your listed points:
\end{lstlisting}
\begin{lstlisting}[label=lst:image_rubric_prompt_diff,caption=Rubric compilation prompt used in image classification (differences), frame=tlrb,backgroundcolor=\color{light-gray}, breaklines]
Below are {num_image} images from dataset B:
{B}

Below are {num_images} images from dataset A:
{A}

Given some samples from image classification dataset A and dataset B, list how dataset B is different from dataset A. Similar characteristics are provided below for reference. Return ${num_points} points that summarize the characteristics of the two datasets (e.g., dataset A is ... dataset B is ...). Focus on the characteristics of the images in terms of how they are structured, styled, or captured (e.g., lighting, background, composition, etc.) rather than the image specifications such as resolution, size, etc. Your generated characteristic points should apply to all the samples from the corresponding dataset (A or B). Output should be a JSON list of strings.

Similar characteristics:
${similar_characteristics}

Your listed points:
\end{lstlisting}
\begin{lstlisting}[label=lst:web_rubric_prompt_com,caption=Rubric compilation prompt used in web navigation (commonalities), frame=tlrb,backgroundcolor=\color{light-gray}, breaklines]
Given some samples of web navigation tasks and the web accessibility tree of dataset A and dataset B, list how B is {feedback} A. Return {num} points that summarize the characteristics of the two datasets. The two accessibility trees generated from the same website are provided. Please only list out characteristics that are related to the proposed web navigation tasks, the accessibility trees are provided to only help you understand the context of the proposed tasks, therefore do not mention the accessibility tree in your response. Please focus on granular and specific characteristics, and your generated characteristic points should apply to all the samples. Output should be a JSON list of strings. 

Accessibility Tree for dataset A:
{A_tree}

Sampled web navigation tasks from dataset A:
{A}

Accessibility Tree for dataset B:
{B_tree}

Sampled web navigation tasks from dataset B:
{B}
\end{lstlisting}
\begin{lstlisting}[label=lst:web_rubric_prompt_diff,caption=Rubric compilation prompt used in web navigation (differences), frame=tlrb,backgroundcolor=\color{light-gray}, breaklines]
Given some samples of proposed web navigation tasks and the web accessibility tree of dataset A and dataset B. Based on the similar characteristics between them. List how B is {feedback} A. Return {num} points that summarize the characteristics of the two datasets. The two accessibility trees of the same website are provided. Please only list out characteristics that are related to the proposed web navigation tasks, the accessibility trees are only provided to help you understand the context of the proposed tasks. Therefore do not list any characteristics that are related to the accessibility tree in your response. Please focus on granular and specific characteristics, and your generated characteristic points should apply to all samples in corresponding dataset (A or B). Output should be a JSON list of strings.

Similar characteristics between A and B:
{similar_points}

Accessibility Tree for dataset A:
{A_tree}

Sampled web navigation tasks from dataset A:
{A}

Accessibility Tree for dataset B:
{B_tree}

Sampled web navigation tasks from dataset B:
{B}
\end{lstlisting}

\subsubsection{\ourmethodnameshort\ Scoring Prompts}
\label{sec:scoring_prompt}
\begin{lstlisting}[label=lst:sa_scorer_prompt,caption=Scorer prompt used in sentiment analysis, frame=tlrb,backgroundcolor=\color{light-gray}, breaklines]
You are given similarities and differences between two financial news headline datasets A and B.

Your task is to judge how likely is the given financial news headline comes from dataset {prediction}. Answer your judgement with one of the following strings: "very unlikely", "unlikely", "unsure", "likely", and "very likely". 

Similar characteristics between dataset A and B:
{similar_characteristics}

Differences between dataset A and B:
{differences}

Financial news headline sample to be judged:
{example}

Your judgement in JSON format:
\end{lstlisting}
\vspace{1em}
\begin{lstlisting}[label=lst:text2sql_scorer_prompt,caption=Scorer prompt used in Text2SQL, frame=tlrb,backgroundcolor=\color{light-gray}, breaklines]
You are given similarities and differences between datasets A and B about database queries in natural language.

Your task is to judge how likely is the given database query in natural language comes from dataset {prediction}. Answer your judgement with one of the following strings: "very unlikely", "unlikely", "unsure", "likely", and "very likely".


Similar characteristics between dataset A and B:
{similar_characteristics}

Differences between dataset A and B:
{differences}

Natural language database query to be judged:
{example}

Your judgement in JSON format:
\end{lstlisting}
\vspace{1em}
\begin{lstlisting}[label=lst:image_scorer_prompt,caption=Scorer prompt used in image classification, frame=tlrb,backgroundcolor=\color{light-gray}, breaklines]
You are given similarities and differences between datasets A and B.

Your task is to judge how likely is the given image comes from dataset {prediction}. Answer your judgement with one of the following strings: "very unlikely", "unlikely", "unsure", "likely", and "very likely".

Format:
{format_instructions}

Similar characteristics between dataset A and B:
{similar_characteristics}

Differences between dataset A and B:
{differences}

Image to be judged:
{image}

your judgement in JSON format:
\end{lstlisting}


\begin{lstlisting}[label=lst:web_scorer_prompt,caption=Scorer prompt used in web navigation, frame=tlrb,backgroundcolor=\color{light-gray}, breaklines]
You are given similar and different characteristics between two datasets A and B consisting of web navigation tasks.

Your objective is to judge how likely is the given web browsing task comes from dataset {prediction}. Answer your judgement with one of the following strings: "very unlikely", "unlikely", "unsure", "likely", and "very likely".

Format:
{format_instructions}

Similar characteristics between dataset A and B:
{similar_characteristics}

Different characteristics between dataset A and B:
{differences}

Web navigation task to be judged:
{example}

Your judgement in JSON format:
\end{lstlisting}
\end{document}